\documentclass{article} % For LaTeX2e
\pdfoutput=1
\usepackage{iclr2017_conference,times}
\usepackage{hyperref}
\usepackage{url}
\usepackage{mathtools}
\usepackage{subcaption}

\usepackage[utf8]{inputenc} % allow utf-8 input
\usepackage[T1]{fontenc}    % use 8-bit T1 fonts
\usepackage{hyperref}       % hyperlinks
\usepackage{url}            % simple URL typesetting
\usepackage{booktabs}       % professional-quality tables
\usepackage{amsfonts}       % blackboard math symbols
\usepackage{nicefrac}       % compact symbols for 1/2, etc.
\usepackage{microtype}      % microtypography
\usepackage{caption}
\usepackage{graphicx}
\usepackage{amsmath,amssymb} % define this before the line numbering.
\usepackage{color}
\usepackage[ruled]{algorithm}
\usepackage{algpseudocode}
\usepackage{tabularx}
\usepackage{multirow}
\usepackage{xcolor}

\usepackage{color}
\definecolor{gray}{rgb}{0.5,0.5,0.5}

\iftrue  % do NOT display comments
        \newcommand{\todo}[1]{}
        \newcommand{\outline}[1]{}
        \newcommand{\textgray}[1]{}
        \newcommand{\commenttext}[1]{}
        \newcommand{\commentfoot}[1]{}
        \newcommand{\commentselfoot}[2]{}
        \newcommand{\topic}[1]{}
\else 
        \newcommand{\todo}[1]{{\textcolor{red}{[[TODO: {#1}]]}}}
        \newcommand{\outline}[1]{{\textcolor{blue}{[[{#1}]]}}}
        \newcommand{\textgray}[1]{\textcolor{gray}{[[{#1}]]}}
        \newcommand{\commenttext}[1]{\textcolor{red}{[[{#1}]]}}
        \newcommand{\commentfoot}[1]{\footnote{\textcolor{red}{\textit{#1}}}}
        \newcommand{\commentselfoot}[2]{{\textcolor{blue}{#1}}\commenttext{#2}}
        \newcommand{\topic}[1]{\textcolor{gray}{\textbf{(#1.)}}}
\fi

\iftrue % use space-saving macro
        \newcommand{\cutsectionup}{\vspace*{-0.05in}}
        \newcommand{\cutsectiondown}{\vspace*{-0.05in}}

\else % do not use space-saving macro
        \newcommand{\cutsectionup}{}
        \newcommand{\cutsectiondown}{}

\fi

\iclrfinalcopy
\title{Decomposing Motion and Content for \\Natural Video Sequence Prediction}

\author{Ruben Villegas\textsuperscript{\textnormal{1}} 
\quad Jimei Yang\textsuperscript{\textnormal{2}} 
\quad Seunghoon Hong\textsuperscript{\textnormal{3,\thanks{This work was done while SH and XL were visiting the University of Michigan.}}} 
\quad Xunyu Lin\textsuperscript{\textnormal{4,*}} 
\quad Honglak Lee\textsuperscript{\textnormal{1,5}} \\
\textsuperscript{\textnormal{1}}University of Michigan, Ann Arbor, USA\\
\textsuperscript{\textnormal{2}}Adobe Research, San Jose, CA 95110 \\
\textsuperscript{\textnormal{3}}POSTECH, Pohang, Korea \\
\textsuperscript{\textnormal{4}}Beihang University, Beijing, China \\
\textsuperscript{\textnormal{5}}Google Brain, Mountain View, CA 94043\\
}

\newcommand{\x}{{\mathbf{x}}}
\newcommand{\y}{{\mathbf{y}}}
\newcommand{\z}{{\mathbf{z}}}

\newcommand{\s}{{\mathbf{s}}}
\newcommand{\rr}{{\mathbf{r}}}
\newcommand{\ff}{{\mathbf{f}}}
\newcommand{\dd}{{\mathbf{d}}}
\newcommand{\sh}[1]{{\color{black}{#1}}}

\newcommand{\ignore}[1]{}

\usepackage{hyperref}

\begin{document}

\maketitle

\begin{abstract}
We propose a deep neural network for the prediction of future frames in natural video sequences. 
To effectively handle complex evolution of pixels in videos, we propose to decompose the motion and content, two key components generating dynamics in videos. 
Our model is built upon the Encoder-Decoder Convolutional Neural Network and Convolutional LSTM for pixel-level prediction, which independently capture the spatial layout of an image and the corresponding temporal dynamics.
By independently modeling motion and content, predicting the next frame reduces to converting the extracted content features into the next frame content by the identified motion features, which simplifies the task of prediction.
Our model is end-to-end trainable over multiple time steps, and naturally learns to decompose motion and content without separate training.
%We evaluate the proposed network architecture on the controlled human activity datasets of KTH and Actions as Space-Time Shapes, and the unconstrained dataset of UCF101 datasets.
We evaluate the proposed network architecture on human activity videos using KTH, Weizmann action, and UCF-101 datasets.
We show state-of-the-art performance in comparison to recent approaches.
To the best of our knowledge, this is the first end-to-end trainable network architecture with motion and content separation to model the spatio-temporal dynamics for pixel-level future prediction in natural videos.
\end{abstract}
% \citep{convlstm}
%~\citep{ActionsAsSpaceTimeShapes_pami07}
%~\citep{Kth}
%\citep{Ucf} 

% Introduction ------------------------------------------------------------------------------------------------------------------------------------------------------------
\cutsectionup
\section{Introduction}
\label{sec:intro}
\cutsectiondown

Understanding videos has been one of the most important tasks in the field of computer vision.
Compared to still images, the temporal component of videos provides much richer descriptions of the visual world, such as interaction between objects, human activities, and so on.
Amongst the various tasks applicable on videos, the task of anticipating the future has recently received increased attention in the research community. 
Most prior works in this direction focus on predicting high-level semantics in a video such as action~\citep{Vondrick15,Ryoo11,Lan14}, event~\citep{Yuen10,Hoai13} and motion~\citep{Pintea14,Walker14,Pickup14,WalkerDGH16}.
Forecasting semantics provides information about \textit{what will happen} in a video, and is essential to automate decision making.
However, the predicted semantics are often specific to a particular task and provide only a partial description of the future. 
Also, training such models often requires heavily labeled training data which leads to tremendous annotation costs especially with videos.

%\ifdefined\paratitle {\color{blue} [our objective: frame prediction. ]\\ } 
In this work, we aim to address the problem of prediction of future frames in natural video sequences.
Pixel-level predictions provide dense and direct description of the visual world, and existing video recognition models can be adopted on top of the predicted frames to infer various semantics of the future.
%In addition, a successful prediction of future frames requires non-trivial internal representations of the input even without specified constraints.
%Specifically, unlike in the problem of image reconstruction \citep{DBLP:conf/icml/VincentLBM08}, learning an identity function on the input will not give a reasonable outcome.
% Pixels in videos exhibit strong temporal and spatial correlations useful for frame prediction, which enables purely unsupervised training of the model by observing raw video frames.
Spatio-temporal correlations in videos provide a self-supervision for frame prediction, which enables purely unsupervised training of a model by observing raw video frames.
Unfortunately, estimating frames is an extremely challenging task; not only because of the inherent uncertainty of the future, but also various factors of variation in videos leading to complicated dynamics in raw pixel values.
%In addition, a successful prediction of future frames requires non-trivial internal representations of the input even without specified constraints.
%Specifically, unlike in the problem of image reconstruction \citep{DBLP:conf/icml/VincentLBM08}, learning an identity function on the input will not give a reasonable outcome.
%There have been a number of recent attempts on frame prediction~\citep{Srivastava15,Mathieu15,Oh15,Goroshin15,Lotter15,Ranzato14}, yet they all use a single encoder that should be able to reason about all the different variations occurring in videos in order to make predictions of the future.
There have been a number of recent attempts on frame prediction~\citep{Srivastava15,Mathieu15,Oh15,Goroshin15,Lotter15,Ranzato14}, which use a single encoder that needs to reason about all the different variations occurring in videos in order to make predictions of the future, or require extra information like foreground-background segmentation masks and static background~\citep{Vondrick16}.
% More recently,~\citep{visualdynamics16} proposed a probabilistic model that observes difference image and raw image to generate the image difference that converts the input frame into the next frame.
% This is the closest work to ours as it uses a motion, and image encoders based on CNNs.
% The main differences are: (1) Our model is not probabilistic, (2) our motion encoder is based Convolutional LSTM~\citep{convlstm}, which is intuitively designed to model sequence dynamics and long term dependencies of its input, (3) our content encoder only observes a single dimension of the input image, and (4) we directly generate the pixel values of the future frames.

% they are often limited to short-term prediction of simple dynamics (at most a few frames) due to the difficulties in modeling complex variations of pixels; it is not clear whether the model is able to capture semantically meaningful dynamics in longer time steps.  
%Several architectures has been tried with different choices of architecture~\citep{}, loss functions~\citep{} and modeling of dynamics~\citep{}. 
%However, the previous attempts in prediction of future frames are often limited to short-term prediction of simple dynamics ($\leq2$ frames), and it is not clear whether the model is able to capture semantically meaningful dynamics in longer time steps.

We propose a Motion-Content Network (MCnet) for robust future frame prediction.
Our intuition is to split the inputs for video prediction into two easily identifiable groups, motion and content, and independently capture each information stream with separate encoder pathways.
%In this architecture, the \textit{contents} pathway encodes the spatial layout of the salient part of an image, while the \textit{motion} pathways encodes the local dynamics of spatial regions.
In this architecture, the \textit{motion} pathway encodes the local dynamics of spatial regions, while the \textit{content} pathway encodes the spatial layout of the salient parts of an image.
The prediction of the future frame is then achieved by transforming the content of the last observed frame given the identified dynamics up to the last observation. %which are captured by contents and motion pathways, respectively.
%The high-level representations of contents and motions are then combined to generate pixel-level prediction of the future frame through decoder. 
%
\iffalse
Modeling the motion and content in videos with separate encoder pathways provides number of benefits for future frame prediction.
First, decomposing the two sources of information significantly simplifies of the prediction task.
Specifically, it allows our network to focus on identifying temporal and spatial features separately, which in turn reduces the prediction problem to converting content features from the last observed time step to the next using the motion features.
\fi
%
Somewhat surprisingly, we show that such a network is end-to-end trainable \emph{without individual path way supervision}. Specifically, we show that an asymmetric architecture for the two pathways enables such decompositions without explicit supervision.
The contributions of this paper are summarized below:
\begin{itemize}
\item We propose MCnet for the task of frame prediction, which separates the information streams (motion and content) into different encoder pathways. 
\item The proposed network is end-to-end trainable and naturally learns to decompose motion and content without separate training, and reduces the task of frame prediction to transforming the last observed frame into the next by the observed motion.
% \item The proposed network reduces the task of frame prediction to transforming the last observed frame into the next by the observed motion.
% \item The proposed model is evaluated on real-world videos for long-term  
\item We evaluate the proposed model on challenging real-world video datasets, and show that it outperforms previous approaches on frame prediction. 
\end{itemize}

The rest of the paper is organized as follows.
We briefly review related work in Section~\ref{sec:relatedwork}, and introduce an overview of the proposed algorithm in Section~\ref{sec:overview}. 
The detailed configuration of the proposed network is described in Section~\ref{sec:architecture}.
Section~\ref{sec:train_infer} describes training and inference procedure.
Section~\ref{sec:experiments} illustrates implementation details and experimental results on challenging benchmarks. %datasets.

% related works ------------------------------------------------------------------------------------------------------------------------------------------------------------
\cutsectionup
\section{Related work}
\commenttext{Section needs revising and update with current literature.}
\label{sec:relatedwork}
\cutsectiondown

% \ifdefined\paratitle {\color{blue} 
%1. general sequence estimation (language modeling, etc.)\\
% 1. general video understanding\\/
% 2. high-level video prediction (two-stream network, etc.)\\
% 3. pixel-level sequence estimation with natural videos (Jun's, etc.)\\
% 4. disentangling motion and contents.\\
%3. other references might be related to this paper\\
% } \fi

% \ifdefined\paratitle {\color{blue} [general video understanding]\\ }\fi

\ifdefined\paratitle {\color{blue} [future prediction in a video]\\ }\fi
The problem of visual future prediction has received growing interests in the computer vision community.
It has led to various tasks depending on the objective of future prediction, such as human activity~\citep{Vondrick15,Ryoo11,Lan14}, event~\citep{Yuen10,Hoai13} and geometric path~\citep{Walker14}.
Although previous work achieved reasonable success in specific tasks, they are often limited to estimating predefined semantics, and require fully-labeled training data. %specifically collected for the each prediction task.  
To alleviate this issue, approaches predicting representation of the future beyond semantic labels have been proposed.
\citet{Walker14} proposed a data-driven approach to predict the motion of a moving object, and coarse hallucination of the predicted motion.
\citet{Vondrick15} proposed a deep regression network to predict feature representations of the future frames.
These approaches are supervised and provide coarse predictions of how the future will look like.
Our work also focuses on unsupervised learning for prediction of the future, but to a more direct visual prediction task: frame prediction.

\ifdefined\paratitle {\color{blue} [frame prediction on videos]\\ }\fi
% Compared to prediction of high-level semantics, pixel-level prediction requires more fine-grained understanding of underlying contents and dynamics, and has been less investigated due to the lack of powerful model for capturing patterns from noisy data.
% Compared to predicting semantics, pixel-level prediction has been less investigated due to the lack of powerful model for modeling noisy v in video pixels.
Compared to predicting semantics, pixel-level prediction has been less investigated due to the difficulties in modeling evolution of raw pixels \sh{over time}.
% Recent success in deep learning has enabled to model complex sequential data, and started to apply in frame prediction recently.   
% Frame prediction has been less investigated due to these challenges  
% Recent approaches on frame prediction are built upon the deep neural network due to its powerfulness in modeling sequential data.
Fortunately, recent advances in deep learning provide a powerful tool for sequence modeling, and enable the creation of novel architectures for modeling complex sequential data.  
\cite{Ranzato14} applied a recurrent neural network developed for language modeling to frame prediction by posing the task as classification of each image region to one of quantized patch dictionaries. 
%\citep{Srivastava15} employed Long Short Term Memory (LSTM) for frame prediction, wher
\cite{Srivastava15} applied a sequence-to-sequence model to video prediction, and showed that Long Short-Term Memory (LSTM) is able to capture pixel dynamics.
\cite{Oh15} proposed an action-conditional encoder-decoder network to predict future frames in Atari games.
In addition to the different choices of architecture, some other works addressed the importance of selecting right objective function:
\cite{Lotter15} used adversarial loss with combined CNN and LSTM architectures, and \cite{Mathieu15} employed similar adversarial loss with additional regularization using a multi-scale encoder-decoder network.
\cite{FinnGL16} constructed a network that predicts transformations on the input pixels for next frame prediction.
\cite{DBLP:journals/corr/PatrauceanHC15} proposed a network that by explicitly predicting optical flow features is able to predict the next frame in a video.
\cite{Vondrick16} proposed a generative adversarial network for video which, by generating a background-foreground mask, is able to generate realistic-looking video sequences.
However, none of the previously mentioned approaches exploit spatial and temporal information separately in an unsupervised fashion.
In terms of the way data is observed, the closest work to ours is \citet{visualdynamics16}.
The differences are (1) Our model is deterministic and theirs is probabilistic, (2) our motion encoder is based on convolutional LSTM \citep{convlstm} which is a more natural module to model long-term dynamics, (3) our content encoder observes a single scale input and theirs observes many scales, and (4) we directly generate image pixels values, which is a more complicated task.
We aim to exploit the existing spatio-temporal correlations in videos by decomposing the motion and content in our network architecture.
% We aim to alleviate this issue by decomposing the factors of variations in video, contents and motion.
% Although the prior approaches have achieved reasonable results on simple videos, most of them fails in prediction of real-world videos after few iterations. 
% It has not been clearly shown that the proposed approaches are capable to capture longer-term motion patterns in real-world videos. 
%and there is still many rooms for improvement for long-term prediction on real-world videos.

% Built upon deep neural network designed for recognition of sequence data, 

\ifdefined\paratitle {\color{blue} [Disentangling motion and contents in videos]\\ }\fi
% The idea of decomposing the videos into spatial and temporal components has been actively investigated in various tasks.
To the best of our knowledge, the idea of separating motion and content has not been investigated in the task of unsupervised deterministic frame prediction. 
The proposed architecture shares similarities to the two-stream CNN~\citep{Simonyan14}, which is designed for action recognition to jointly exploit the information from frames and their temporal dynamics. 
\sh{However, in contrast to their network we aim to learn features for temporal dynamics directly from the raw pixels, and we use the identified features from the motion in combination with spatial features to make pixel-level predictions of the future. }
% Our work is extension of these ideas to a more challenging problem, frame prediction.
% frame prediction requires to establish more close and direct connection between spatial and temporal components. 

% general sequence estimation

% video prediction

% other references

% % Architectures ------------------------------------------------------------------------------------------------------------------------------------------------------------
\cutsectionup
\section{Algorithm Overview}
\label{sec:overview}
\cutsectiondown

%\ifdefined\paratitle {\color{blue} [definition of frame prediction]\\ } 
%This paper tackles the problem of frame prediction in natural video sequences.
In this section, we formally define the task of frame prediction and the role of each component in the proposed architecture.
Let $\x_t\in\mathrm{R}^{w\times h\times c}$ denote the $t$-th frame in an input video $\x$, where $w, h$, and $c$ denote width, height, and number of channels, respectively.
The objective of frame prediction is to generate the future frame $\hat{\x}_{t+1}$ given the input frames $\x_{1:t}$. 

At the $t$-th time step, our network observes a history of previous consecutive frames up to frame $t$, and generates the prediction of the next frame $\hat{\x}_{t+1}$ as follows:

\begin{itemize}
\item
\textbf{Motion Encoder} recurrently takes an image difference input between frame $\x_t$ and $\x_{t-1}$ starting from $t=2$, and produces the hidden representation $\dd_t$ encoding the temporal dynamics of the scene components (Section~\ref{sec:dynamic_network}).

\item   
\textbf{Content Encoder} takes the last observed frame $\x_t$ as an input, and outputs the hidden representation $\s_t$ that encodes the spatial layout of the scene (Section~\ref{sec:contents_network}).  

\item
\textbf{Multi-Scale Motion-Content Residual} takes the computed features, from both the motion and content encoders, at every scale right before pooling and computes residuals $\rr_t$ \citep{resnets} to aid the information loss caused by pooling in the encoding phase (Section~\ref{sec:mcres_network}). 

\item  
\textbf{Combination Layers and Decoder} takes the outputs from both encoder pathways and residual connections,  $\dd_t$, $\s_t$, and $\rr_t$, and combines them to produce a pixel-level prediction of the next frame $\hat{\x}_{t+1}$ (Section~\ref{sec:decoder_network}). 
\end{itemize}

\begin{figure}[!t]
    \hspace*{-.1cm}
    \centering
    \begin{subfigure}{0.48\linewidth}
        \vspace{-7pt}
	    \includegraphics[width=1\linewidth]{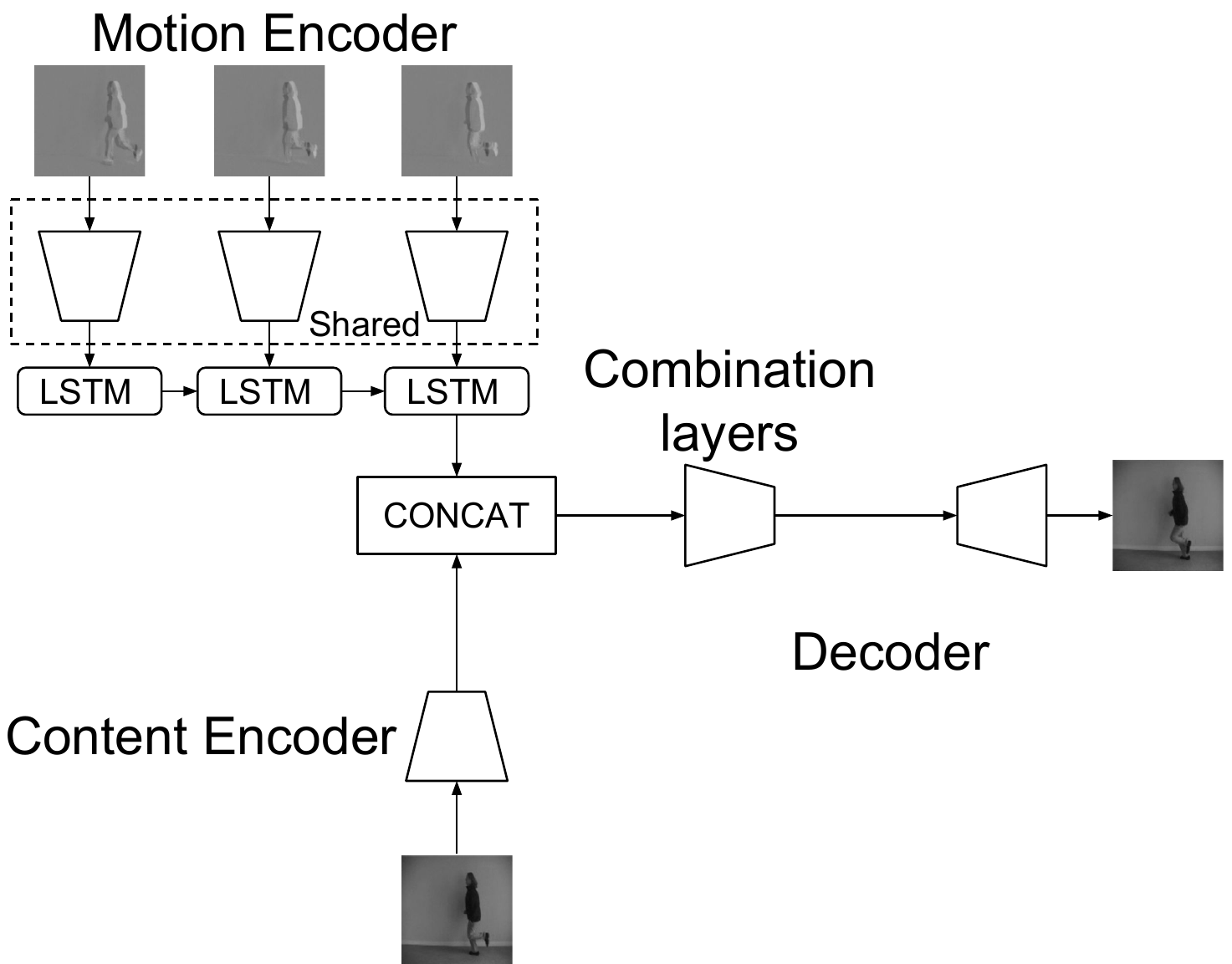}
	    \caption{Base MCnet}
	\end{subfigure}
	\hspace*{.4cm}
    \begin{subfigure}{0.48\linewidth}
        \vspace{-7pt}
	    \includegraphics[width=1\linewidth]{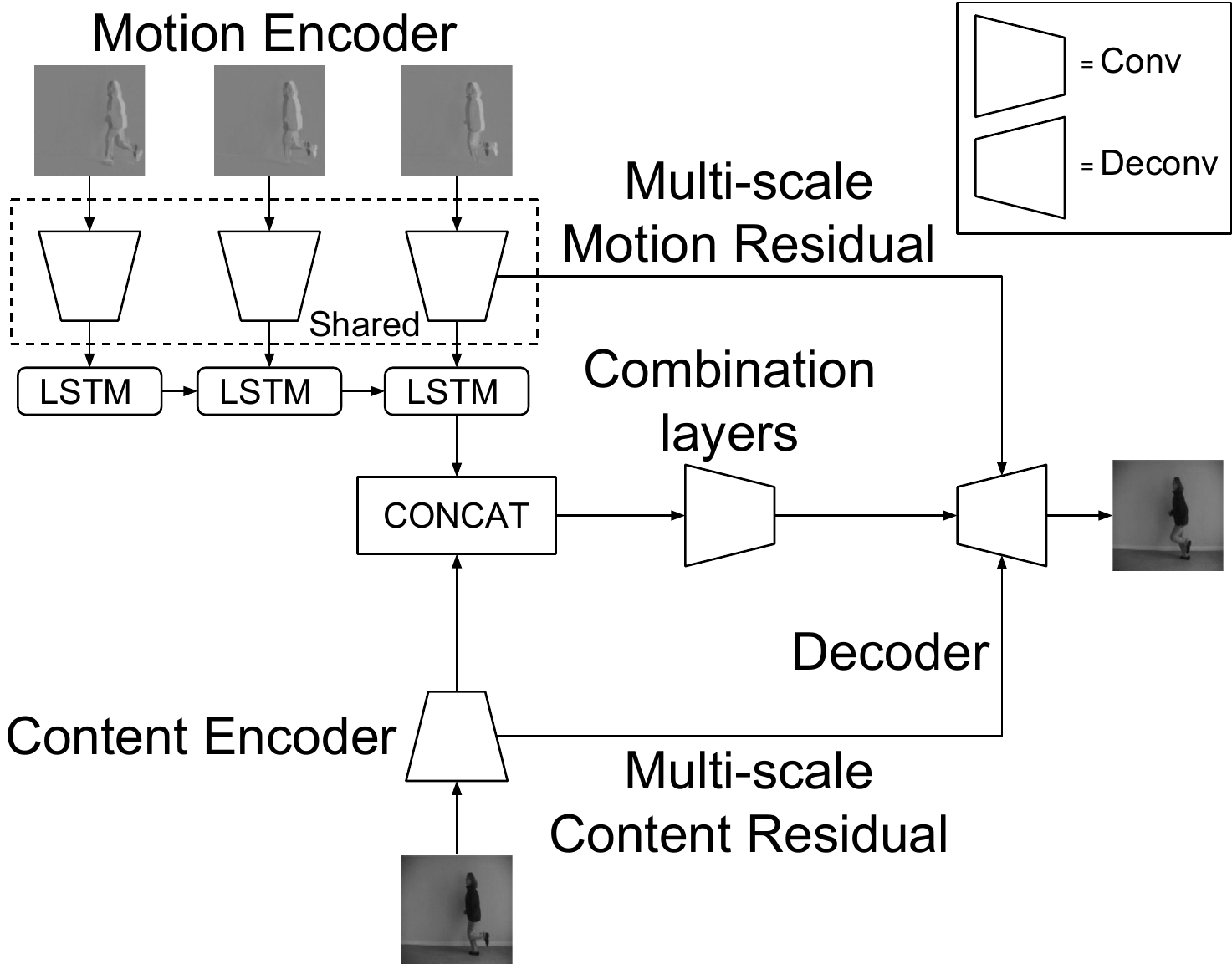}
	    \caption{MCnet with Multi-scale Motion-Content Residuals}
	\end{subfigure} 

	\vspace{-5pt}
    \caption{Overall architecture of the proposed network. (a) illustrates MCnet without the Motion-Content Residual \textit{skip connections}, and (b) illustrates MCnet with such connections. Our network observes a history of image differences through the motion encoder and last observed image through the content encoder. Subsequently, our network proceeds to compute motion-content features and communicates them to the decoder for the prediction of the next frame.}
\label{fig:arch}
\vspace{-15pt}
\end{figure}

The overall architecture of the proposed algorithm is described in Figure~\ref{fig:arch}.
The prediction of multiple frames, $\hat{\x}_{t+1:t+T}$, can be achieved by recursively performing the above procedures over $T$ time steps (Section~\ref{sec:train_infer}). 
Each component in the proposed architecture is described in the following section.

\cutsectionup
\section{Architecture}
\label{sec:architecture}
\cutsectiondown

This section describes the detailed configuration of the proposed architecture, including the two encoder pathways, multi-scale residual connections, combination layers, and decoder.  

\subsection{Motion Encoder}
\label{sec:dynamic_network}

The motion encoder captures the temporal dynamics of the scene's components by recurrently observing subsequent difference images computed from $\x_{t-1}$ and $\x_{t}$, and outputs motion features by
\begin{equation}
\left[\dd_t,\mathbf{c}_{t}\right]=f^{\text{dyn}}\left(\x_t-\x_{t-1}, \dd_{t-1}, \mathbf{c}_{t-1}\right),
\label{eq:motion_encoder}
\end{equation}
where $\x_t-\x_{t-1}$ denotes element-wise subtraction between frames at time $t$ and $t-1$, $\dd_t\in\mathbb{R}^{{w'}\times{h'}\times{c'}}$ is the feature tensor encoding the motion across the observed difference image inputs, and $\mathbf{c}_t\in\mathbb{R}^{{w'}\times{h'}\times{c'}}$ is a memory cell that retains information of the dynamics observed through time. 
$f^{\text{dyn}}$ is implemented in a fully-convolutional way to allow our model to identify local dynamics of frames rather than complicated global motion. For this, we use an encoder CNN with a Convolutional LSTM \citep{convlstm} layer on top.

\subsection{Content Encoder}
\label{sec:contents_network}

The content encoder extracts important spatial features from a single frame, such as the spatial layout of the scene and salient objects in a video. 
Specifically, it takes the last observed frame $\x_t$ as an input, and produces content features by
\begin{equation}
\s_t=f^{\text{cont}}\left(\x_{t}\right),
\label{eq:content_encoder}
\end{equation}
where $\s_t\in\mathbb{R}^{{w'}\times{h'}\times{c'}}$ is the feature encoding the spatial content of the last observed frame, and $f^{\text{cont}}$ is implemented by a Convolutional Neural Network (CNN) that specializes on extracting features from single frame.
%Note that the content features are not updated when predicting the entire multi-step future frames (i.e., at each time step, the predicted future frame is not fed back to the content encoder).

It is important to note that our model employs an \textit{asymmetric} architecture for the motion and content encoder.
The content encoder takes the last observed frame, which keeps the most critical clue to reconstruct spatial layout of near future, but has no information about dynamics. 
On the other hand, the motion encoder takes a history of previous image differences, which are less informative about the future spatial layout compared to the last observed frame, yet contain important spatio-temporal variations occurring over time.
This asymmetric architecture encourages encoders to exploit each of two pieces of critical information to predict the future content and motion individually, and enables the model to learn motion and content decomposition naturally without any supervision.

\subsection{Multi-scale Motion-Content Residual}
\label{sec:mcres_network}

To prevent information loss after the pooling operations in our motion and content encoders, we use residual connections \citep{resnets}.
The residual connections in our network communicate motion-content features at every scale into the decoder layers after unpooling operations. The residual feature at layer $l$ is computed by
\begin{equation}
\rr_t^l=f^{\text{res}}\left(\left[\s_t^l,\dd_t^l\right]\right)^l,
\label{eq:res_connect}
\end{equation}
where $\rr_t^l$ is the residual output at layer $l$, $\left[\s_t^l,\dd_t^l\right]$ is the  concatenation of the motion and content features along the depth dimension at layer $l$ of their respective encoders, $f^{\text{res}}\left(.\right)^l$ the residual function at layer $l$ implemented as consecutive convolution layers and rectification with a final linear layer.

\subsection{Combination Layers and Decoder}
\label{sec:decoder_network}

The outputs from the two encoder pathways, $\dd_t$ and $\s_t$, encode a high-level representation of motion and content, respectively.
Given these representations, the objective of the decoder is to generate a pixel-level prediction of the next frame $\hat{\x}_{t+1}\in\mathbb{R}^{{w}\times{h}\times{c}}$. 
To this end, it first combines the motion and content back into a unified representation by
\begin{equation}
\ff_t=g^{\text{comb}}\left(\left[\dd_t,\s_t\right]\right),
\label{eq:comb_decoder}
\end{equation}
where $\left[\dd_t,\s_t\right]\in\mathbb{R}^{w'\times h'\times 2c'}$ denotes the concatenation of the higher-level motion and content features in the depth dimension, and $\ff_t\in\mathbb{R}^{w'\times h'\times c'}$ denotes the combined high-level representation of motion and content.
$g^{\text{comb}}$ is implemented by a CNN with bottleneck layers~\citep{bottleneck}; it first projects both $\dd_t$ and $\s_t$ into a lower-dimensional embedding space, and then puts it back to the original size to construct the combined feature $\ff_t$. 
Intuitively, $\ff_t$ can be viewed as the content feature of the next time step, $\s_{t+1}$, which is generated by transforming $\s_t$ using the observed dynamics encoded in $\dd_t$.
Then our decoder places $\ff_t$ back into the original pixel space by
\begin{equation}
{\hat{\x}}_{t+1}=g^{\text{dec}}\left(\ff_t,\rr_t\right),
\label{eq:rec_decoder}
\end{equation}
where $\rr_t$ is a list containing the residual connections from every layer of the motion and content encoders before pooling sent to every layer of the decoder after unpooling.
We employ the deconvolution network~\citep{Zeiler11} for our decoder network $g^{\text{dec}}$,
which is composed of multiple successive operations of deconvolution, rectification and unpooling with the addition of the motion-content residual connections after each unpooling operation. The output layer is passed through a $\tanh\left(.\right)$ activation function.
Unpooling with fixed switches are used to upsample the intermediate activation maps.

\cutsectionup
\section{Inference and Training}
\label{sec:train_infer}
\cutsectiondown

Section~\ref{sec:architecture} describes the procedures for single frame prediction, while this section presents the extension of our algorithm for the prediction of multiple time steps. 

\subsection{Multi-step prediction}
\commenttext{Add Algorithm figure}

Given an input video, our network observes the first $n$ frames as image difference between frame $\x_t$ and $\x_{t-1}$, starting from $t=2$ up to $t=n$, to encode initial temporal dynamics through the motion encoder. The last frame $\x_n$ is given to the content encoder to be transformed into the first prediction $\hat{\x}_{t+1}$ by the identified motion features.

For each time step $t\in\left[n+1, n+T \right]$, where $T$ is the desired number of prediction steps, our network takes the difference image between the first prediction $\hat{\x}_{t+1}$ and the previous image $\x_{t}$, and the first prediction $\hat{\x}_{t+1}$ itself to predict the next frame $\hat{\x}_{t+2}$, and so forth.

% \ifdefined\paratitle {\color{blue} [Training]\\ }   
\subsection{Training Objective} 

To train our network, we use an objective function composed of different sub-losses similar to \citet{Mathieu15}. Given the training data $D=\{\x^{(i)}_{1,...,T}\}_{i=1}^{N}$, our model is trained to minimize the prediction loss by
\begin{equation}
\mathcal{L} = \alpha\mathcal{L}_{\text{img}}+\beta\mathcal{L}_{\text{GAN}},
\label{eq:full_loss}
\end{equation}
where $\alpha$ and $\beta$ are hyper-parameters that control the effect of each sub-loss during optimization. $\mathcal{L}_{\text{img}}$ is the loss in image space from \citet{Mathieu15} defined by
\begin{equation}
\mathcal{L}_{\text{img}} = \mathcal{L}_{p}\left(\x_{t+k},\hat{\x}_{t+k}\right)+\mathcal{L}_{gdl}\left(\x_{t+k},\hat{\x}_{t+k}\right),
\label{eq:loss_img}
\end{equation}
%\vspace*{-0.202in}
%
\begin{align}
\text{where \quad} \mathcal{L}_{p}\left(\y,\z\right)= & \sum_{k=1}^{T}||\y-\z||_{p}^{p}, \label{eq:lp}\\ %\text{\quad and}
\mathcal{L}_{gdl}\left(\y,\z\right)= & \sum_{i,j}^{h,w}\left| \, (|\y_{i,j}-\y_{i-1,j}|-|\z_{i,j}-\z_{i-1,j}|) \, \right|^{\lambda} \label{eq:lgdl}\\
 & +\left| \, (|\y_{i,j-1}-\y_{i,j}|-|\z_{i,j-1}-\z_{i,j}|) \, \right|^{\lambda}. \nonumber
\end{align}
Here, $\x_{t+k}$ and $\hat{\x}_{t+k}$ are the target and predicted frames, respectively, and $p$ and $\lambda$ are hyper-parameters for $\mathcal{L}_p$ and $\mathcal{L}_{gdl}$, respectively.
Intuitively, $\mathcal{L}_{p}$ guides our network to match the average pixel values directly, while $\mathcal{L}_{gdl}$ guides our network to match the gradients of such pixel values.
Overall, $\mathcal{L}_{\text{img}}$ guides our network to learn parameters towards generating the correct average sequence given the input.
Training to generate average sequences, however, results in somewhat blurry generations which is the reason we use an additional sub-loss.
$\mathcal{L}_{\text{GAN}}$ is the generator loss in adversarial training to allow our model to predict realistic looking frames and it is defined by
\begin{equation}
\mathcal{L}_{\text{GAN}} = -\log D\left(\left[\x_{1:t},G\left(\x_{1:t}\right)\right]\right) ,
\label{eq:loss_adv}
\end{equation}
where $\x_{1:t}$ is the concatenation of the input images, $\x_{t+1:t+T}$ is the concatenation of the ground-truth future images, $G\left(\x_{1:t}\right)=\hat{\x}_{t+1:t+T}$ is the concatenation of all predicted images along the depth dimension, and $D\left(.\right)$ is the discriminator in adversarial training.
The discriminative loss in adversarial training is defined by
\begin{equation}\label{eq:loss_disc}
\begin{aligned}
\mathcal{L}_{\text{disc}} &= -\log D\left(\left[\x_{1:t},\x_{t+1:t+T}\right]\right) -\log\left(1-D\left(\left[\x_{1:t},G\left(\x_{1:t}\right)\right]\right)\right).
\end{aligned}
\end{equation}
$\mathcal{L}_{\text{GAN}}$, in addition to $\mathcal{L}_{\text{img}}$, allows our network to not only generate the target sequence, but also simultaneously enforce realism in the images through visual sharpness that fools the human eye.
Note that our model uses its predictions as input for the next time-step during the training, which enables the gradients to flow through time and makes the network robust for error propagation during prediction.
For more a detailed description about adversarial training, please refer to Appendix \ref{sec:GANs}.

% \ignore{
% \lipsum[1]
% \begin{wrapfigure}{R}{0.5\textwidth}
% \begin{minipage}{0.5\textwidth}
% \begin{algorithm}
% \caption{Recursive Prediction Procedure \label{alg1}}
% \begin{algorithmic}
% \State input: $\textbf{x}_{1:n}$
% \State output: $\hat{\textbf{x}}_{n+1:n+T}$
% \State  initialize: $\x_{t-n+1:t}\gets\textbf{x}_{1:n},\ \x_{t}\gets\x_{n}$
% \For{$t$=n+1 to $n+T$}
%     \State{
%         $\begin{aligned}
%         &\textbf{h}_t^{cont}    &\gets& \ f^{cont}(\x_{t}) \\
%         &\textbf{h}_t^{dyn}     &\gets& \ f^{dyn}(\x_{t-n+1:t}) \\
% %        &\textbf{h}_t^{comb}    &\gets& \ g^{comb}([\textbf{h}_t^{dyn},\textbf{h}_t^{cont}]) \\
% %        &\hat{\textbf{x}}_{t} &\gets& \ g^{dec}(\textbf{h}_t^{comb}) \\
% 	&\hat{\x}_{t+1} &\gets& \ g^{dec}(\ g^{comb}([\textbf{h}_t^{dyn},\textbf{h}_t^{cont}])) \\
%         % \State $\hat{\textbf{x}}_{t+1}=g^{dec}(g^{comb}([f^{cont}(\textbf{x}^c), f^{dyn}(\textbf{x}^m_{t-n+1:t})]))
%         &\x_{t-n+1:t} &\gets& \ \left[\x_{t-n+2:t},\hat{\x}_{t+1}\right] \\
%         &\x_{t}           &\gets& \ \hat{\x}_{t+1}
%         \end{aligned}$
%     }
% \EndFor
% \end{algorithmic}
% \label{alg:multi_pred}
% \end{algorithm}
% \end{minipage}
% \end{wrapfigure}
% \lipsum
% }

% Experiments ------------------------------------------------------------------------------------------------------------------------------------------------------------
\cutsectionup
% !TEX root = video_prediction.tex
\section{Experiments}
\label{sec:experiments}
\cutsectiondown

In this section, we present experiments using our network for video generation. We first evaluate our network, MCnet, on the KTH~\citep{Kth} and Weizmann action~\citep{ActionsAsSpaceTimeShapes_pami07} datasets, and compare against a  baseline convolutional LSTM (ConvLSTM)~\citep{convlstm}.
We then proceed to evaluate on the more challenging UCF-101~\citep{Ucf} dataset, in which we compare against the same ConvLSTM baseline and also the current state-of-the-art method by~\citet{Mathieu15}.
For all our experiments, we use $\alpha=1$, $\lambda=1$, and $p=2$ in the loss functions.

In addition to the results in this section, we also provide more qualitative comparisons in the supplementary material and in the videos on the project website: \url{https://sites.google.com/a/umich.edu/rubenevillegas/iclr2017}.

\paragraph{Architectures.} The content encoder of MCnet is built with the same architecture as VGG16~\citep{Vgg16} up to the third pooling layer.
The motion encoder of MCnet is also similar to VGG16 up to the third pooling layer, except that we replace its consecutive $3$x$3$ convolutions with single $5$x$5$, $5$x$5$, and $7$x$7$ convolutions in each layer.
%The motion encoder is also based on VGG16 up to the third pooling layer, except that we fuse the consecutive $3$x$3$ convolution into the respective receptive field size (i.e. 3x3 followed by 3x3 is fused into a single convolution of 5x5).
The combination layers are composed of $3$ consecutive $3$x$3$ convolutions (256, 128, and 256 channels in each layer).
The multi-scale residuals are composed of $2$ consecutive $3$x$3$ convolutions.
The decoder is the mirrored architecture of the content encoder where we perform unpooling followed by deconvolution.
For the baseline ConvLSTM, we use the same architecture as the motion encoder, residual connections, and decoder, except we increase the number of channels in the encoder in order to have an overall comparable number of parameters with MCnet.
%In the ConvLSTM KTH experiments encoder, the output channel dimensions are 200, 274 and 300, respectively.
%For the UCF101 experiments, the encoder output channel dimensions are 200, 300 and 300, respectively.

% KTH experiments
\subsection{KTH and Weizmann action datasets}

\label{sec:kth}

\begin{figure*}[h!]
\vspace{-.7cm}
\centering
\includegraphics[width=0.49\linewidth] {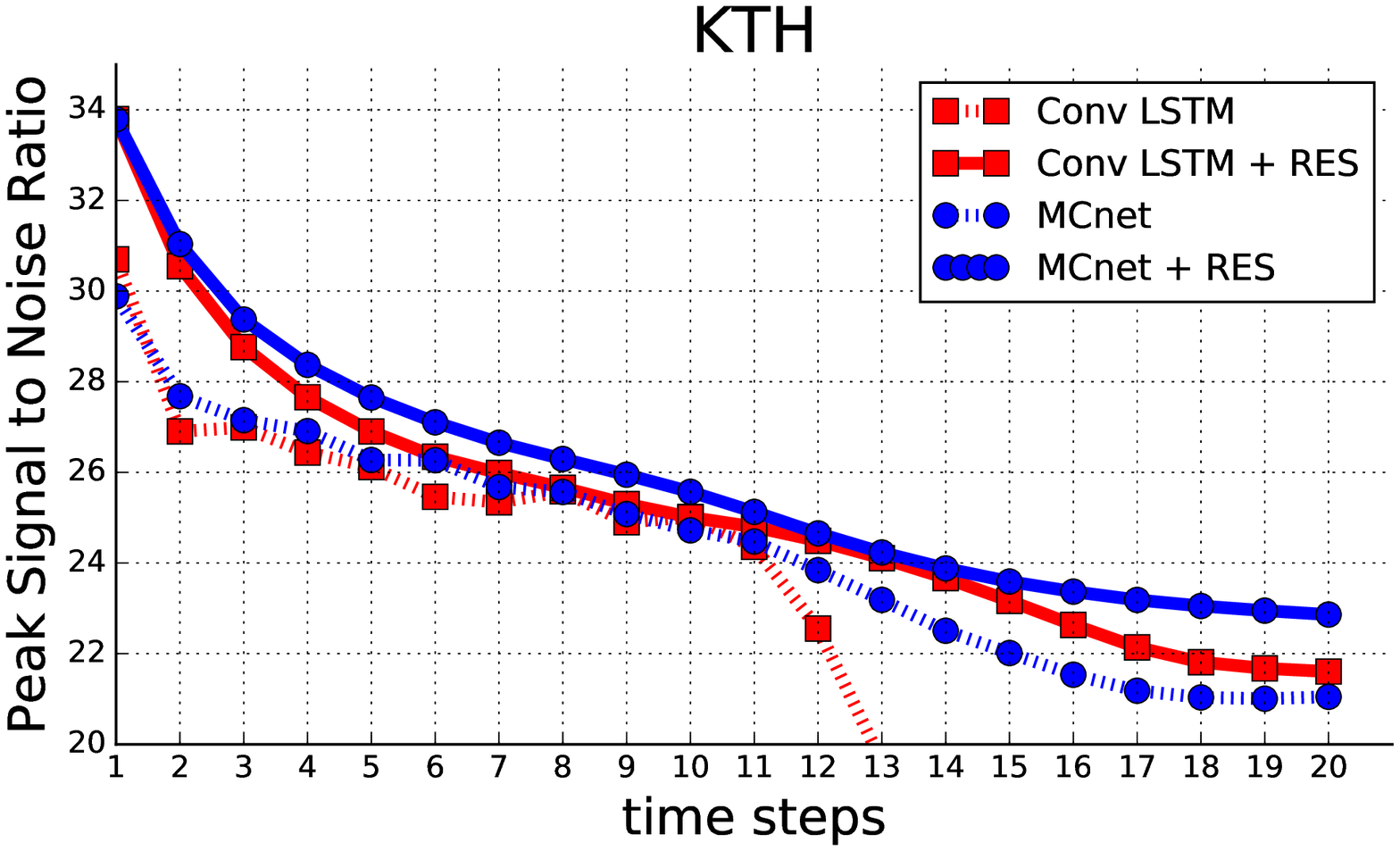} \hspace{.1cm}
\includegraphics[width=0.49\linewidth] {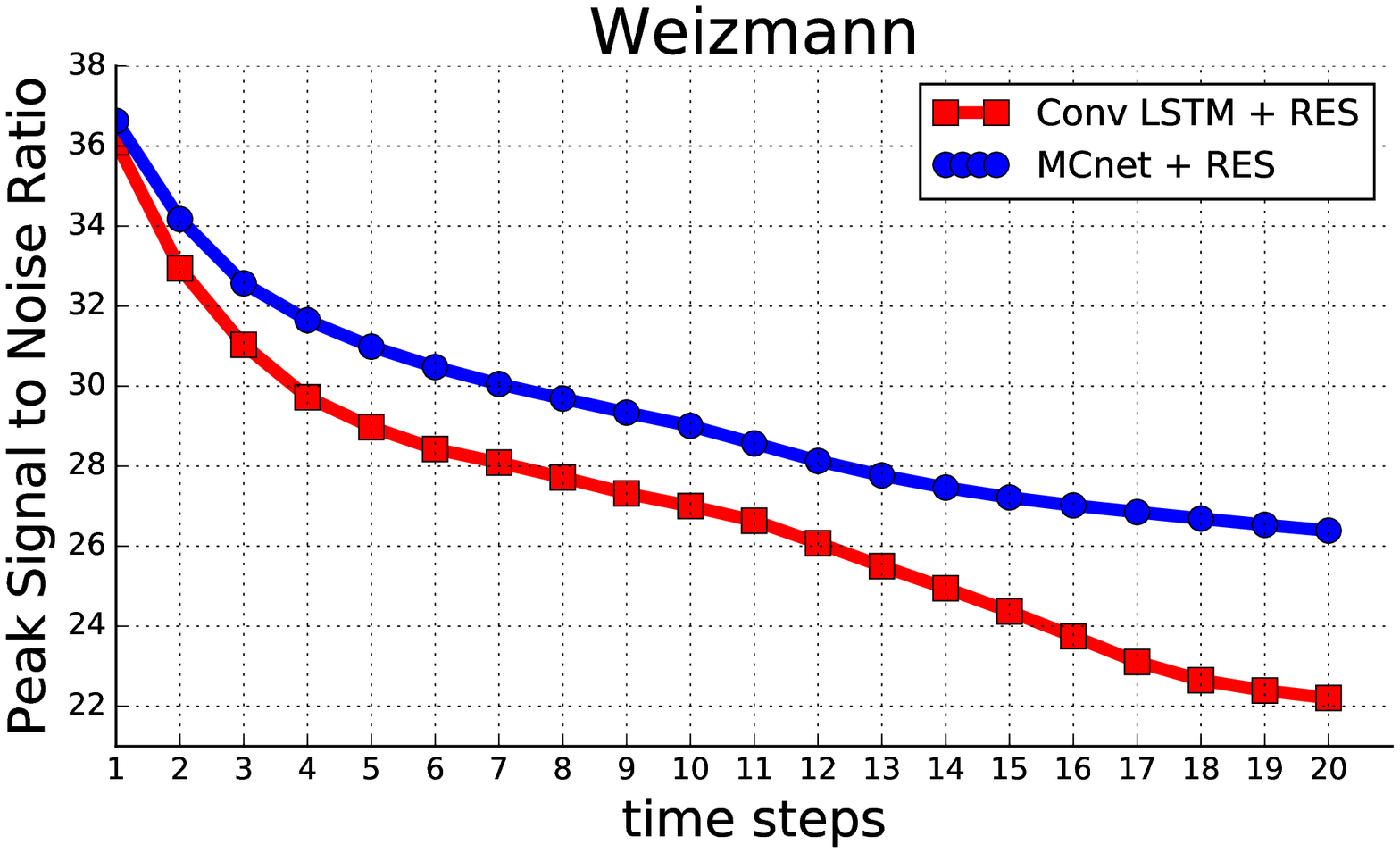} \hspace{0.1cm} \\
\includegraphics[width=0.49\linewidth] {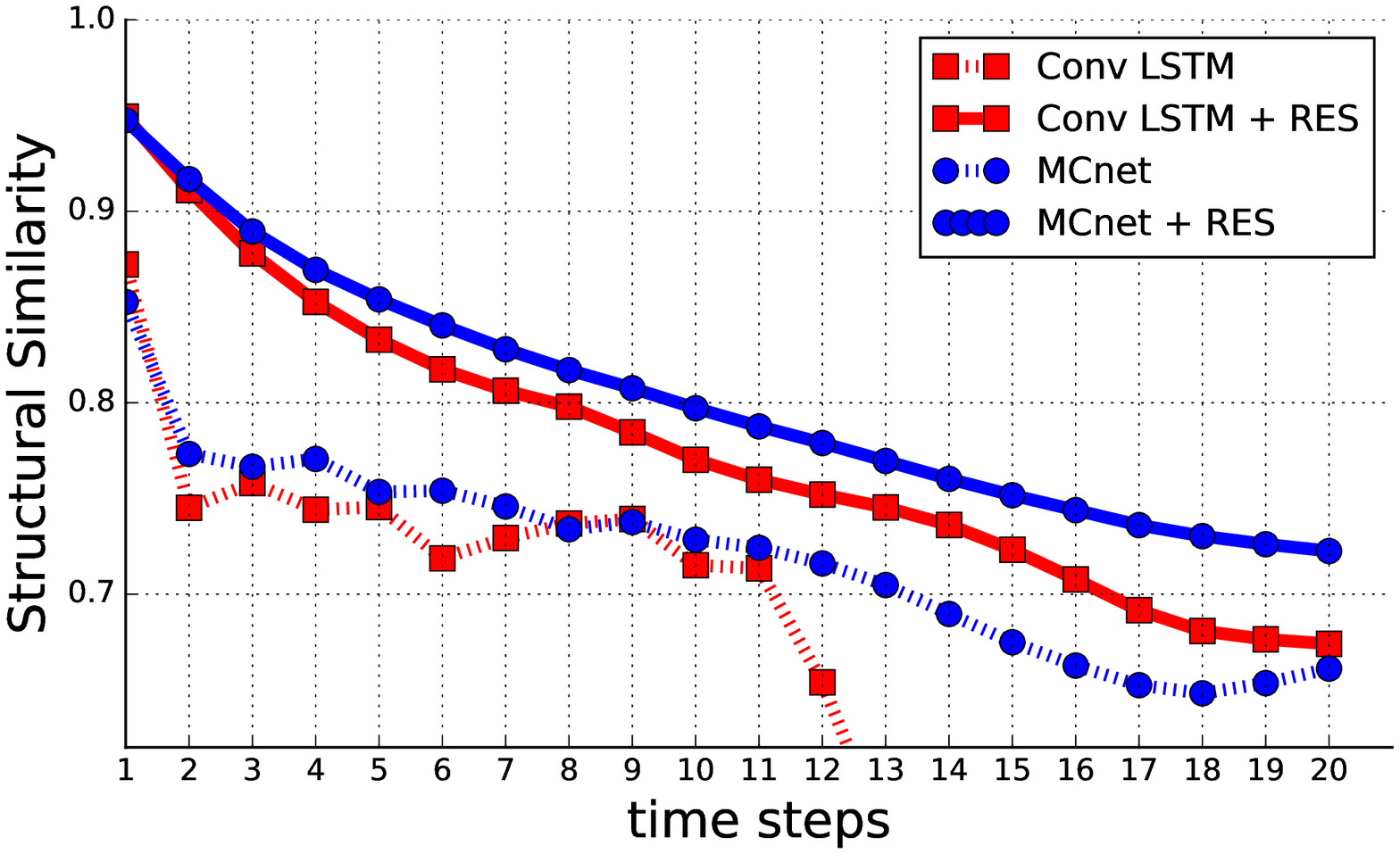} \hspace{.1cm}
\includegraphics[width=0.49\linewidth] {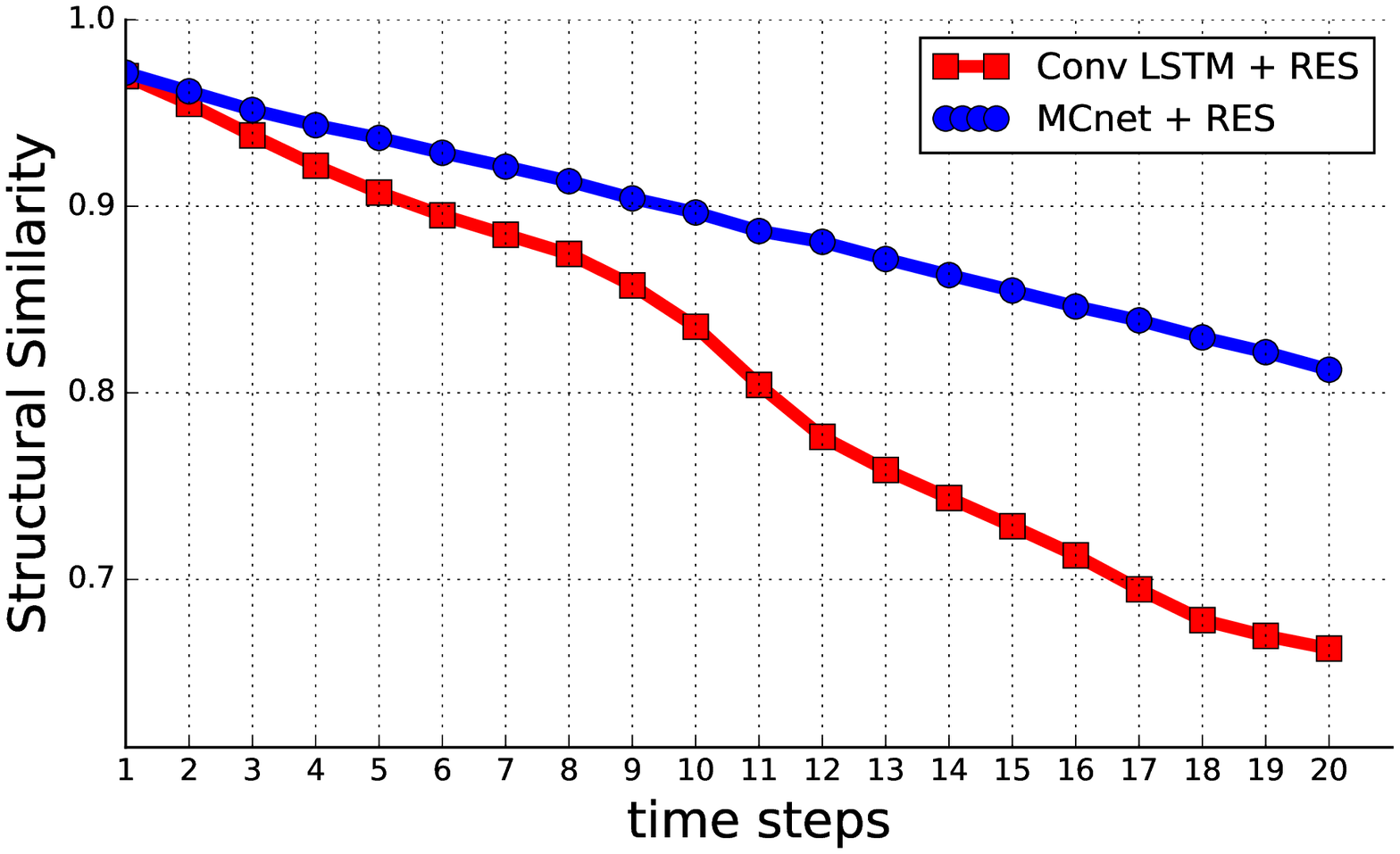} \hspace{0.1cm}
\vspace{-.2cm}
\caption{
Quantitative comparison between MCnet and ConvLSTM baseline with and without multi-scale residual connections (indicated by "+ RES"). Given 10 input frames, the models predict 20 frames recursively, one by one. Left column: evaluation on KTH dataset~\citep{Kth}. Right colum: evaluation on Weizmann~\citep{ActionsAsSpaceTimeShapes_pami07} dataset.}
\label{fig:kth_quantitative}
\vspace{-.7cm}
\end{figure*}

\begin{figure}[h!]
    \vspace{-.4cm}
    \hspace*{-.7cm}
    \centering
    \begin{subfigure}{0.04\linewidth}
        \raggedleft
        \rotatebox{90}{
        \hspace{-.4cm}
        \parbox{2cm}{\centering G.T.} \hspace{-.3cm} \parbox{2cm}{\centering ConvLSTM} \hspace{-.3cm} \parbox{2cm}{\centering MCnet}
        }
    \end{subfigure}
    \begin{subfigure}{0.13\linewidth}
        \caption*{t=12}
        \vspace{-7pt}
	    \includegraphics[width=1\linewidth]{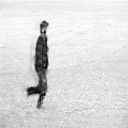} 
   		\includegraphics[width=1\linewidth]{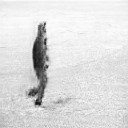}
   		\includegraphics[width=1\linewidth]{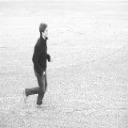} 
	\end{subfigure} 
    \begin{subfigure}{0.13\linewidth}
        \caption*{t=15}
        \vspace{-7pt}
	    \includegraphics[width=1\linewidth]{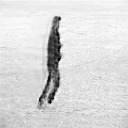} 
   		\includegraphics[width=1\linewidth]{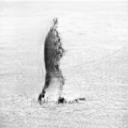}
   		\includegraphics[width=1\linewidth]{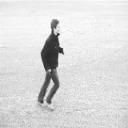}
	\end{subfigure} 
    \begin{subfigure}{0.13\linewidth}
        \caption*{t=18}
        \vspace{-7pt}
	    \includegraphics[width=1\linewidth]{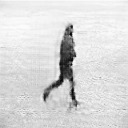} 
   		\includegraphics[width=1\linewidth]{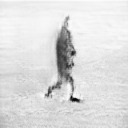}
   		\includegraphics[width=1\linewidth]{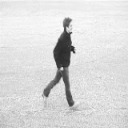}
	\end{subfigure} 
    \begin{subfigure}{0.13\linewidth}
        \vspace{20pt}
        \caption*{t=21}
        \vspace{-7pt}
	    \includegraphics[width=1\linewidth]{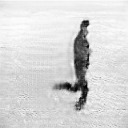} 
   		\includegraphics[width=1\linewidth]{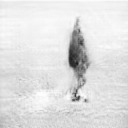}
   		\includegraphics[width=1\linewidth]{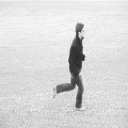}
   		\caption*{Jogging}
	\end{subfigure}
	\begin{subfigure}{0.13\linewidth}
        \caption*{t=24}
        \vspace{-7pt}
	    \includegraphics[width=1\linewidth]{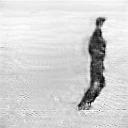} 
   		\includegraphics[width=1\linewidth]{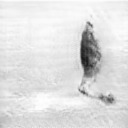}
   		\includegraphics[width=1\linewidth]{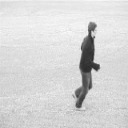}
	\end{subfigure}
	\begin{subfigure}{0.13\linewidth}
        \caption*{t=27}
        \vspace{-7pt}
	    \includegraphics[width=1\linewidth]{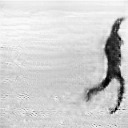} 
   		\includegraphics[width=1\linewidth]{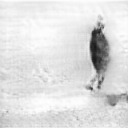}
   		\includegraphics[width=1\linewidth]{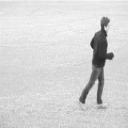}
	\end{subfigure}
	\begin{subfigure}{0.13\linewidth}
        \caption*{t=30}
        \vspace{-7pt}
	    \includegraphics[width=1\linewidth]{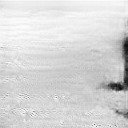} 
   		\includegraphics[width=1\linewidth]{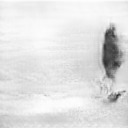}
   		\includegraphics[width=1\linewidth]{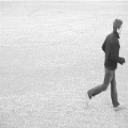}
	\end{subfigure}
    \vspace{.1cm}
    \hspace*{-.7cm}
    \centering
    \\
    \vspace{-.4cm}
    \begin{subfigure}{0.04\linewidth}
        \raggedleft
        \rotatebox{90}{
        \hspace{.2cm}
        \parbox{2cm}{\centering G.T.} \hspace{-.3cm} \parbox{2cm}{\centering ConvLSTM} \hspace{-.3cm} \parbox{2cm}{\centering MCnet}
        }
    \end{subfigure}
    \begin{subfigure}{0.13\linewidth}
        \vspace{-7pt}
	    \includegraphics[width=1\linewidth]{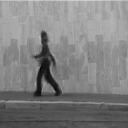} 
   		\includegraphics[width=1\linewidth]{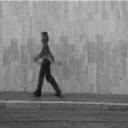}
   		\includegraphics[width=1\linewidth]{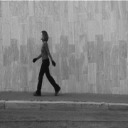} 
	\end{subfigure} 
    \begin{subfigure}{0.13\linewidth}
        \vspace{-7pt}
	    \includegraphics[width=1\linewidth]{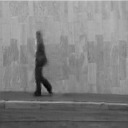} 
   		\includegraphics[width=1\linewidth]{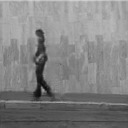}
   		\includegraphics[width=1\linewidth]{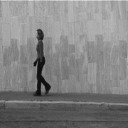}
	\end{subfigure} 
    \begin{subfigure}{0.13\linewidth}
        \vspace{-7pt}
	    \includegraphics[width=1\linewidth]{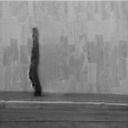} 
   		\includegraphics[width=1\linewidth]{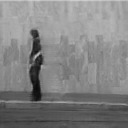}
   		\includegraphics[width=1\linewidth]{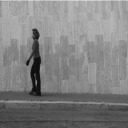}
	\end{subfigure} 
    \begin{subfigure}{0.13\linewidth}
        \vspace{9pt}
	    \includegraphics[width=1\linewidth]{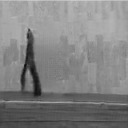} 
   		\includegraphics[width=1\linewidth]{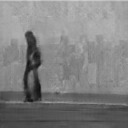}
   		\includegraphics[width=1\linewidth]{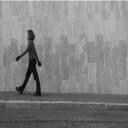}
   		\caption*{Walking}
	\end{subfigure}
	\begin{subfigure}{0.13\linewidth}
        \vspace{-7pt}
	    \includegraphics[width=1\linewidth]{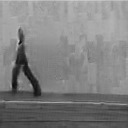} 
   		\includegraphics[width=1\linewidth]{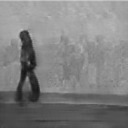}
   		\includegraphics[width=1\linewidth]{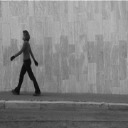}
	\end{subfigure}
	\begin{subfigure}{0.13\linewidth}
        \vspace{-7pt}
	    \includegraphics[width=1\linewidth]{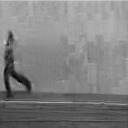} 
   		\includegraphics[width=1\linewidth]{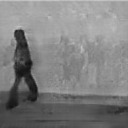}
   		\includegraphics[width=1\linewidth]{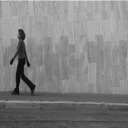}
	\end{subfigure}
	\begin{subfigure}{0.13\linewidth}
        \vspace{-7pt}
	    \includegraphics[width=1\linewidth]{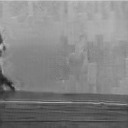} 
   		\includegraphics[width=1\linewidth]{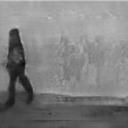}
   		\includegraphics[width=1\linewidth]{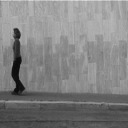}
	\end{subfigure}

	\vspace{-5pt}
    \caption{Qualitative comparison between our MCNet model and ConvLSTM. We display predictions starting from the $12^{\text{th}}$ frame, in every $3$ timesteps. The first $3$ rows correspond to KTH dataset for the action of jogging and the last $3$ rows correspond to Weizmann dataset for the action of walking.}
\label{fig:kth_qualitative}
\vspace{-.6cm}
\end{figure}

\paragraph{Experimental settings.}
The KTH human action dataset~\citep{Kth} contains 6 categories of periodic motions on a simple background: running, jogging, walking, boxing, hand-clapping and hand-waiving.
We use person 1-16 for training and 17-25 for testing, and also resize frames to 128x128 pixels.
We train our network and baseline by observing 10 frames and predicting 10 frames into the future on the KTH dataset.
We set $\beta=0.02$ for training.
We also select the walking, running, one-hand waving, and two-hands waving sequences from the Weizmann action dataset~\citep{ActionsAsSpaceTimeShapes_pami07} for testing the networks' generalizability.

For all the experiments, we test the networks on predicting 20 time steps into the future.
As for evaluation, we use the same SSIM and PSNR metrics as in~\citet{Mathieu15}.
The evaluation on KTH was performed on sub-clips within each video in the testset.
We sample sub-clips every 3 frames for running and jogging, and sample sub-clips every 20 frames (skipping the frames we have already predicted) for walking, boxing, hand-clapping, and hand-waving.
Sub-clips for running, jogging, and walking were manually trimmed to ensure humans are always present in the frames.
The evaluation on Weizmann was performed on all sub-clips in the selected sequences.

\paragraph{Results.}
Figure~\ref{fig:kth_quantitative} summarizes the quantitative comparisons among our MCnet, ConvLSTM baseline and their residual variations.
In the KTH test set, our network outperforms the ConvLSTM baseline by a small margin.
However, when we test the residual versions of MCnet and ConvLSTM on the dataset~\citep{ActionsAsSpaceTimeShapes_pami07} with similar motions, we can see that our network can generalize well to the unseen contents by showing clear improvements, especially in long-term prediction.
One reason for this result is that the test and training partitions of the KTH dataset have simple and similar image contents so that ConvLSTM can memorize the average background and human appearance to make reasonable predictions.
However, when tested on unseen data, ConvLSTM has to internally take care of both scene dynamics and image contents in a mingled representation, which gives it a hard time for generalization.
In contrast, the reason our network outperforms the ConvLSTM baseline on unseen data is that our network focuses on identifying general motion features and applying them to a learned content representation.

Figure \ref{fig:kth_qualitative} presents qualitative results of multi-step prediction by our network and ConvLSTM.
As expected, prediction results by our full architecture preserves human shapes more accurately than the baseline.
It is worth noticing that our network produces very sharp prediction over long-term time steps; it shows that MCnet is able to capture periodic motion cycles, which reduces the uncertainty of future prediction significantly. 
More qualitative comparisons are shown in the supplementary material and the \href{https://goo.gl/nG8ve1}{\color{blue} project website}. %~\url{https://sites.google.com/a/umich.edu/rubenevillegas/iclr2017}.
%More qualitative comparisons are shown in Appendix~\ref{supp:kth}.

\subsection{UCF-101 dataset} 
\label{sec:ucf}

\paragraph{Experimental settings.}
This section presents results on the challenging real-world videos in the UCF-101~\citep{Ucf} dataset.
Having collected from YouTube, the dataset contains 101 realistic human actions taken in a wild and exhibits various challenges, such as background clutter, occlusion, and complicated motion.
We employed the same network architecture as in the KTH dataset, but resized frames to 240x320 pixels, and trained the network to observe 4 frames and predict a single frame.
We set $\beta=0.001$ for training.
We also trained our convolutional LSTM baseline in the same way.
Following the same protocol as \citet{Mathieu15} for data pre-processing and evaluation metrics on full images, all networks were trained on Sports-1M \citep{sports1m} dataset and tested on UCF-101 unless otherwise stated.\footnote{We use the code and model released by \citet{Mathieu15} at \url{https://github.com/coupriec/VideoPredictionICLR2016}}

\begin{figure*}[h!]
\centering
\includegraphics[width=0.49\linewidth] {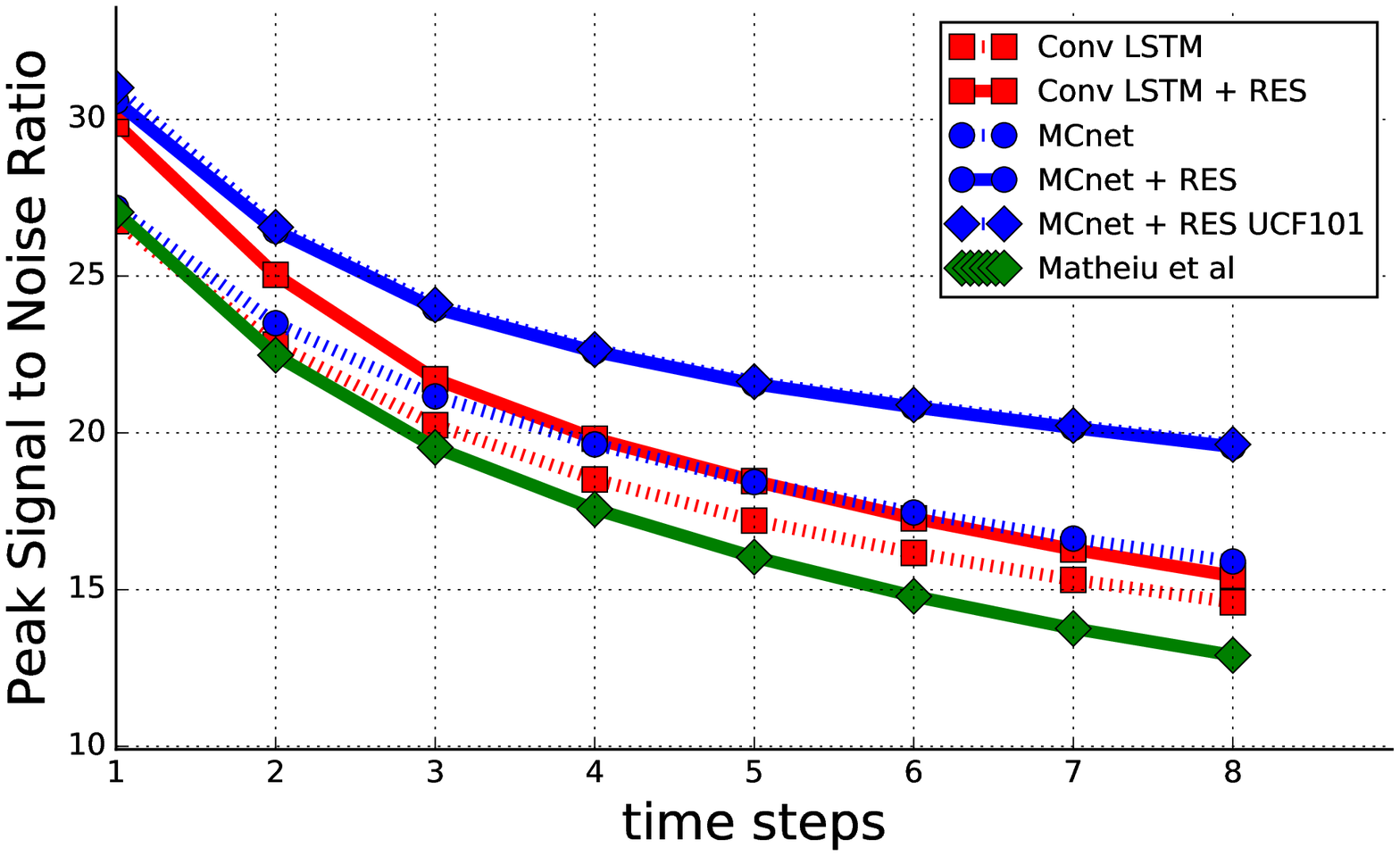} \hspace{-0.1cm}
\includegraphics[width=0.49\linewidth] {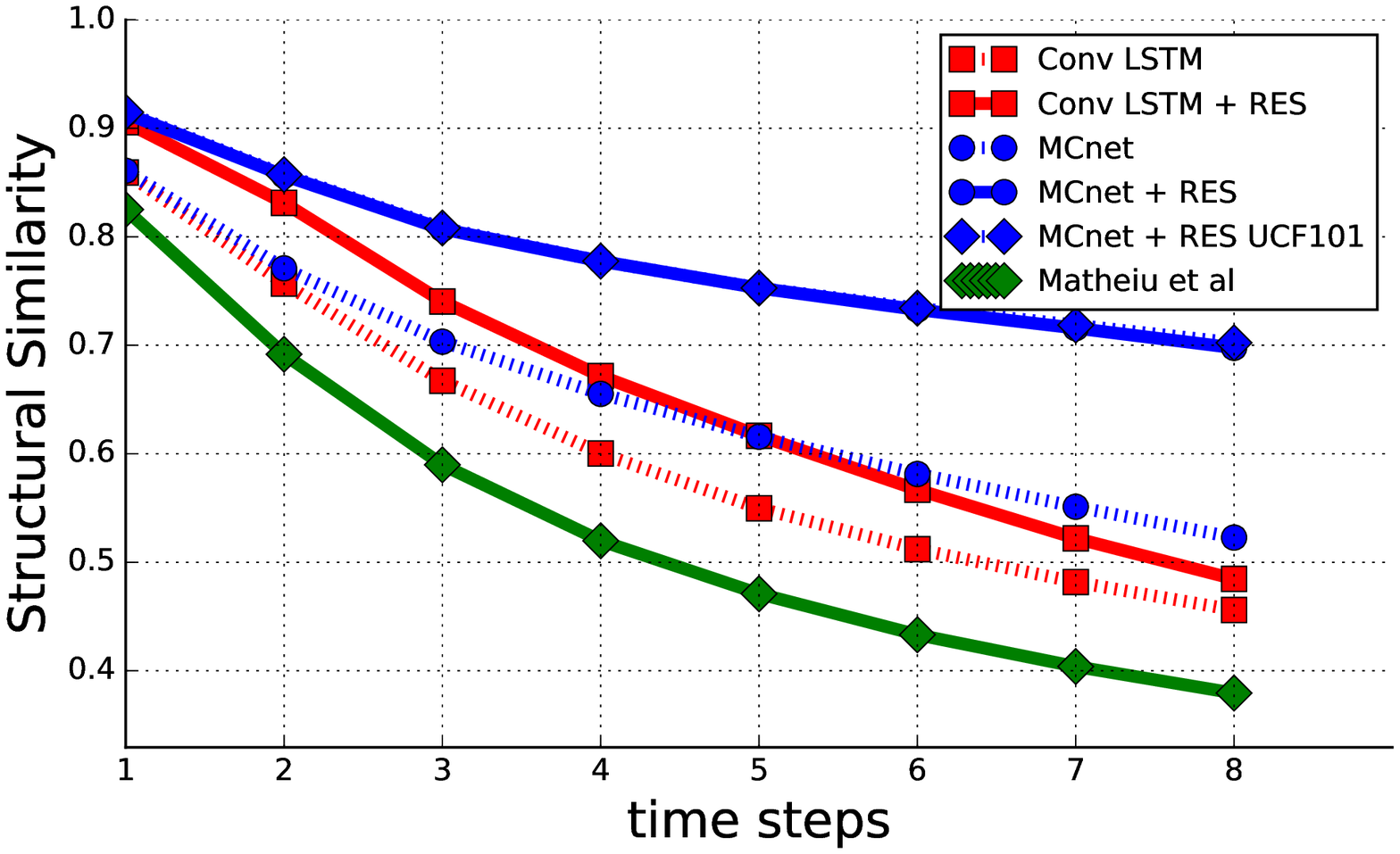} \hspace{0.1cm}
\caption{
Quantitative comparison between our model, convolutional LSTM~\cite{convlstm}, and \cite{Mathieu15}. Given 4 input frames, the models predict 8 frames recursively, one by one.}
\vspace{-.6cm}
\label{fig:ucf101_quantitative}
\end{figure*}

\begin{figure}[htb!]
    \hspace*{-1.cm}
    \centering
    \begin{subfigure}{0.04\linewidth}
        \raggedleft
        \rotatebox{90}{
        \parbox{0.1cm}{\rotatebox{-90}{t=11}} \hspace{2.1cm} \parbox{0.1cm}{\rotatebox{-90}{t=9}} \hspace{2.3cm} \parbox{0.1cm}{\rotatebox{-90}{t=7}} \hspace{2.1cm} \parbox{0.1cm}{\rotatebox{-90}{t=5}} \hspace{.6cm}
        }
    \end{subfigure}
    \begin{subfigure}{0.23\linewidth}
        \caption*{G.T.}
        \vspace{-7pt}
	    \includegraphics[width=1\linewidth]{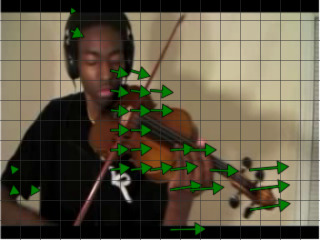}
   		\includegraphics[width=1\linewidth]{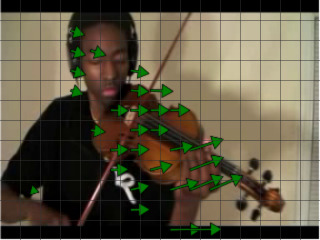}
   		\includegraphics[width=1\linewidth]{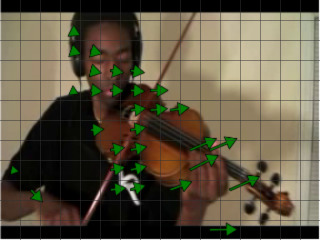}
   		\includegraphics[width=1\linewidth]{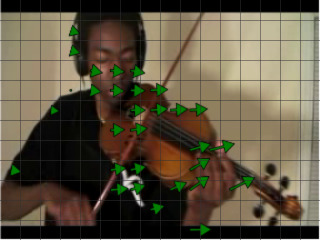} 
	\end{subfigure}
    \begin{subfigure}{0.23\linewidth}
        \caption*{MCnet}
        \vspace{-7pt}
	    \includegraphics[width=1\linewidth]{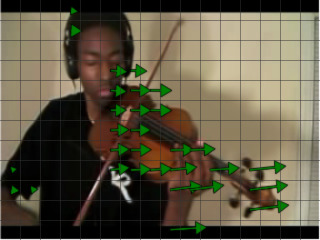} 
   		\includegraphics[width=1\linewidth]{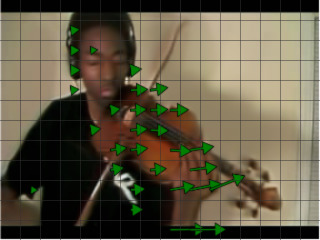}
   		\includegraphics[width=1\linewidth]{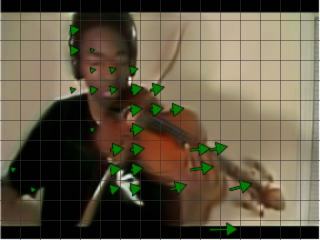}
   		\includegraphics[width=1\linewidth]{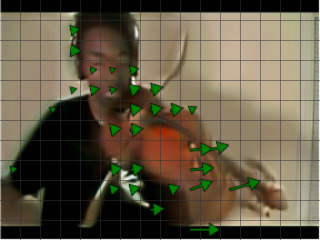}
	\end{subfigure}
	\begin{subfigure}{0.23\linewidth}
        \caption*{ConvLSTM}
        \vspace{-7pt}
	    \includegraphics[width=1\linewidth]{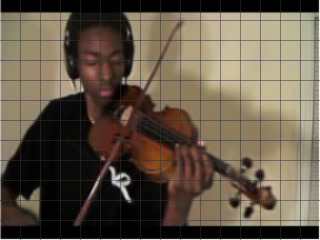} 
   		\includegraphics[width=1\linewidth]{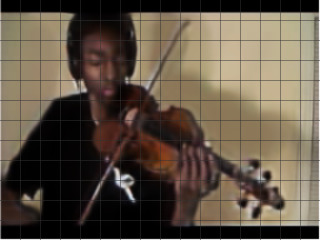}
   		\includegraphics[width=1\linewidth]{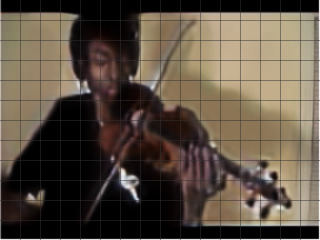}
   		\includegraphics[width=1\linewidth]{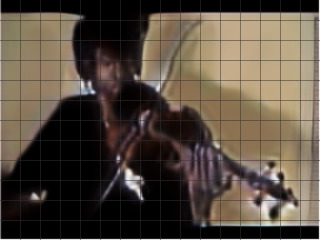}
	\end{subfigure}
	\begin{subfigure}{0.23\linewidth}
        \caption*{\cite{Mathieu15}}
        \vspace{-7pt}
	    \includegraphics[width=1\linewidth]{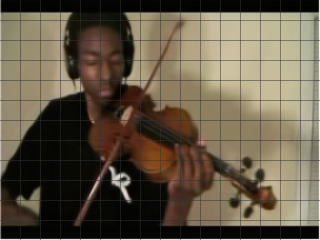} 
   		\includegraphics[width=1\linewidth]{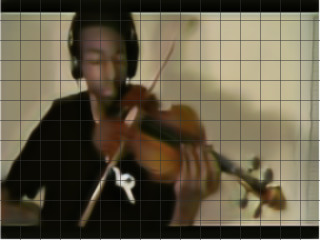}
   		\includegraphics[width=1\linewidth]{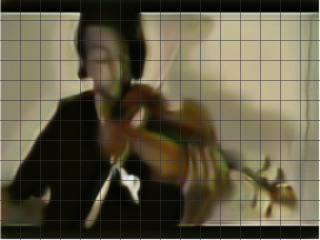}
   		\includegraphics[width=1\linewidth]{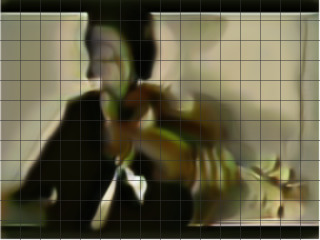}
	\end{subfigure}
    \vspace{.1cm}
    \hspace*{-1cm} \\
    \centering
    \hspace*{-.7cm}
% ----------------------------------------------------------------------------------------------------------------

    \hspace*{-1.1cm}
    \begin{subfigure}{0.04\linewidth}
        \raggedleft
        \rotatebox{90}{
        \parbox{0.1cm}{\rotatebox{-90}{t=11}} \hspace{2.1cm} \parbox{0.1cm}{\rotatebox{-90}{t=9}} \hspace{2.3cm} \parbox{0.1cm}{\rotatebox{-90}{t=7}} \hspace{2.1cm} \parbox{0.1cm}{\rotatebox{-90}{t=5}} \hspace{.6cm}
        }
    \end{subfigure}
    \begin{subfigure}{0.23\linewidth}
        \caption*{}
        \vspace{-7pt}
	    \includegraphics[width=1\linewidth]{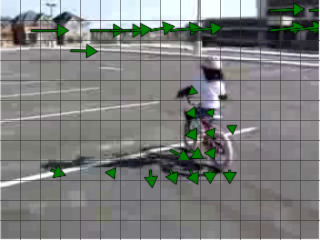}
   		\includegraphics[width=1\linewidth]{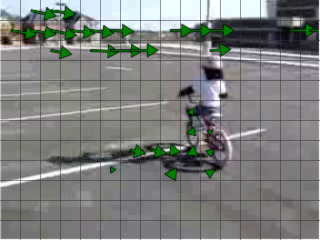}
   		\includegraphics[width=1\linewidth]{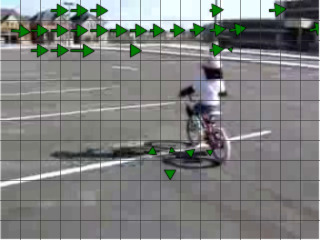}
   		\includegraphics[width=1\linewidth]{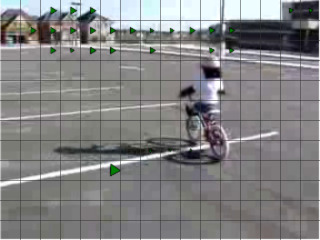} 
	\end{subfigure} 
    \begin{subfigure}{0.23\linewidth}
        \caption*{}
        \vspace{-7pt}
	    \includegraphics[width=1\linewidth]{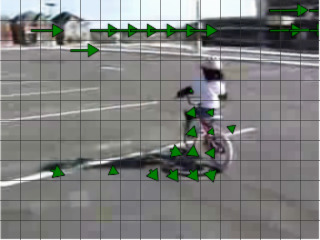} 
   		\includegraphics[width=1\linewidth]{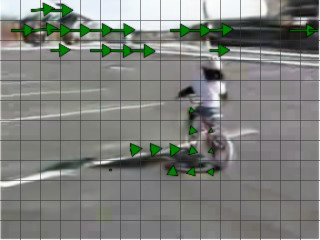}
   		\includegraphics[width=1\linewidth]{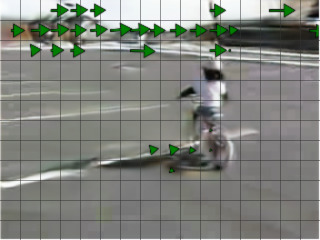}
   		\includegraphics[width=1\linewidth]{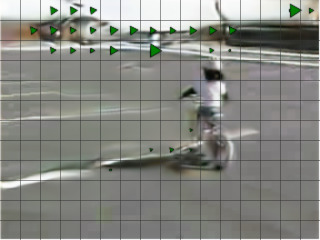}
	\end{subfigure}
	\begin{subfigure}{0.23\linewidth}
        \caption*{}
        \vspace{-7pt}
	    \includegraphics[width=1\linewidth]{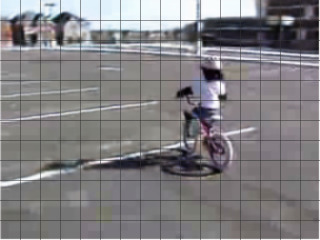} 
   		\includegraphics[width=1\linewidth]{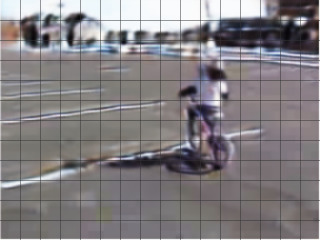}
   		\includegraphics[width=1\linewidth]{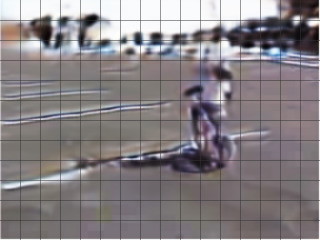}
   		\includegraphics[width=1\linewidth]{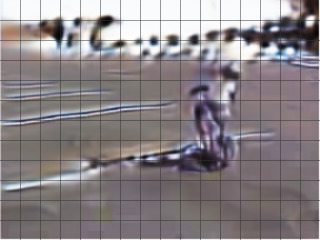}
	\end{subfigure}
	\begin{subfigure}{0.23\linewidth}
        \caption*{}
        \vspace{-7pt}
	    \includegraphics[width=1\linewidth]{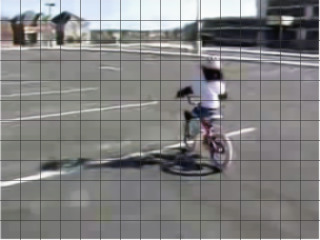} 
   		\includegraphics[width=1\linewidth]{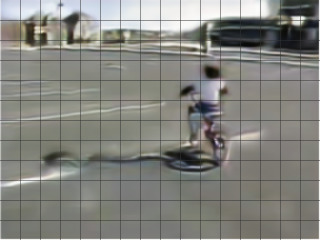}
   		\includegraphics[width=1\linewidth]{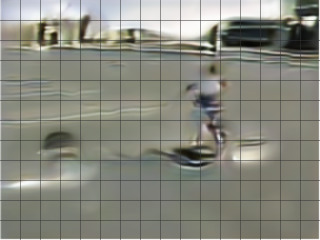}
   		\includegraphics[width=1\linewidth]{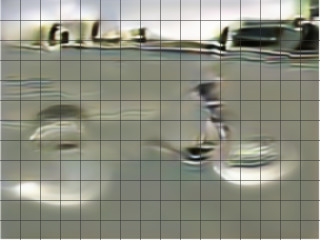}
	\end{subfigure}
	\hspace*{-1.1cm}

	\vspace{-5pt}
    \caption{Qualitative comparisons among MCnet and ConvLSTM and \cite{Mathieu15}. We display predicted frames (in every other frame) starting from the $5^{\text{th}}$ frame. The green arrows denote the top-30 closest optical flow vectors within image patches between MCnet and ground-truth. More clear motion prediction can be seen in the \href{https://goo.gl/nG8ve1}{\color{blue} project website}.}
\label{fig:ucf101_qualitative}
\vspace{-0cm}
\end{figure}

\paragraph{Results.}
Figure \ref{fig:ucf101_quantitative} shows the quantitative comparisons between our network trained for single-step-prediction and \citet{Mathieu15}.
We can clearly see the advantage of our network over the baseline. The separation of motion and contents in two encoder pathways allows our network to identify key motion and content features, which are then fed into the decoder to yield predictions of higher quality compared to the baseline.\footnote{We were not able to get the model fine-tuned on UCF-101 from the authors so it is not included in Figure \ref{fig:ucf101_quantitative}}
In other words, our network only moves what shows motion in the past, and leaves the rest untouched.
We also trained a residual version of MCnet on UCF-101, indicated by ``MCnet + RES UCF101", to compare how well our model generalizes when trained and tested on the same or different dataset(s). 
To our surprise, when tested with UCF-101, the MCnet trained on Sports-1M (MCnet + RES) roughly matches the performance of the MCnet trained on UCF-101 (MCnet + RES UCF101), which suggests that our model learns effective representations which can generalize to new datasets.
Figure \ref{fig:ucf101_qualitative} presents qualitative comparisons between frames generated by our network and \citet{Mathieu15}.
Since the ConvLSTM and \citet{Mathieu15} lack explicit motion and content modules, they lose sense of the dynamics in the video and therefore the contents become distorted quickly.
More qualitative comparisons are shown in the supplementary material and the \href{https://goo.gl/nG8ve1}{\color{blue} project website}.
%More qualitative comparisons are shown in the supplementary material and the project website: \url{https://sites.google.com/a/umich.edu/rubenevillegas/iclr2017}.

\vspace{-.03cm}

\newpage

\vspace{-1cm}

\newpage

% Conclusion ------------------------------------------------------------------------------------------------------------------------------------------------------------
\cutsectionup
\section{Conclusion}
\label{sec:conclusion}
\commenttext{Tentative conclusion: may need to edit further}
\cutsectiondown

\label{sec:conclusion}
We proposed a motion-content network for pixel-level prediction of future frames in natural video sequences. 
The proposed model employs two separate encoding pathways, and learns to decompose motion and content without explicit constraints or separate training.  
Experimental results suggest that separate modeling of motion and content improves the quality of the pixel-level future prediction, 
% by identifying the \textit{content} of the last observed frame and converting them into the next frame based on the features from the \textit{motion} encoder.
% Our network architecture can serve as a base model to more complicated networks that can benefit from the separation of motion and contents in videos.
% Exploiting benefits of multiple encoders, future work will involve imposing stochasticity in our architecture to allow the modeling of the inherit uncertainty in natural videos as we move further into the future.
% Adding stochasticity to our network can potentially improve the prediction the average motion present as we move further into the future caused by the many possible futures.
% Adding stochasticity to our network, especially for motion encoder, would facilitate modeling of more complicated dynamics in a video, and enable sharper prediction by preventing the model from averaging all possible futures; it would be one promising future research direction of this paper. 
% encoding pathways for contents and motion can be extended to model different characteristics of both factors (e.g. modeling with different level of stochasticity).
%
%Overall, our model achieves state-of-the-art performance in predicting future frames in challenging real-world video datasets.
and our model overall achieves state-of-the-art performance in predicting future frames in challenging real-world video datasets.

\cutsectionup
\section{Acknowledgements}
\cutsectiondown

This work was supported in part by ONR N00014-13-1-0762, NSF CAREER IIS-1453651, gifts from the Bosch Research and Technology Center, and Sloan Research Fellowship.
We also thank NVIDIA for donating K40c and TITAN X GPUs.
We thank Ye Liu, Junhyuk Oh, Xinchen Yan, Lajanugen Logeswaran, Yuting Zhang, Sungryull Sohn, Kibok Lee, Rui Zhang, and other collaborators for helpful discussions.
R. Villegas was partly supported by the Rackham Merit Fellowship.

\bibliography{iclr2017_conference}
\bibliographystyle{abbrvnat}

\clearpage
\newpage

\section{Appendix}
\begin{appendix}

%\section{KTH} \label{supp:kth}
\begin{figure}[!htb]
    \hspace*{-.7cm}
    \centering
    \begin{subfigure}{0.04\linewidth}
        \raggedleft
        \rotatebox{90}{
        \hspace{-.4cm}
        \parbox{2cm}{\centering G.T.} \hspace{-.3cm} \parbox{2cm}{\centering ConvLSTM} \hspace{-.3cm} \parbox{2cm}{\centering MCnet}
        }
    \end{subfigure}
    \begin{subfigure}{0.13\linewidth}
        \caption*{t=12}
        \vspace{-7pt}
	    \includegraphics[width=1\linewidth]{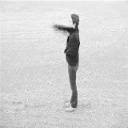} 
   		\includegraphics[width=1\linewidth]{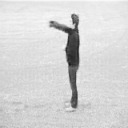}
   		\includegraphics[width=1\linewidth]{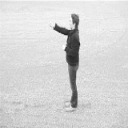} 
	\end{subfigure} 
    \begin{subfigure}{0.13\linewidth}
        \caption*{t=15}
        \vspace{-7pt}
	    \includegraphics[width=1\linewidth]{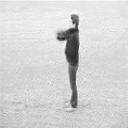} 
   		\includegraphics[width=1\linewidth]{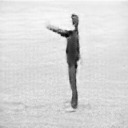}
   		\includegraphics[width=1\linewidth]{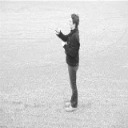}
	\end{subfigure} 
    \begin{subfigure}{0.13\linewidth}
        \caption*{t=18}
        \vspace{-7pt}
	    \includegraphics[width=1\linewidth]{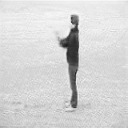} 
   		\includegraphics[width=1\linewidth]{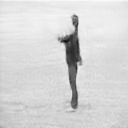}
   		\includegraphics[width=1\linewidth]{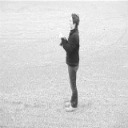}
	\end{subfigure} 
    \begin{subfigure}{0.13\linewidth}
        \vspace{20pt}
        \caption*{t=21}
        \vspace{-7pt}
	    \includegraphics[width=1\linewidth]{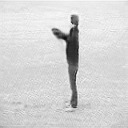} 
   		\includegraphics[width=1\linewidth]{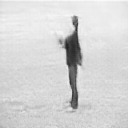}
   		\includegraphics[width=1\linewidth]{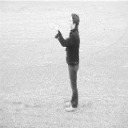}
   		\caption*{Boxing}
	\end{subfigure}
	\begin{subfigure}{0.13\linewidth}
        \caption*{t=24}
        \vspace{-7pt}
	    \includegraphics[width=1\linewidth]{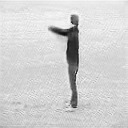} 
   		\includegraphics[width=1\linewidth]{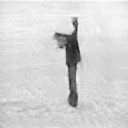}
   		\includegraphics[width=1\linewidth]{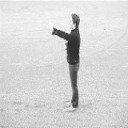}
	\end{subfigure}
	\begin{subfigure}{0.13\linewidth}
        \caption*{t=27}
        \vspace{-7pt}
	    \includegraphics[width=1\linewidth]{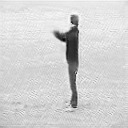} 
   		\includegraphics[width=1\linewidth]{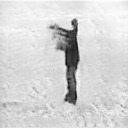}
   		\includegraphics[width=1\linewidth]{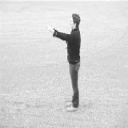}
	\end{subfigure}
	\begin{subfigure}{0.13\linewidth}
        \caption*{t=30}
        \vspace{-7pt}
	    \includegraphics[width=1\linewidth]{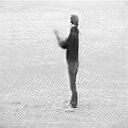} 
   		\includegraphics[width=1\linewidth]{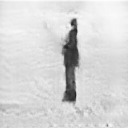}
   		\includegraphics[width=1\linewidth]{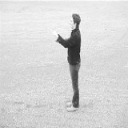}
	\end{subfigure}
    \vspace{.1cm}
    \hspace*{-.7cm} \\
    \centering
    \hspace*{-.7cm}
%-----------------------------------------------------------------------------------------------
    \begin{subfigure}{0.04\linewidth}
        \raggedleft
        \rotatebox{90}{
        \hspace{.1cm}
        \parbox{2cm}{\centering G.T.} \hspace{-.3cm} \parbox{2cm}{\centering ConvLSTM} \hspace{-.3cm} \parbox{2cm}{\centering MCnet}
        }
    \end{subfigure}
    \begin{subfigure}{0.13\linewidth}
        \vspace{-7pt}
	    \includegraphics[width=1\linewidth]{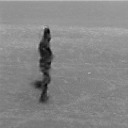} 
   		\includegraphics[width=1\linewidth]{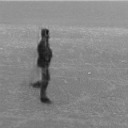}
   		\includegraphics[width=1\linewidth]{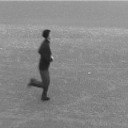} 
	\end{subfigure} 
    \begin{subfigure}{0.13\linewidth}
        \vspace{-7pt}
	    \includegraphics[width=1\linewidth]{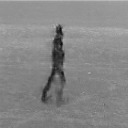} 
   		\includegraphics[width=1\linewidth]{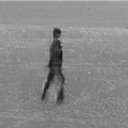}
   		\includegraphics[width=1\linewidth]{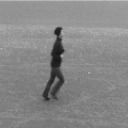}
	\end{subfigure} 
    \begin{subfigure}{0.13\linewidth}
        \vspace{-7pt}
	    \includegraphics[width=1\linewidth]{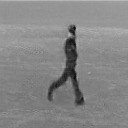} 
   		\includegraphics[width=1\linewidth]{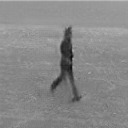}
   		\includegraphics[width=1\linewidth]{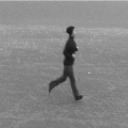}
	\end{subfigure} 
    \begin{subfigure}{0.13\linewidth}
        \vspace{9pt}
	    \includegraphics[width=1\linewidth]{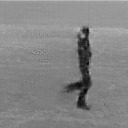} 
   		\includegraphics[width=1\linewidth]{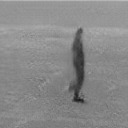}
   		\includegraphics[width=1\linewidth]{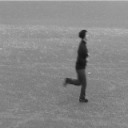}
   		\caption*{Running}
	\end{subfigure}
	\begin{subfigure}{0.13\linewidth}
        \vspace{-7pt}
	    \includegraphics[width=1\linewidth]{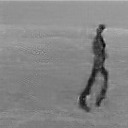} 
   		\includegraphics[width=1\linewidth]{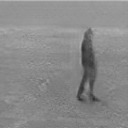}
   		\includegraphics[width=1\linewidth]{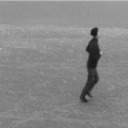}
	\end{subfigure}
	\begin{subfigure}{0.13\linewidth}
        \vspace{-7pt}
	    \includegraphics[width=1\linewidth]{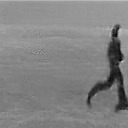} 
   		\includegraphics[width=1\linewidth]{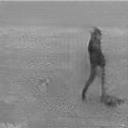}
   		\includegraphics[width=1\linewidth]{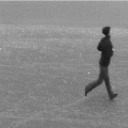}
	\end{subfigure}
	\begin{subfigure}{0.13\linewidth}
        \vspace{-7pt}
	    \includegraphics[width=1\linewidth]{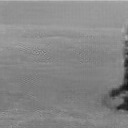} 
   		\includegraphics[width=1\linewidth]{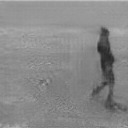}
   		\includegraphics[width=1\linewidth]{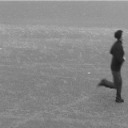}
	\end{subfigure}
    \vspace{.1cm}
    \hspace*{-.7cm} \\
    \centering
    \hspace*{-.7cm}
%---------------------------------------------------------------------------------------------
    \begin{subfigure}{0.04\linewidth}
        \raggedleft
        \rotatebox{90}{
        \hspace{.1cm}
        \parbox{2cm}{\centering G.T.} \hspace{-.3cm} \parbox{2cm}{\centering ConvLSTM} \hspace{-.3cm} \parbox{2cm}{\centering MCnet}
        }
    \end{subfigure}
    \begin{subfigure}{0.13\linewidth}
        \vspace{-7pt}
	    \includegraphics[width=1\linewidth]{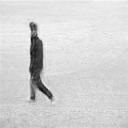} 
   		\includegraphics[width=1\linewidth]{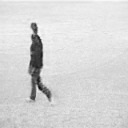}
   		\includegraphics[width=1\linewidth]{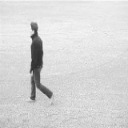} 
	\end{subfigure} 
    \begin{subfigure}{0.13\linewidth}
        \vspace{-7pt}
	    \includegraphics[width=1\linewidth]{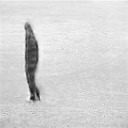} 
   		\includegraphics[width=1\linewidth]{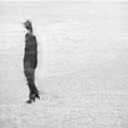}
   		\includegraphics[width=1\linewidth]{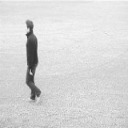}
	\end{subfigure} 
    \begin{subfigure}{0.13\linewidth}
        \vspace{-7pt}
	    \includegraphics[width=1\linewidth]{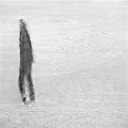} 
   		\includegraphics[width=1\linewidth]{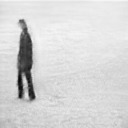}
   		\includegraphics[width=1\linewidth]{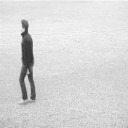}
	\end{subfigure} 
    \begin{subfigure}{0.13\linewidth}
        \vspace{9pt}
	    \includegraphics[width=1\linewidth]{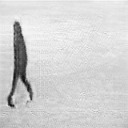} 
   		\includegraphics[width=1\linewidth]{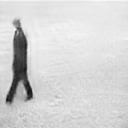}
   		\includegraphics[width=1\linewidth]{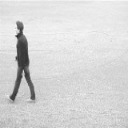}
   		\caption*{Walking}
	\end{subfigure}
	\begin{subfigure}{0.13\linewidth}
        \vspace{-7pt}
	    \includegraphics[width=1\linewidth]{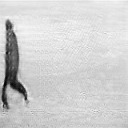} 
   		\includegraphics[width=1\linewidth]{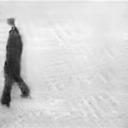}
   		\includegraphics[width=1\linewidth]{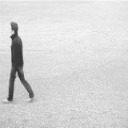}
	\end{subfigure}
	\begin{subfigure}{0.13\linewidth}
        \vspace{-7pt}
	    \includegraphics[width=1\linewidth]{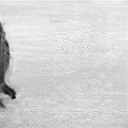} 
   		\includegraphics[width=1\linewidth]{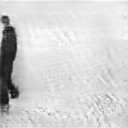}
   		\includegraphics[width=1\linewidth]{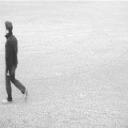}
	\end{subfigure}
	\begin{subfigure}{0.13\linewidth}
        \vspace{-7pt}
	    \includegraphics[width=1\linewidth]{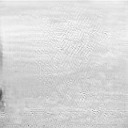} 
   		\includegraphics[width=1\linewidth]{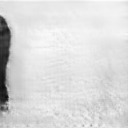}
   		\includegraphics[width=1\linewidth]{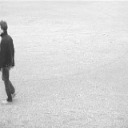}
	\end{subfigure}
    \vspace{.1cm}
    \hspace*{-.7cm} \\
    \centering

%-----------------------------------------------------------------------------------------------
    \caption{Qualitative comparisons on KTH testset. We display predictions starting from the $12^{\text{th}}$ frame, for every $3$ timesteps. More clear motion prediction can be seen in the \href{https://goo.gl/nG8ve1}{\color{blue} project website}.}
\label{fig:kth_qualitative2}
\vspace{-.5cm}
\end{figure}

\begin{figure}[t!]
    \vspace{-10cm}
    \hspace*{-.7cm}
    \centering
%---------------------------------------------------------------------------------------------
    \begin{subfigure}{0.04\linewidth}
        \raggedleft
        \rotatebox{90}{
        \hspace{-.4cm}
        \parbox{2cm}{\centering G.T.} \hspace{-.3cm} \parbox{2cm}{\centering ConvLSTM} \hspace{-.3cm} \parbox{2cm}{\centering MCnet}
        }
    \end{subfigure}
    \begin{subfigure}{0.13\linewidth}
        \caption*{t=12}
        \vspace{-7pt}
	    \includegraphics[width=1\linewidth]{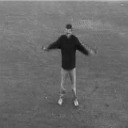} 
   		\includegraphics[width=1\linewidth]{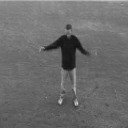}
   		\includegraphics[width=1\linewidth]{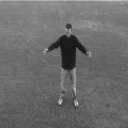} 
	\end{subfigure} 
    \begin{subfigure}{0.13\linewidth}
        \caption*{t=15}
        \vspace{-7pt}
	    \includegraphics[width=1\linewidth]{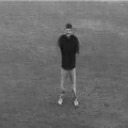} 
   		\includegraphics[width=1\linewidth]{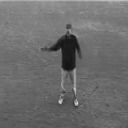}
   		\includegraphics[width=1\linewidth]{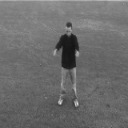}
	\end{subfigure} 
    \begin{subfigure}{0.13\linewidth}
        \caption*{t=18}
        \vspace{-7pt}
	    \includegraphics[width=1\linewidth]{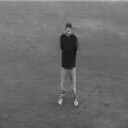} 
   		\includegraphics[width=1\linewidth]{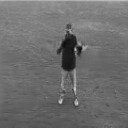}
   		\includegraphics[width=1\linewidth]{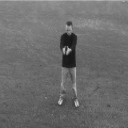}
	\end{subfigure} 
    \begin{subfigure}{0.13\linewidth}
        \vspace{21pt}
        \caption*{t=21}
        \vspace{-7pt}
	    \includegraphics[width=1\linewidth]{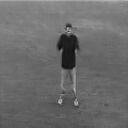} 
   		\includegraphics[width=1\linewidth]{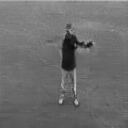}
   		\includegraphics[width=1\linewidth]{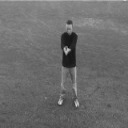}
   		\caption*{Handclapping}
	\end{subfigure}
	\begin{subfigure}{0.13\linewidth}
	    \caption*{t=24}
        \vspace{-7pt}
	    \includegraphics[width=1\linewidth]{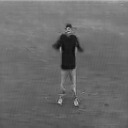} 
   		\includegraphics[width=1\linewidth]{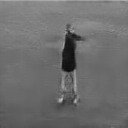}
   		\includegraphics[width=1\linewidth]{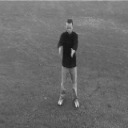}
	\end{subfigure}
	\begin{subfigure}{0.13\linewidth}
	    \caption*{t=27}
        \vspace{-7pt}
	    \includegraphics[width=1\linewidth]{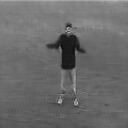} 
   		\includegraphics[width=1\linewidth]{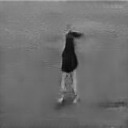}
   		\includegraphics[width=1\linewidth]{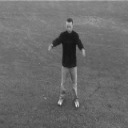}
	\end{subfigure}
	\begin{subfigure}{0.13\linewidth}
	    \caption*{t=30}
        \vspace{-7pt}
	    \includegraphics[width=1\linewidth]{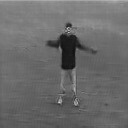} 
   		\includegraphics[width=1\linewidth]{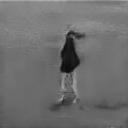}
   		\includegraphics[width=1\linewidth]{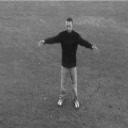}
	\end{subfigure}
    \vspace{.1cm}
    \hspace*{-.7cm} \\
    \centering
    \hspace*{-.7cm}
%-----------------------------------------------------------------------------------------------
    \begin{subfigure}{0.04\linewidth}
        \raggedleft
        \rotatebox{90}{
        \hspace{.1cm}
        \parbox{2cm}{\centering G.T.} \hspace{-.3cm} \parbox{2cm}{\centering ConvLSTM} \hspace{-.3cm} \parbox{2cm}{\centering MCnet}
        }
    \end{subfigure}
    \begin{subfigure}{0.13\linewidth}
        \vspace{-7pt}
	    \includegraphics[width=1\linewidth]{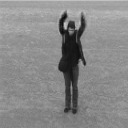} 
   		\includegraphics[width=1\linewidth]{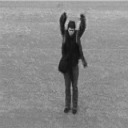}
   		\includegraphics[width=1\linewidth]{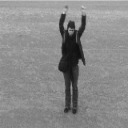} 
	\end{subfigure} 
    \begin{subfigure}{0.13\linewidth}
        \vspace{-7pt}
	    \includegraphics[width=1\linewidth]{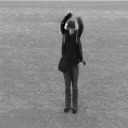} 
   		\includegraphics[width=1\linewidth]{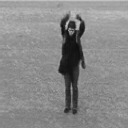}
   		\includegraphics[width=1\linewidth]{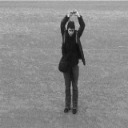}
	\end{subfigure} 
    \begin{subfigure}{0.13\linewidth}
        \vspace{-7pt}
	    \includegraphics[width=1\linewidth]{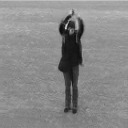} 
   		\includegraphics[width=1\linewidth]{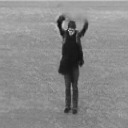}
   		\includegraphics[width=1\linewidth]{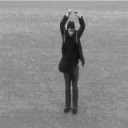}
	\end{subfigure} 
    \begin{subfigure}{0.13\linewidth}
        \vspace{9pt}
	    \includegraphics[width=1\linewidth]{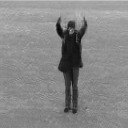} 
   		\includegraphics[width=1\linewidth]{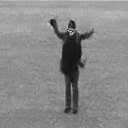}
   		\includegraphics[width=1\linewidth]{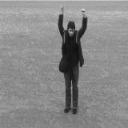}
   		\caption*{Handwaving}
	\end{subfigure}
	\begin{subfigure}{0.13\linewidth}
        \vspace{-7pt}
	    \includegraphics[width=1\linewidth]{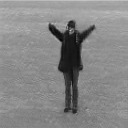} 
   		\includegraphics[width=1\linewidth]{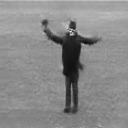}
   		\includegraphics[width=1\linewidth]{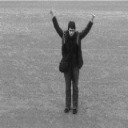}
	\end{subfigure}
	\begin{subfigure}{0.13\linewidth}
        \vspace{-7pt}
	    \includegraphics[width=1\linewidth]{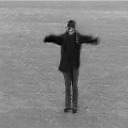} 
   		\includegraphics[width=1\linewidth]{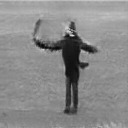}
   		\includegraphics[width=1\linewidth]{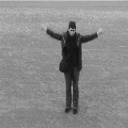}
	\end{subfigure}
	\begin{subfigure}{0.13\linewidth}
        \vspace{-7pt}
	    \includegraphics[width=1\linewidth]{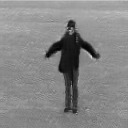} 
   		\includegraphics[width=1\linewidth]{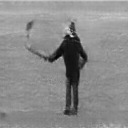}
   		\includegraphics[width=1\linewidth]{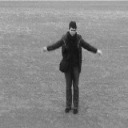}
	\end{subfigure}
    \vspace{.1cm}
    \hspace*{-.7cm} \\
    \centering

%-----------------------------------------------------------------------------------------------
    \caption{Qualitative comparisons on KTH testset. We display predictions starting from the $12^{\text{th}}$ frame, for every $3$ timesteps. More clear motion prediction can be seen in the \href{https://goo.gl/nG8ve1}{\color{blue} project website}.}
\label{fig:kth_qualitative3}
\vspace{-.5cm}
\end{figure}

\clearpage
\newpage

\begin{figure}[!hbt]
    \hspace*{-.7cm}
    \hspace*{-1.1cm}
    \centering
    \begin{subfigure}{0.04\linewidth}
        \raggedleft
        \rotatebox{90}{
        \parbox{0.1cm}{\rotatebox{-90}{t=11}} \hspace{2.1cm} \parbox{0.1cm}{\rotatebox{-90}{t=9}} \hspace{2.3cm} \parbox{0.1cm}{\rotatebox{-90}{t=7}} \hspace{2.1cm} \parbox{0.1cm}{\rotatebox{-90}{t=5}} \hspace{.6cm}
        }
    \end{subfigure}
    \begin{subfigure}{0.23\linewidth}
        \caption*{G.T.}
        \vspace{-7pt}
	    \includegraphics[width=1\linewidth]{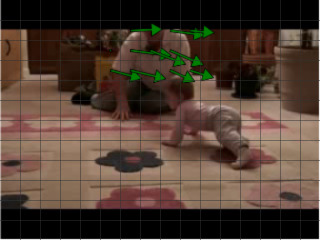}
   		\includegraphics[width=1\linewidth]{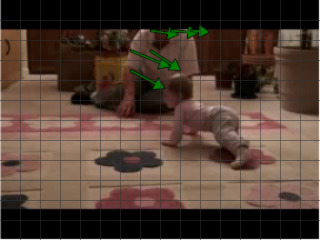}
   		\includegraphics[width=1\linewidth]{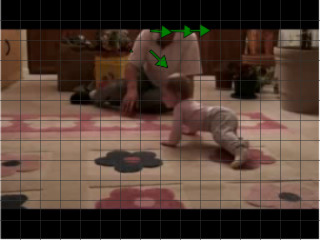}
   		\includegraphics[width=1\linewidth]{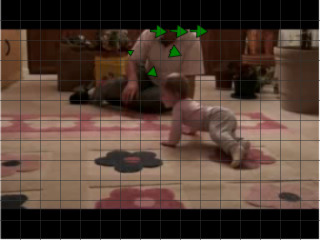} 
	\end{subfigure}
    \begin{subfigure}{0.23\linewidth}
        \caption*{MCnet}
        \vspace{-7pt}
	    \includegraphics[width=1\linewidth]{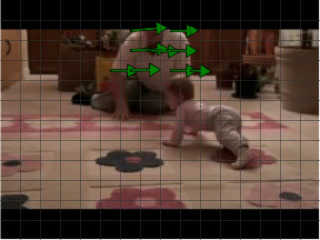} 
   		\includegraphics[width=1\linewidth]{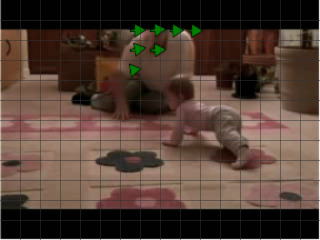}
   		\includegraphics[width=1\linewidth]{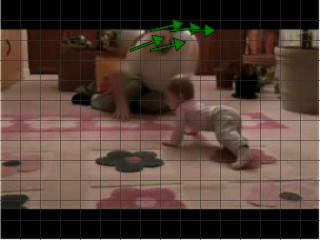}
   		\includegraphics[width=1\linewidth]{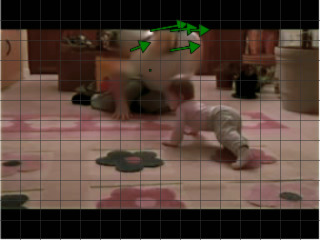}
	\end{subfigure}
	\begin{subfigure}{0.23\linewidth}
        \caption*{ConvLSTM}
        \vspace{-7pt}
	    \includegraphics[width=1\linewidth]{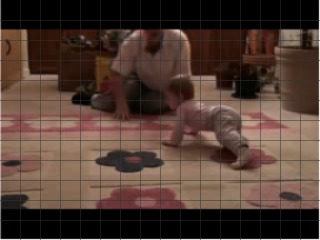} 
   		\includegraphics[width=1\linewidth]{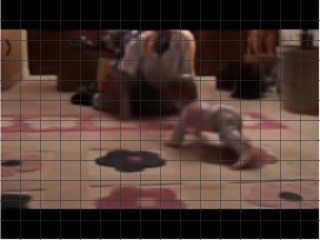}
   		\includegraphics[width=1\linewidth]{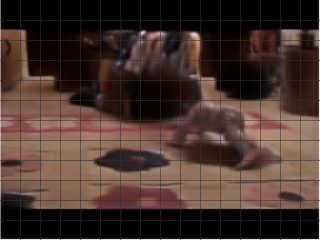}
   		\includegraphics[width=1\linewidth]{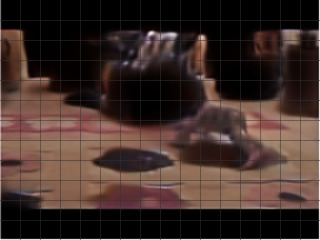}
	\end{subfigure}
	\begin{subfigure}{0.23\linewidth}
        \caption*{\cite{Mathieu15}}
        \vspace{-7pt}
	    \includegraphics[width=1\linewidth]{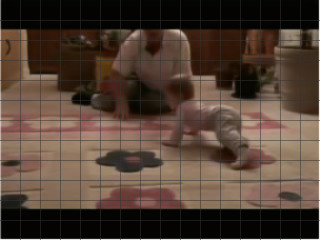} 
   		\includegraphics[width=1\linewidth]{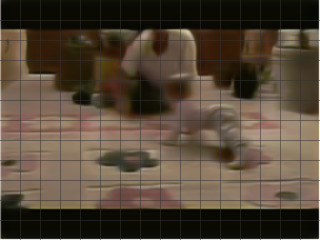}
   		\includegraphics[width=1\linewidth]{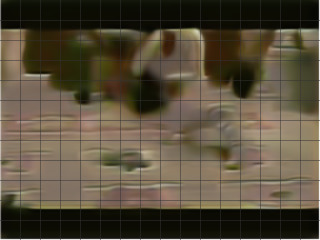}
   		\includegraphics[width=1\linewidth]{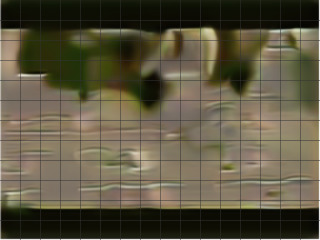}
	\end{subfigure}
    \vspace{.1cm}
    \hspace*{-1cm} \\
    \centering
    \hspace*{-.7cm}
% ----------------------------------------------------------------------------------------------------------------

    \hspace*{-.8cm}
    \begin{subfigure}{0.04\linewidth}
        \raggedleft
        \rotatebox{90}{
        \parbox{0.1cm}{\rotatebox{-90}{t=11}} \hspace{2.1cm} \parbox{0.1cm}{\rotatebox{-90}{t=9}} \hspace{2.3cm} \parbox{0.1cm}{\rotatebox{-90}{t=7}} \hspace{2.1cm} \parbox{0.1cm}{\rotatebox{-90}{t=5}} \hspace{.6cm}
        }
    \end{subfigure}
    \begin{subfigure}{0.23\linewidth}
        \caption*{}
        \vspace{-7pt}
	    \includegraphics[width=1\linewidth]{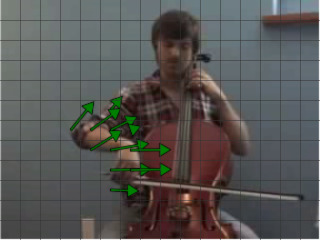}
   		\includegraphics[width=1\linewidth]{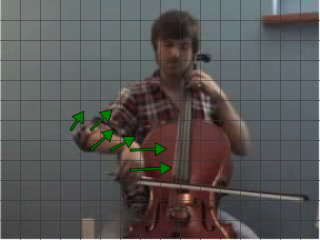}
   		\includegraphics[width=1\linewidth]{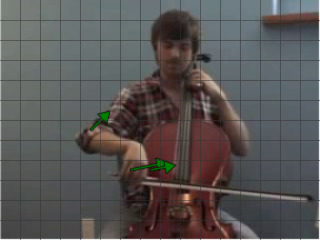}
   		\includegraphics[width=1\linewidth]{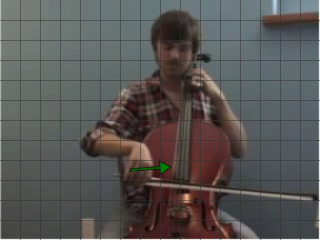} 
	\end{subfigure}
    \begin{subfigure}{0.23\linewidth}
        \caption*{}
        \vspace{-7pt}
	    \includegraphics[width=1\linewidth]{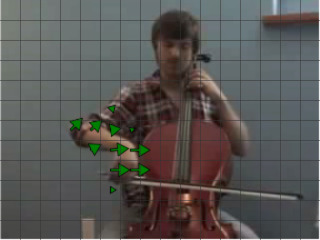} 
   		\includegraphics[width=1\linewidth]{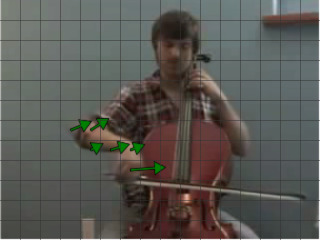}
   		\includegraphics[width=1\linewidth]{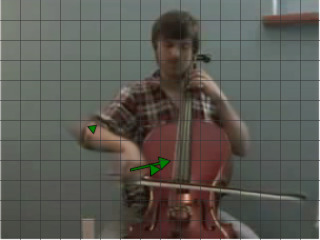}
   		\includegraphics[width=1\linewidth]{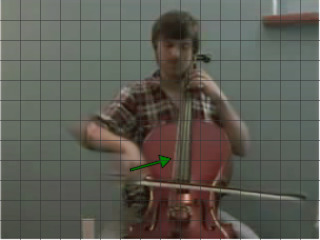}
	\end{subfigure}
	\begin{subfigure}{0.23\linewidth}
        \caption*{}
        \vspace{-7pt}
	    \includegraphics[width=1\linewidth]{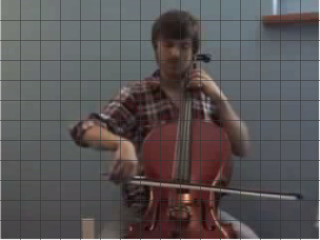} 
   		\includegraphics[width=1\linewidth]{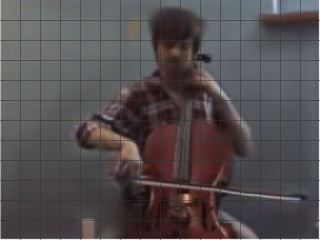}
   		\includegraphics[width=1\linewidth]{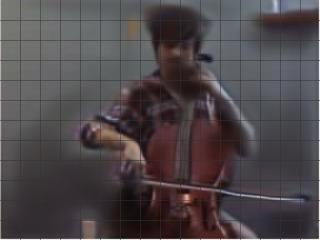}
   		\includegraphics[width=1\linewidth]{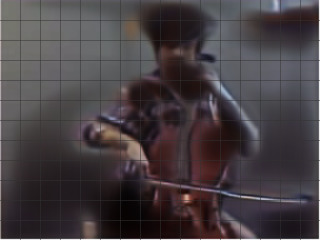}
	\end{subfigure}
	\begin{subfigure}{0.23\linewidth}
        \caption*{}
        \vspace{-7pt}
	    \includegraphics[width=1\linewidth]{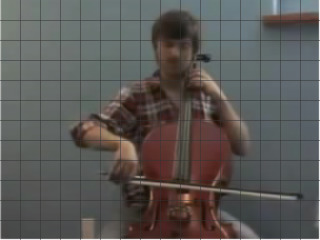} 
   		\includegraphics[width=1\linewidth]{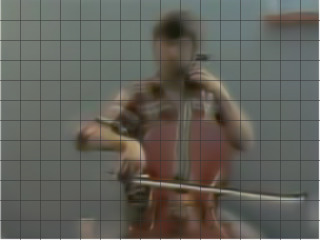}
   		\includegraphics[width=1\linewidth]{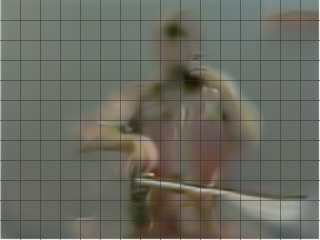}
   		\includegraphics[width=1\linewidth]{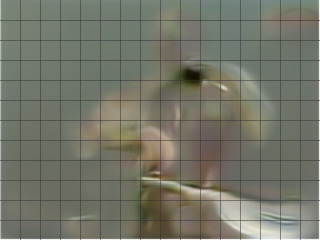}
	\end{subfigure}

	\vspace{-5pt}
    \caption{Qualitative comparisons on UCF-101. We display predictions (in every other frame) starting from the $5^{\text{th}}$ frame. The green arrows denote the top-30 closest optical flow vectors within image patches between MCnet and ground-truth. More clear motion prediction can be seen in the \href{https://goo.gl/nG8ve1}{\color{blue} project website}.} %~\url{https://sites.google.com/a/umich.edu/rubenevillegas/iclr2017}.
\label{fig:ucf101_qualitative2}
\vspace{-.5cm}
\end{figure}

\newpage
\section{Qualitative and quantitative comparison with considerable camera motion and analysis} \label{sec:extquant}
\begin{figure}[!hbt]
    \hspace*{-.7cm}
    \centering
    \begin{subfigure}{0.04\linewidth}
        \raggedleft
        \rotatebox{90}{
        \parbox{0.1cm}{\rotatebox{-90}{t=11}} \hspace{2.1cm} \parbox{0.1cm}{\rotatebox{-90}{t=9}} \hspace{2.3cm} \parbox{0.1cm}{\rotatebox{-90}{t=7}} \hspace{2.1cm} \parbox{0.1cm}{\rotatebox{-90}{t=5}} \hspace{.6cm}
        }
    \end{subfigure}
    \begin{subfigure}{0.23\linewidth}
        \caption*{G.T.}
        \vspace{-7pt}
	    \includegraphics[width=1\linewidth]{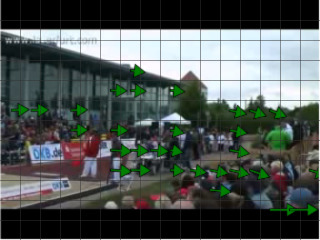}
   		\includegraphics[width=1\linewidth]{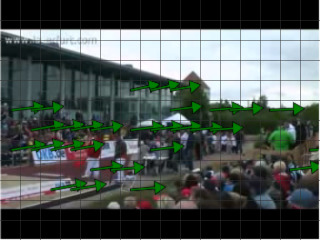}
   		\includegraphics[width=1\linewidth]{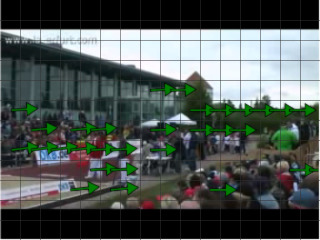}
   		\includegraphics[width=1\linewidth]{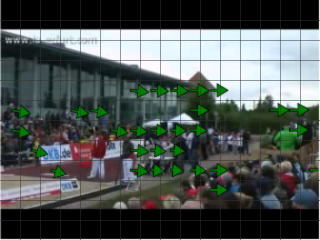} 
	\end{subfigure}  
    \begin{subfigure}{0.23\linewidth}
        \caption*{MCnet}
        \vspace{-7pt}
	    \includegraphics[width=1\linewidth]{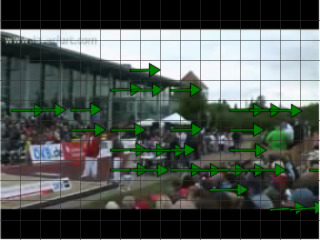} 
   		\includegraphics[width=1\linewidth]{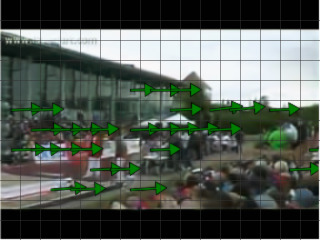}
   		\includegraphics[width=1\linewidth]{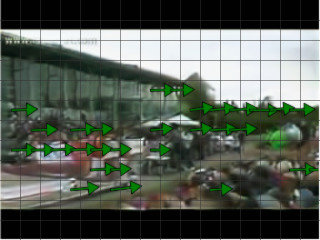}
   		\includegraphics[width=1\linewidth]{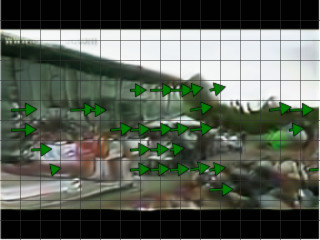}
	\end{subfigure}
	\begin{subfigure}{0.23\linewidth}
        \caption*{ConvLSTM}
        \vspace{-7pt}
	    \includegraphics[width=1\linewidth]{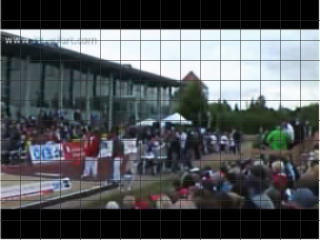} 
   		\includegraphics[width=1\linewidth]{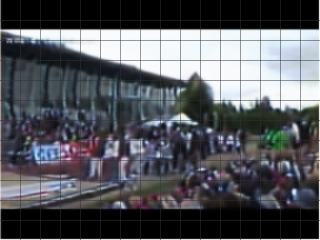}
   		\includegraphics[width=1\linewidth]{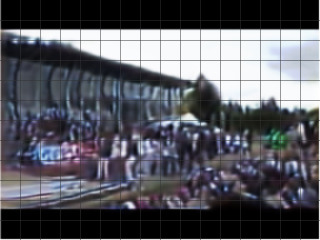}
   		\includegraphics[width=1\linewidth]{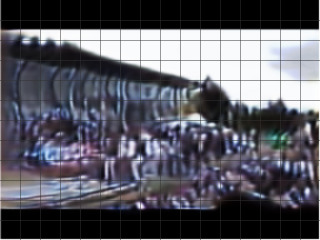}
	\end{subfigure}
	\begin{subfigure}{0.23\linewidth}
        \caption*{\cite{Mathieu15}}
        \vspace{-7pt}
	    \includegraphics[width=1\linewidth]{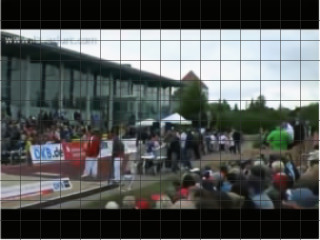} 
   		\includegraphics[width=1\linewidth]{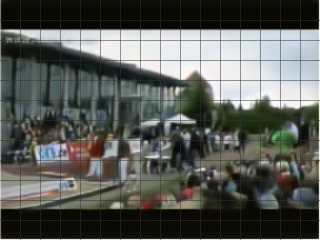}
   		\includegraphics[width=1\linewidth]{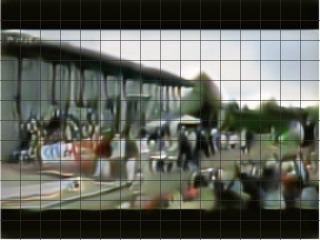}
   		\includegraphics[width=1\linewidth]{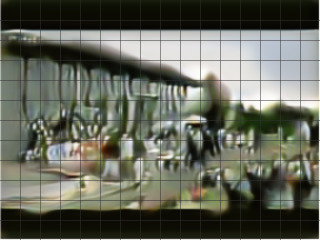}
	\end{subfigure}
	\vspace{-5pt}
    \caption{Qualitative comparisons on UCF-101. We display predictions (in every other frame) starting from the $5^{\text{th}}$ frame. The green arrows denote the top-30 closest optical flow vectors within image patches between MCnet and ground-truth. More clear motion prediction can be seen in the \href{https://goo.gl/nG8ve1}{\color{blue} project website}.} %~\url{https://sites.google.com/a/umich.edu/rubenevillegas/iclr2017}.
\label{fig:ucf101_qualitative3}
\vspace{-.5cm}
\end{figure}

\begin{figure}[!hbt]
    \hspace*{-1cm}
    \centering
    \begin{subfigure}{0.04\linewidth}
        \raggedleft
        \rotatebox{90}{
        \parbox{0.1cm}{\rotatebox{-90}{t=11}} \hspace{2.1cm} \parbox{0.1cm}{\rotatebox{-90}{t=9}} \hspace{2.3cm} \parbox{0.1cm}{\rotatebox{-90}{t=7}} \hspace{2.1cm} \parbox{0.1cm}{\rotatebox{-90}{t=5}} \hspace{.6cm}
        }
    \end{subfigure}
    \begin{subfigure}{0.23\linewidth}
        \caption*{G.T.}
        \vspace{-7pt}
	    \includegraphics[width=1\linewidth]{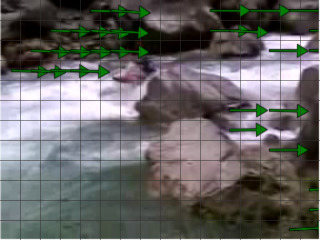}
   		\includegraphics[width=1\linewidth]{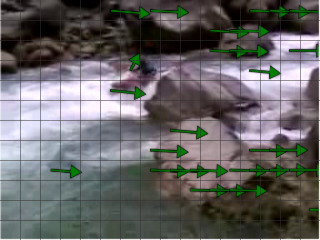}
   		\includegraphics[width=1\linewidth]{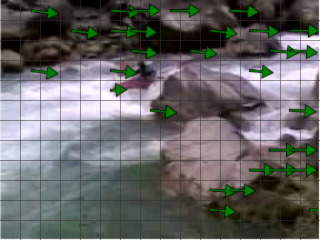}
   		\includegraphics[width=1\linewidth]{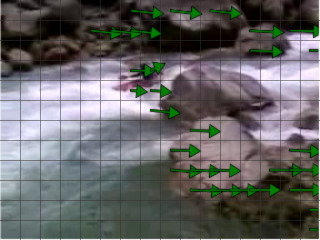} 
	\end{subfigure}
    \begin{subfigure}{0.23\linewidth}
        \caption*{MCnet}
        \vspace{-7pt}
	    \includegraphics[width=1\linewidth]{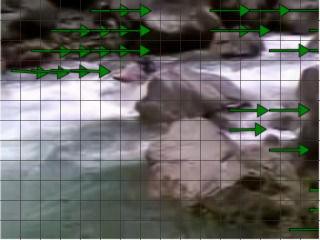} 
   		\includegraphics[width=1\linewidth]{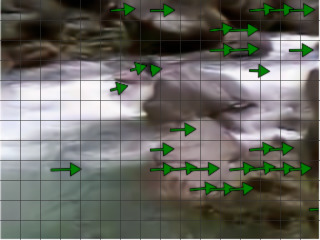}
   		\includegraphics[width=1\linewidth]{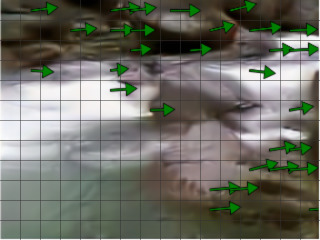}
   		\includegraphics[width=1\linewidth]{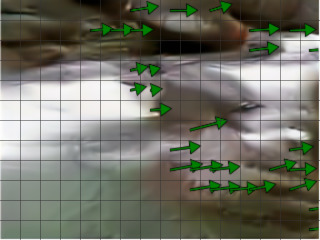}
	\end{subfigure}
	\begin{subfigure}{0.23\linewidth}
        \caption*{ConvLSTM}
        \vspace{-7pt}
	    \includegraphics[width=1\linewidth]{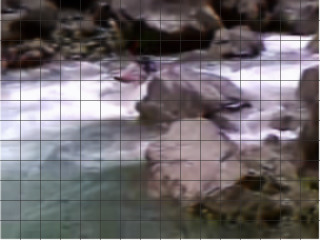} 
   		\includegraphics[width=1\linewidth]{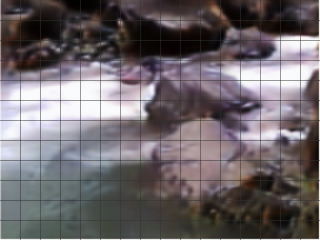}
   		\includegraphics[width=1\linewidth]{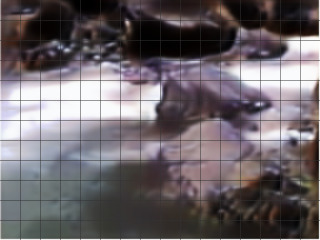}
   		\includegraphics[width=1\linewidth]{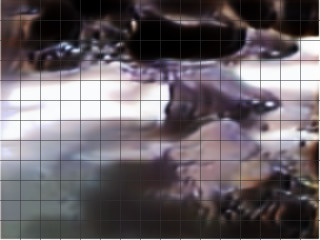}
	\end{subfigure}
	\begin{subfigure}{0.23\linewidth}
        \caption*{\cite{Mathieu15}}
        \vspace{-7pt}
	    \includegraphics[width=1\linewidth]{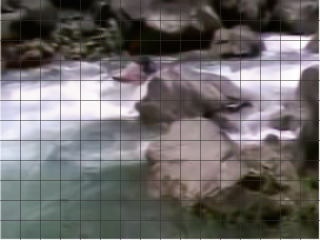} 
   		\includegraphics[width=1\linewidth]{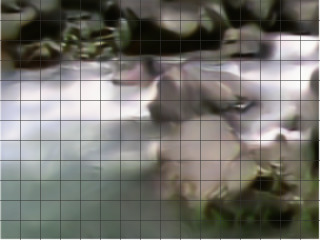}
   		\includegraphics[width=1\linewidth]{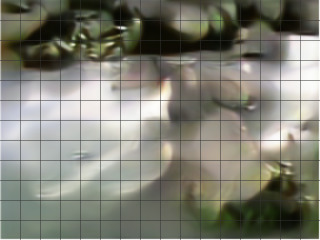}
   		\includegraphics[width=1\linewidth]{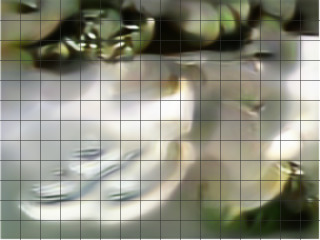}
	\end{subfigure}
    \vspace{.1cm}
    \hspace*{-1cm} \\
    \centering
    \hspace*{-.7cm}
% ----------------------------------------------------------------------------------------------------------------

    \hspace*{-.8cm}
    \begin{subfigure}{0.04\linewidth}
        \raggedleft
        \rotatebox{90}{
        \parbox{0.1cm}{\rotatebox{-90}{t=11}} \hspace{2.1cm} \parbox{0.1cm}{\rotatebox{-90}{t=9}} \hspace{2.3cm} \parbox{0.1cm}{\rotatebox{-90}{t=7}} \hspace{2.1cm} \parbox{0.1cm}{\rotatebox{-90}{t=5}} \hspace{.6cm}
        }
    \end{subfigure}
    \begin{subfigure}{0.23\linewidth}
        \caption*{}
        \vspace{-7pt}
	    \includegraphics[width=1\linewidth]{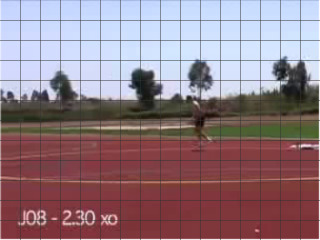}
   		\includegraphics[width=1\linewidth]{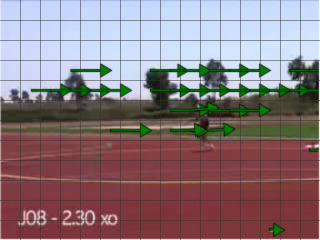}
   		\includegraphics[width=1\linewidth]{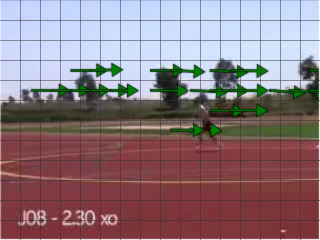}
   		\includegraphics[width=1\linewidth]{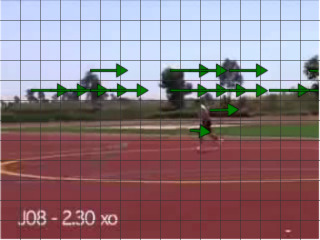} 
	\end{subfigure} 
    \begin{subfigure}{0.23\linewidth}
        \caption*{}
        \vspace{-7pt}
	    \includegraphics[width=1\linewidth]{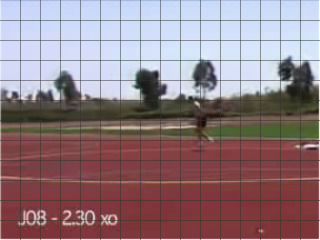} 
   		\includegraphics[width=1\linewidth]{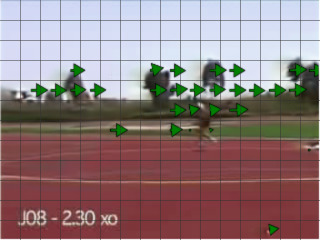}
   		\includegraphics[width=1\linewidth]{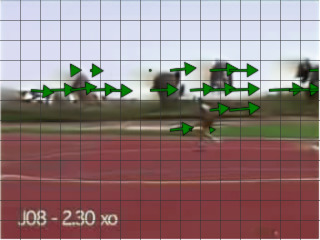}
   		\includegraphics[width=1\linewidth]{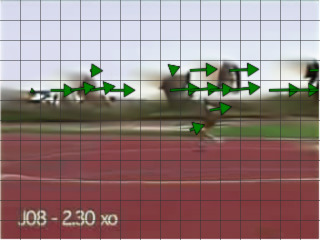}
	\end{subfigure}
	\begin{subfigure}{0.23\linewidth}
        \caption*{}
        \vspace{-7pt}
	    \includegraphics[width=1\linewidth]{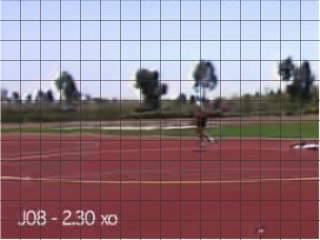} 
   		\includegraphics[width=1\linewidth]{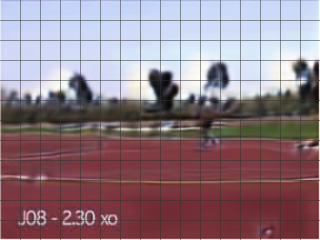}
   		\includegraphics[width=1\linewidth]{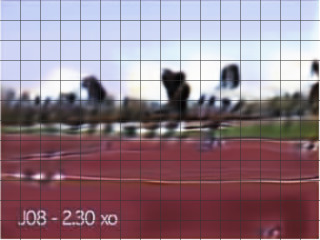}
   		\includegraphics[width=1\linewidth]{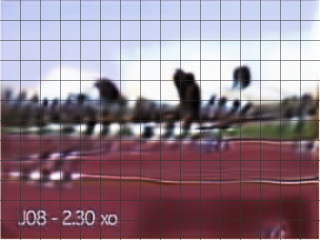}
	\end{subfigure}
	\begin{subfigure}{0.23\linewidth}
        \caption*{}
        \vspace{-7pt}
	    \includegraphics[width=1\linewidth]{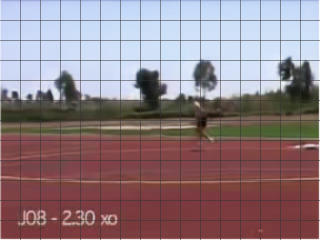} 
   		\includegraphics[width=1\linewidth]{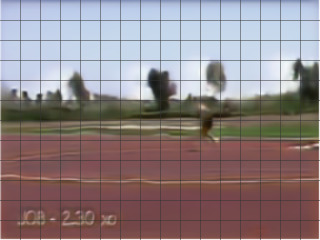}
   		\includegraphics[width=1\linewidth]{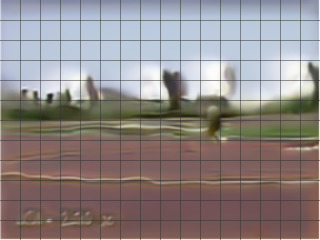}
   		\includegraphics[width=1\linewidth]{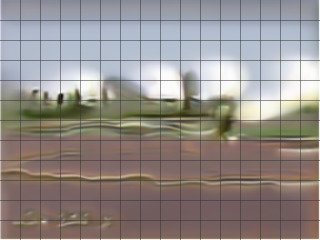}
	\end{subfigure}
	\hspace{-.8cm}

	\vspace{-5pt}
    \caption{Qualitative comparisons on UCF-101. We display predictions (in every other frame) starting from the $5^{\text{th}}$ frame. The green arrows denote the top-30 closest optical flow vectors within image patches between MCnet and ground-truth. More clear motion prediction can be seen in the \href{https://goo.gl/nG8ve1}{\color{blue} project website}.} %~\url{https://sites.google.com/a/umich.edu/rubenevillegas/iclr2017}.
\label{fig:ucf101_qualitative4}
\vspace{-.5cm}
\end{figure}

In this section, we show frame prediction examples in which considerable camera motion occurs.
We analyze the effects of camera motion on our best network and the corresponding baselines. First, we analyze qualitative examples on UCF101 (more complicated camera motion) and then on KTH (zoom-in and zoom-out camera effect).
\paragraph{UCF101 Results.}
As seen in Figure \ref{fig:ucf101_qualitative3} and Figure \ref{fig:ucf101_qualitative4}, our model handles foreground and camera motion for a few steps.
We hypothesize that for the first few steps, motion signals from images are clear.
However, as images are predicted, motion signals start to deteriorate due to prediction errors.
When a considerable amount of camera motion is present in image sequences, the motion signals are very dense.
As predictions evolve into the future, our motion encoder has to handle large motion deterioration due to prediction errors, which cause motion signals to get easily confused and lost quickly.
\vspace*{1cm}

\newpage

\begin{figure}[hbt!]
    \vspace{.1cm}
    \hspace*{-.7cm}
    \centering
%---------------------------------------------------------------------------------------------
    \begin{subfigure}{0.04\linewidth}
        \raggedleft
        \rotatebox{90}{
        \hspace{-.4cm}
        \parbox{2cm}{\centering G.T.} \hspace{-.3cm} \parbox{2cm}{\centering ConvLSTM} \hspace{-.3cm} \parbox{2cm}{\centering MCnet}
        }
    \end{subfigure}
    \begin{subfigure}{0.13\linewidth}
        \caption*{t=12}
        \vspace{-7pt}
	    \includegraphics[width=1\linewidth]{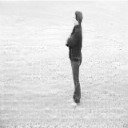} 
   		\includegraphics[width=1\linewidth]{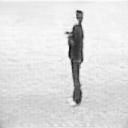}
   		\includegraphics[width=1\linewidth]{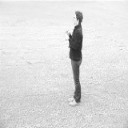} 
	\end{subfigure} 
    \begin{subfigure}{0.13\linewidth}
        \caption*{t=15}
        \vspace{-7pt}
	    \includegraphics[width=1\linewidth]{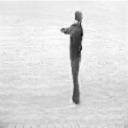} 
   		\includegraphics[width=1\linewidth]{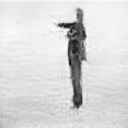}
   		\includegraphics[width=1\linewidth]{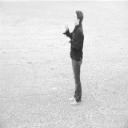}
	\end{subfigure} 
    \begin{subfigure}{0.13\linewidth}
        \caption*{t=18}
        \vspace{-7pt}
	    \includegraphics[width=1\linewidth]{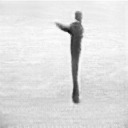} 
   		\includegraphics[width=1\linewidth]{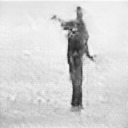}
   		\includegraphics[width=1\linewidth]{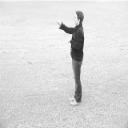}
	\end{subfigure} 
    \begin{subfigure}{0.13\linewidth}
        \vspace{21pt}
        \caption*{t=21}
        \vspace{-7pt}
	    \includegraphics[width=1\linewidth]{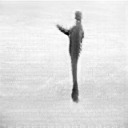} 
   		\includegraphics[width=1\linewidth]{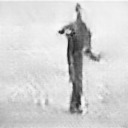}
   		\includegraphics[width=1\linewidth]{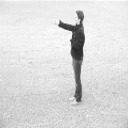}
   		\caption*{Boxing}
	\end{subfigure}
	\begin{subfigure}{0.13\linewidth}
	    \caption*{t=24}
        \vspace{-7pt}
	    \includegraphics[width=1\linewidth]{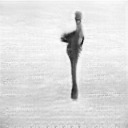} 
   		\includegraphics[width=1\linewidth]{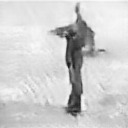}
   		\includegraphics[width=1\linewidth]{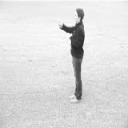}
	\end{subfigure}
	\begin{subfigure}{0.13\linewidth}
	    \caption*{t=27}
        \vspace{-7pt}
	    \includegraphics[width=1\linewidth]{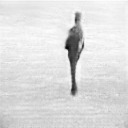} 
   		\includegraphics[width=1\linewidth]{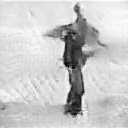}
   		\includegraphics[width=1\linewidth]{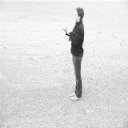}
	\end{subfigure}
	\begin{subfigure}{0.13\linewidth}
	    \caption*{t=30}
        \vspace{-7pt}
	    \includegraphics[width=1\linewidth]{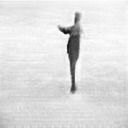} 
   		\includegraphics[width=1\linewidth]{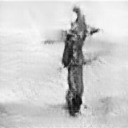}
   		\includegraphics[width=1\linewidth]{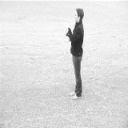}
	\end{subfigure}
    \vspace{.1cm}
    \hspace*{-.7cm} \\
    \centering
    \hspace*{-.7cm}
%-----------------------------------------------------------------------------------------------
    \begin{subfigure}{0.04\linewidth}
        \raggedleft
        \rotatebox{90}{
        \hspace{.1cm}
        \parbox{2cm}{\centering G.T.} \hspace{-.3cm} \parbox{2cm}{\centering ConvLSTM} \hspace{-.3cm} \parbox{2cm}{\centering MCnet}
        }
    \end{subfigure}
    \begin{subfigure}{0.13\linewidth}
        \vspace{-7pt}
	    \includegraphics[width=1\linewidth]{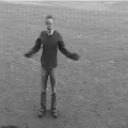} 
   		\includegraphics[width=1\linewidth]{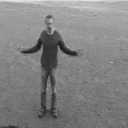}
   		\includegraphics[width=1\linewidth]{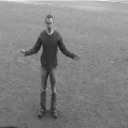} 
	\end{subfigure} 
    \begin{subfigure}{0.13\linewidth}
        \vspace{-7pt}
	    \includegraphics[width=1\linewidth]{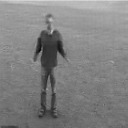} 
   		\includegraphics[width=1\linewidth]{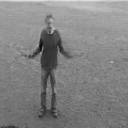}
   		\includegraphics[width=1\linewidth]{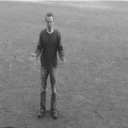}
	\end{subfigure} 
    \begin{subfigure}{0.13\linewidth}
        \vspace{-7pt}
	    \includegraphics[width=1\linewidth]{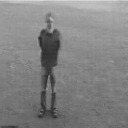} 
   		\includegraphics[width=1\linewidth]{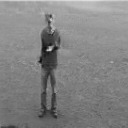}
   		\includegraphics[width=1\linewidth]{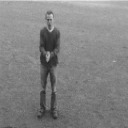}
	\end{subfigure} 
    \begin{subfigure}{0.13\linewidth}
        \vspace{9pt}
	    \includegraphics[width=1\linewidth]{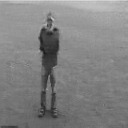} 
   		\includegraphics[width=1\linewidth]{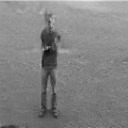}
   		\includegraphics[width=1\linewidth]{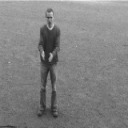}
   		\caption*{Handclapping}
	\end{subfigure}
	\begin{subfigure}{0.13\linewidth}
        \vspace{-7pt}
	    \includegraphics[width=1\linewidth]{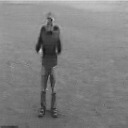} 
   		\includegraphics[width=1\linewidth]{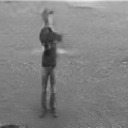}
   		\includegraphics[width=1\linewidth]{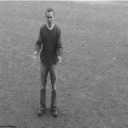}
	\end{subfigure}
	\begin{subfigure}{0.13\linewidth}
        \vspace{-7pt}
	    \includegraphics[width=1\linewidth]{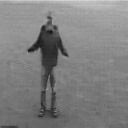} 
   		\includegraphics[width=1\linewidth]{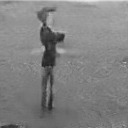}
   		\includegraphics[width=1\linewidth]{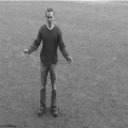}
	\end{subfigure}
	\begin{subfigure}{0.13\linewidth}
        \vspace{-7pt}
	    \includegraphics[width=1\linewidth]{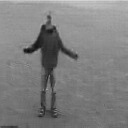} 
   		\includegraphics[width=1\linewidth]{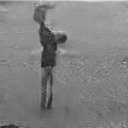}
   		\includegraphics[width=1\linewidth]{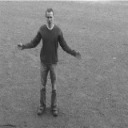}
	\end{subfigure}
    \vspace{.1cm}
    \hspace*{-.7cm} \\
    \centering

%-----------------------------------------------------------------------------------------------
    \caption{Qualitative comparisons on KTH testset. We display predictions starting from the $12^{\text{th}}$ frame, in every $3$ timesteps. More clear motion prediction can be seen in the \href{https://goo.gl/nG8ve1}{\color{blue} project website}.} %~\url{https://sites.google.com/a/umich.edu/rubenevillegas/iclr2017}}
\label{fig:kth_qualitative4}
\vspace{-.5cm}
\end{figure}

\newpage
\paragraph{KTH Results.}
We were unable to find videos with background motion in the KTH dataset, but we found videos where the camera is zooming in or out for the actions of boxing, handclapping, and handwaving.
In Figure \ref{fig:kth_qualitative4}, we display qualitative for such videos.
Our model is able to predict the zoom change in the cameras, while continuing the action motion.
In comparison to the performance observed in UCF101, the background does not change much.
Thus, the motion signals are well localized in the foreground motion (human), and do not get confused with the background and lost as quickly.

\newpage

\section{Extended quantitative evaluation}
In this section, we show additional quantitative comparison with a baseline based on copying the last observed frame through time for KTH and UCF101 datasets.
Copying the last observed frame through time ensures perfect background prediction in videos where most of the motion comes from foreground (i.e. person performing an action).
However, if such foreground composes a small part of the video, it will result in high prediction quality score regardless of the simple copying action.

In Figure \ref{fig:extra_quantitative} below, we can see the quantitative comparison in the datasets.
Copying the last observed frame through time does a reasonable job in both datasets, however, the impact is larger in UCF101.
% Videos in the KTH dataset comprise simple backgrounds with minimal camera motion, which allows our network to easily predict both foreground and background motion, resulting in better image quality scores, though not by a large margin compared to the simple copy baseline.
Videos in the KTH dataset comprise simple background with minimal camera motion, which allows our network to easily predict both foreground and background motion, resulting in better image quality scores.
% However, videos in UCF101 contain more complicated and diverse background which in combination with camera motion present a much greater challenge to video prediction networks during learning and evaluation.
However, videos in UCF101 contain more complicated and diverse background which in combination with camera motion present a much greater challenge to video prediction networks.
From the qualitative results in Section~\ref{sec:extquant} and Figures~\ref{fig:ucf101_qualitative}, \ref{fig:ucf101_qualitative2}, \ref{fig:ucf101_qualitative3}, and \ref{fig:ucf101_qualitative4}, we can see that our network performs better in videos that contain isolated areas of motion compared to videos with dense motion.
% A simple copy paste operation of the last observed frame, ensures very high prediction scores in videos where motion only happens in small areas even though no motion has been predicted.
A simple copy$/$paste operation of the last observed frame, ensures very high prediction scores in videos where very small motion occur.
% Such high scores, push the simple baseline to a high overall performance.
The considerable score boost by videos with small motion causes the simple copy$/$paste baseline to outperform MCnet in the overall performance on UCF101.

% However, UCF101 dataset complicated backgrounds and larger camera motion, requires our network
% UCF101 videos contain complicated and varied backgrounds.
% Predicting the motion of such backgrounds in combination with foreground motion is a more challenging task than predicting mostly foreground motion.
% Background prediction noise quickly takes over, and causes foreground and background motion to become unidentifiable.
% Therefore, our model loses sense of the structures moving in such videos quickly in comparison to videos with minimal camera motion.

\begin{figure*}[h!]
\centering
\includegraphics[width=0.49\linewidth] {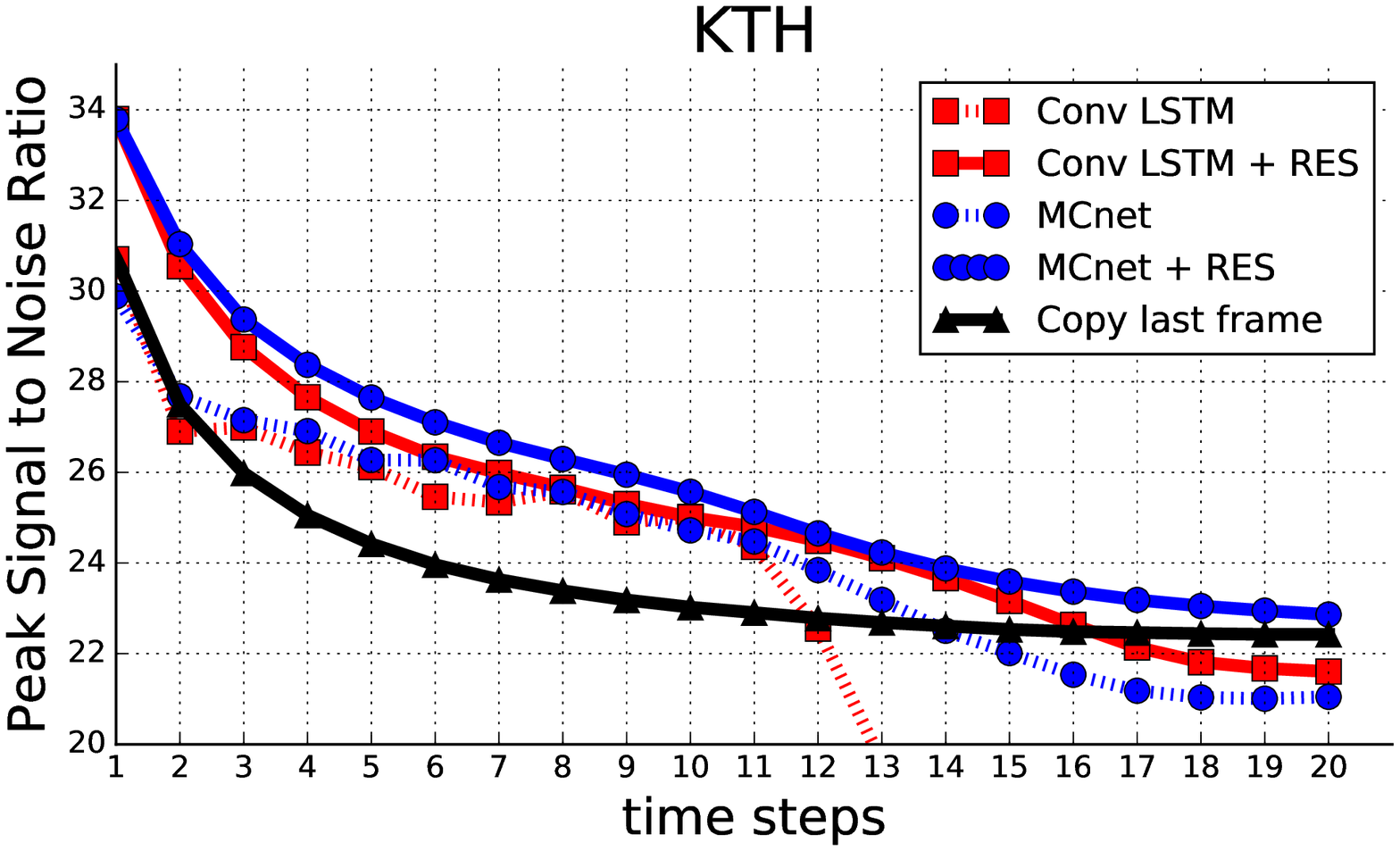} \hspace{0.1cm}
\includegraphics[width=0.49\linewidth] {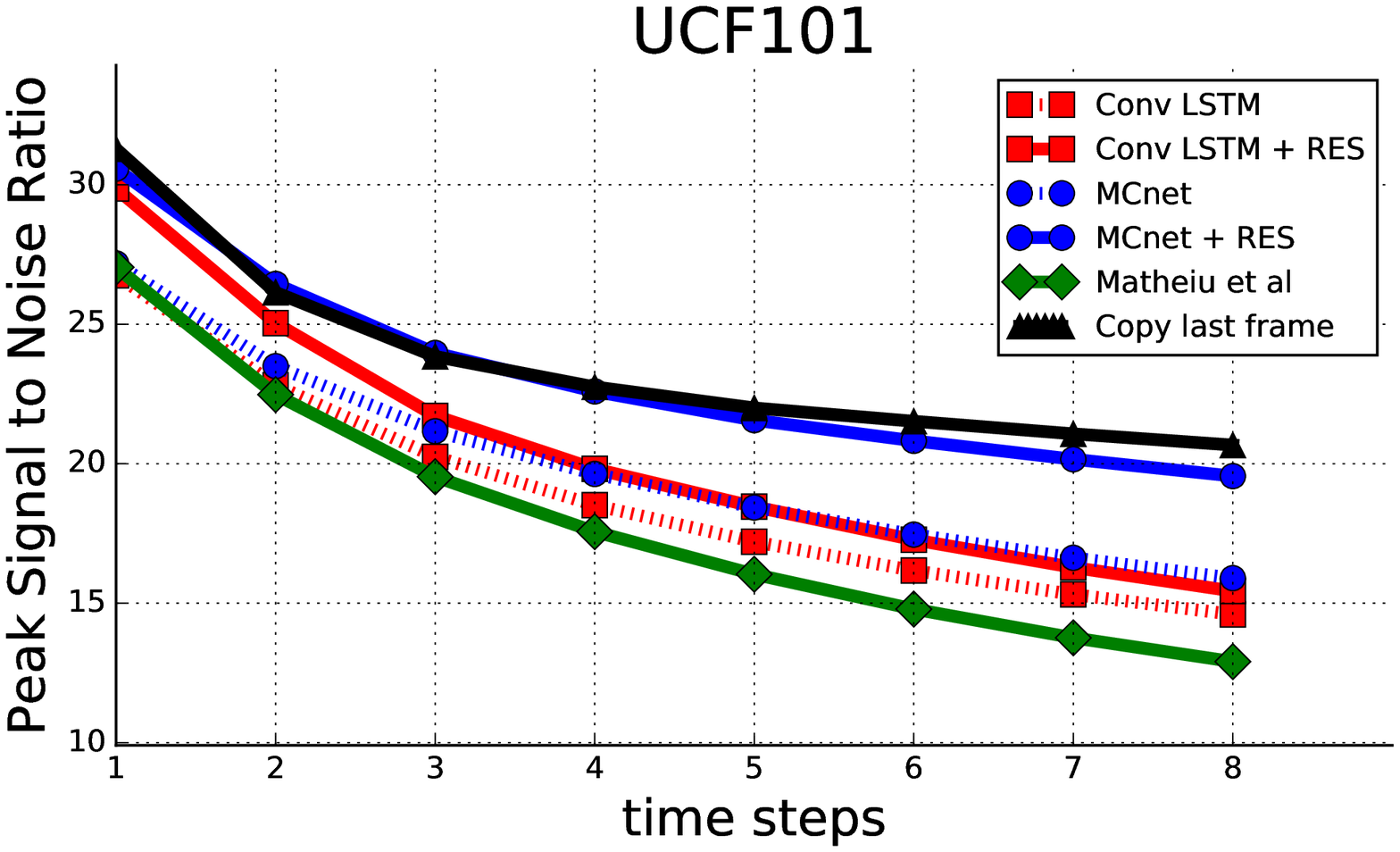} \hspace{0.1cm} \\
\includegraphics[width=0.49\linewidth] {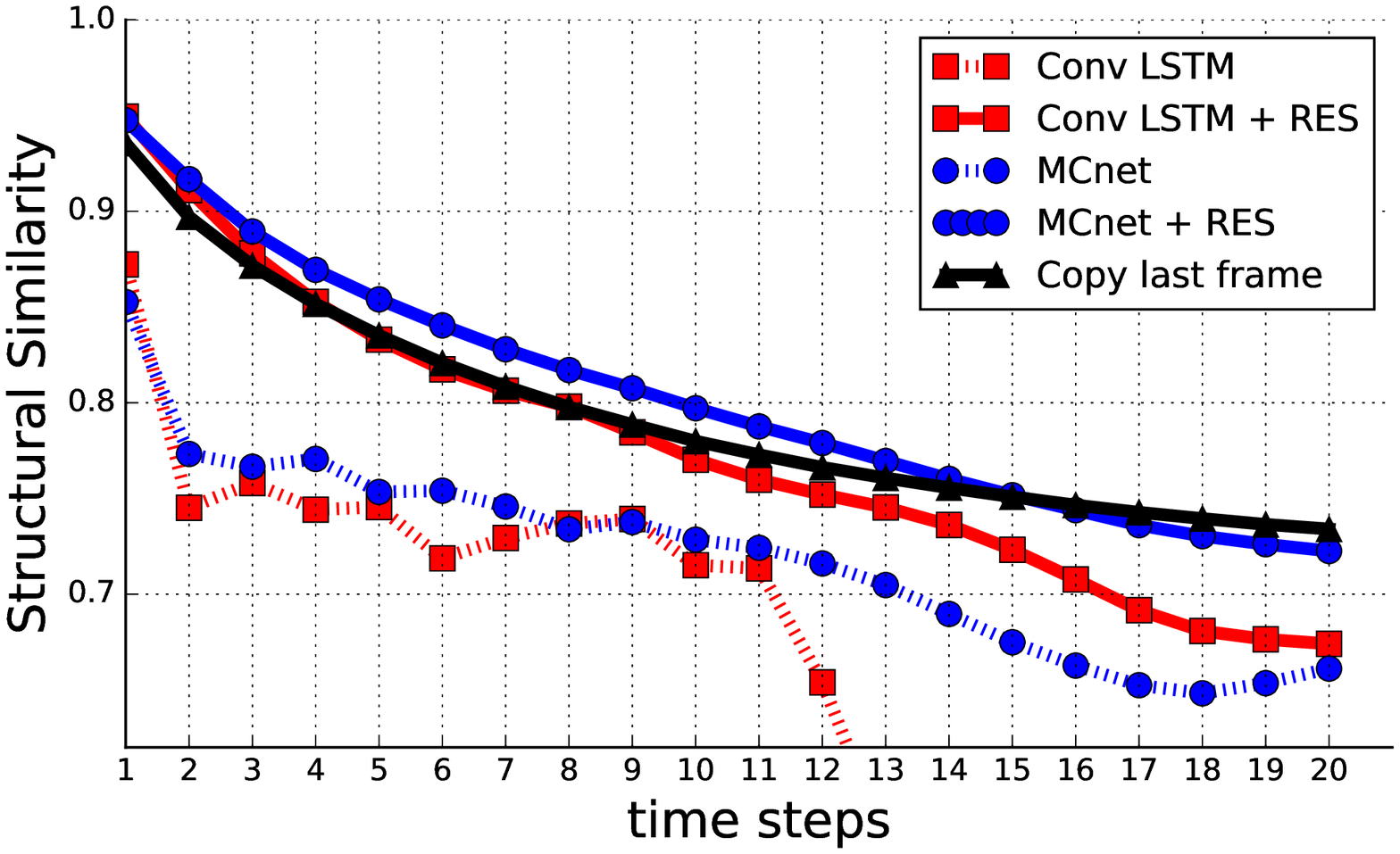} \hspace{0.1cm}
\includegraphics[width=0.49\linewidth] {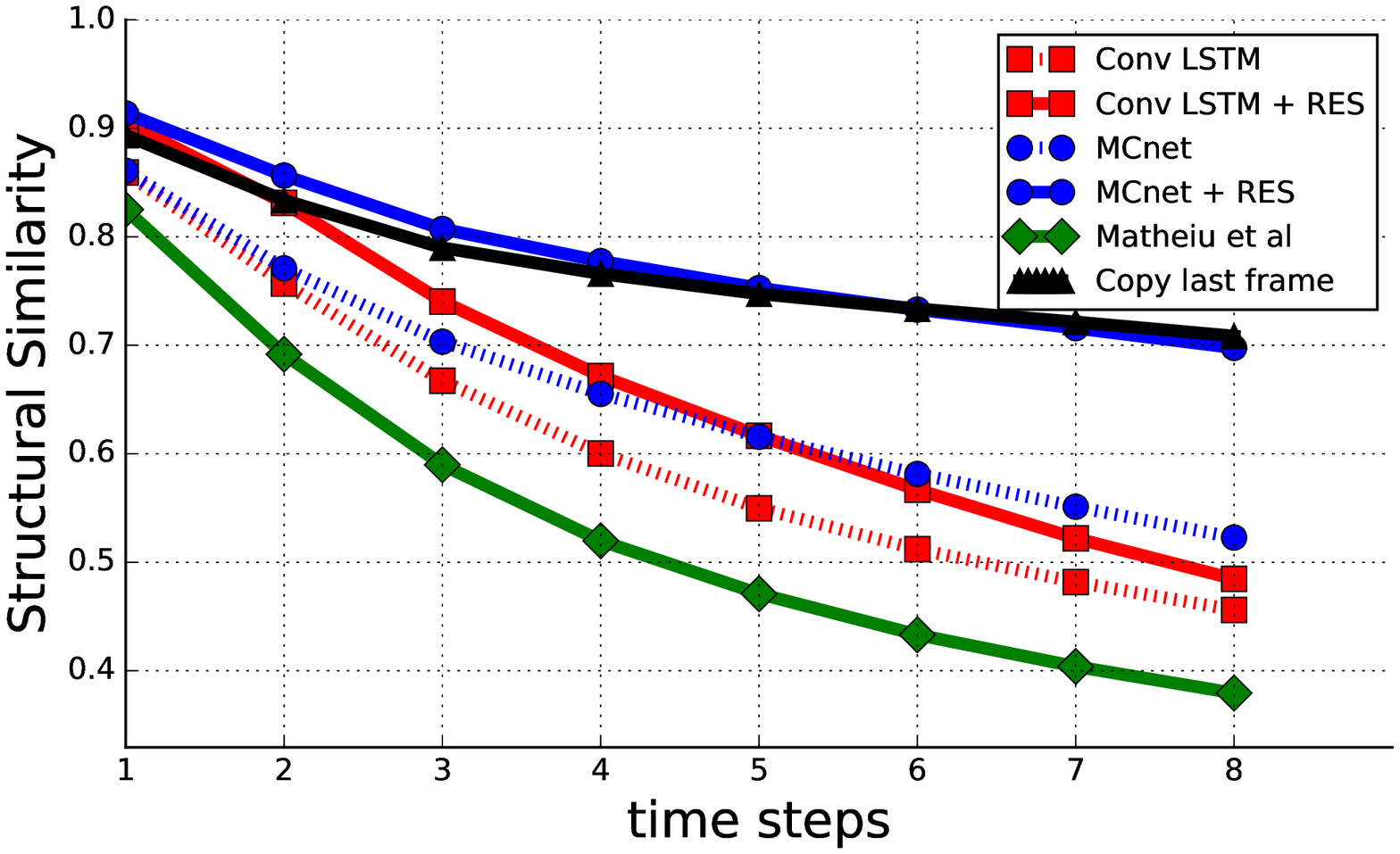} \hspace{0.1cm}
\vspace{-.2cm}
\caption{Extended quantitative comparison including a baseline based on copying the last observed frame through time.}
\label{fig:extra_quantitative}
\end{figure*}

\newpage

\section{UCF101 Motion Disambiguation Experiments}
Due to the observed bias from videos with small motion, we perform experiments by measuring the image quality scores on areas of motion.
These experiments are similar to the ones performed in \cite{Mathieu15}.
We compute DeepFlow optical flow \citep{deepflow} between the previous and the current groundtruth image of interest, compute the magnitude, and normalize it to $[0,1]$.
The computed optical flow magnitude is used to mask the pixels where motion was observed.
We set the pixels where the optical flow magnitude is less than 0.2, and leave all other pixels untouched in both the groundtruth and predicted images.
Additionally, we separate the test videos by the average $\ell_2$-norm of time difference between target frames.
We separate the test videos into deciles based of the computed average $\ell_2$-norms, and compute image quality on each decile.
Intuitively, the $1^{st}$ decile contains videos with the least overall of motion (i.e. frames that show the smallest change over time), and the $10^{th}$ decile contains videos with the most overall motion (i.e. frames that show the largest change over time).

\begin{figure*}[htb!]
\centering
\vspace{.2cm}
\includegraphics[width=0.49\linewidth] {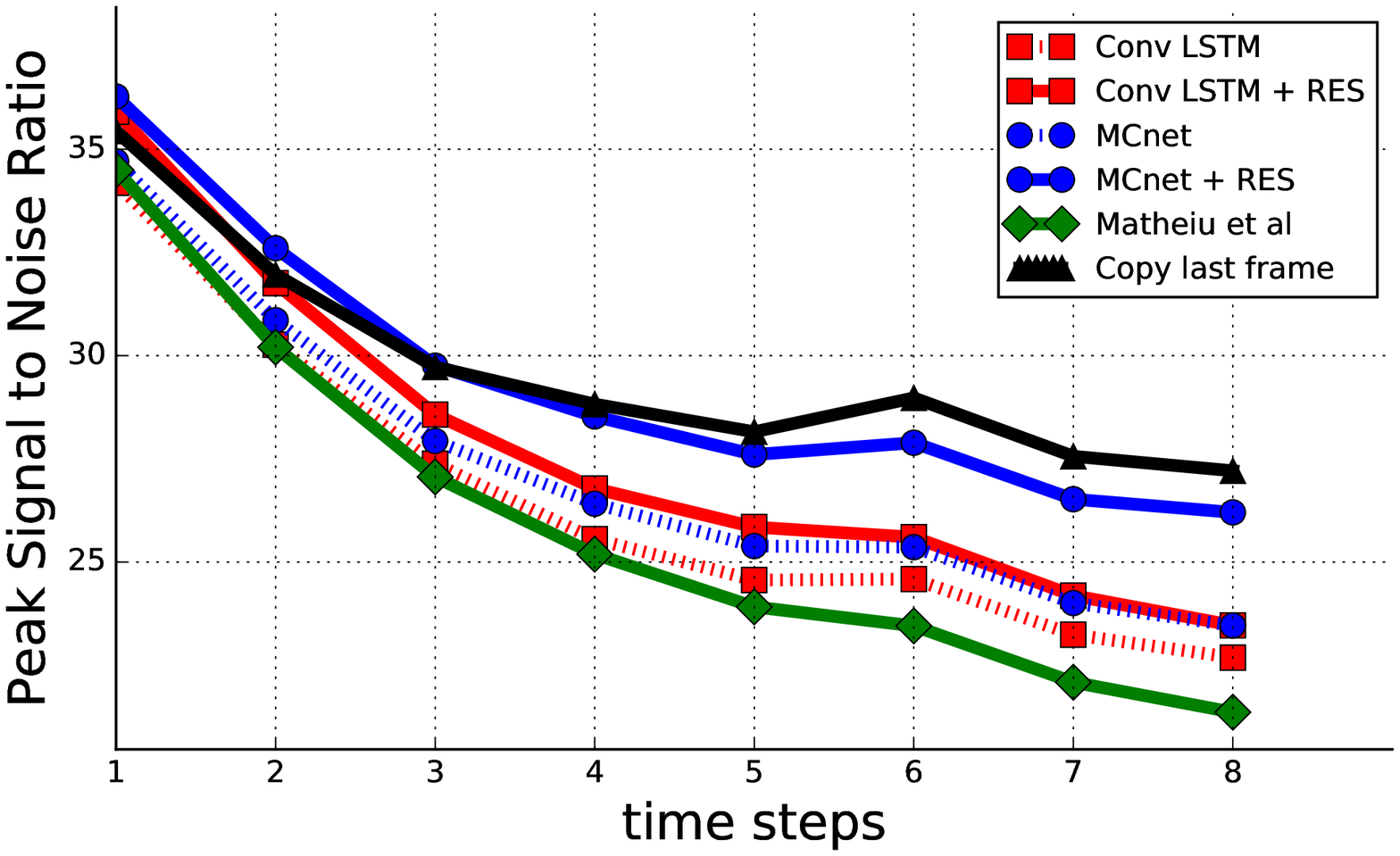} \hspace{0.1cm}
\includegraphics[width=0.49\linewidth] {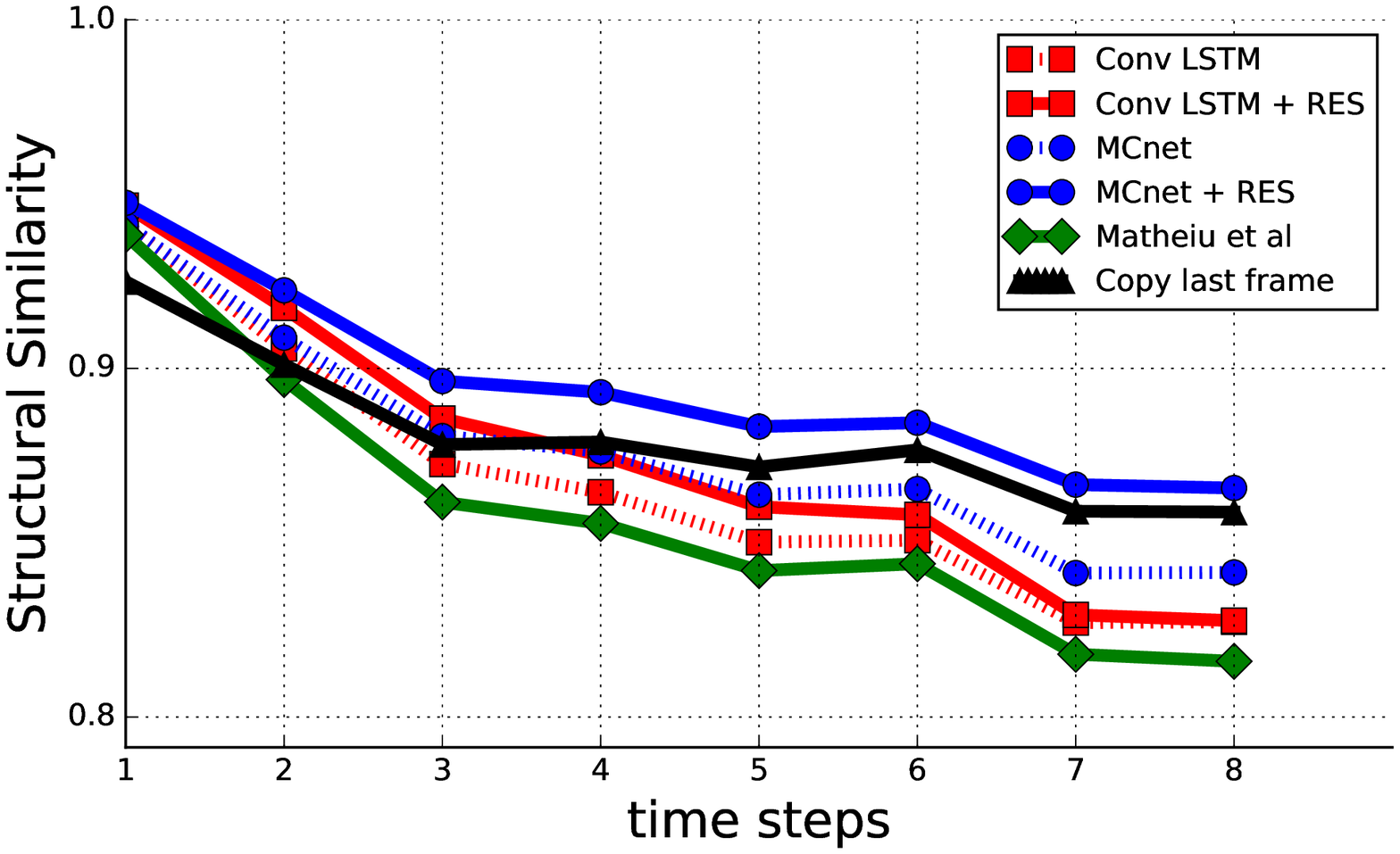} \hspace{0.1cm}
\vspace{-.2cm}
\caption{Extended quantitative comparison on UCF101 including a baseline based on copying the last observed frame through time using motion based pixel mask.}
\label{fig:extra_quantitative2}
\end{figure*}

As shown in Figure~\ref{fig:extra_quantitative2}, when we only evaluate on pixels where rough motion is observed, MCnet reflects higher PSNR and SSIM, and clearly outperforms all the baselines in terms of SSIM.
The SSIM results show that our network is able to predict a structure (i.e. textures, edges, etc) similar to the grountruth images within the areas of motion.
The PSNR results, however, show that our method outperforms the simple copy$/$paste baseline for the first few steps, but then our method performs slightly worse.
The discrepancies observed between PSNR and SSIM scores could be due to the fact that some of the predicted images may not reflect the exact pixel values of the groundtruth regardless of the structures being similar.
SSIM scores are known to take into consideration features in the image that go beyond directly matching pixel values, reflecting more accurately how humans perceived image quality.

\begin{figure*}[htb!]
\vspace{-.6cm}
\centering
\caption*{$10^{th}$ decile}
\vspace{-.4cm}
\includegraphics[width=0.49\linewidth] {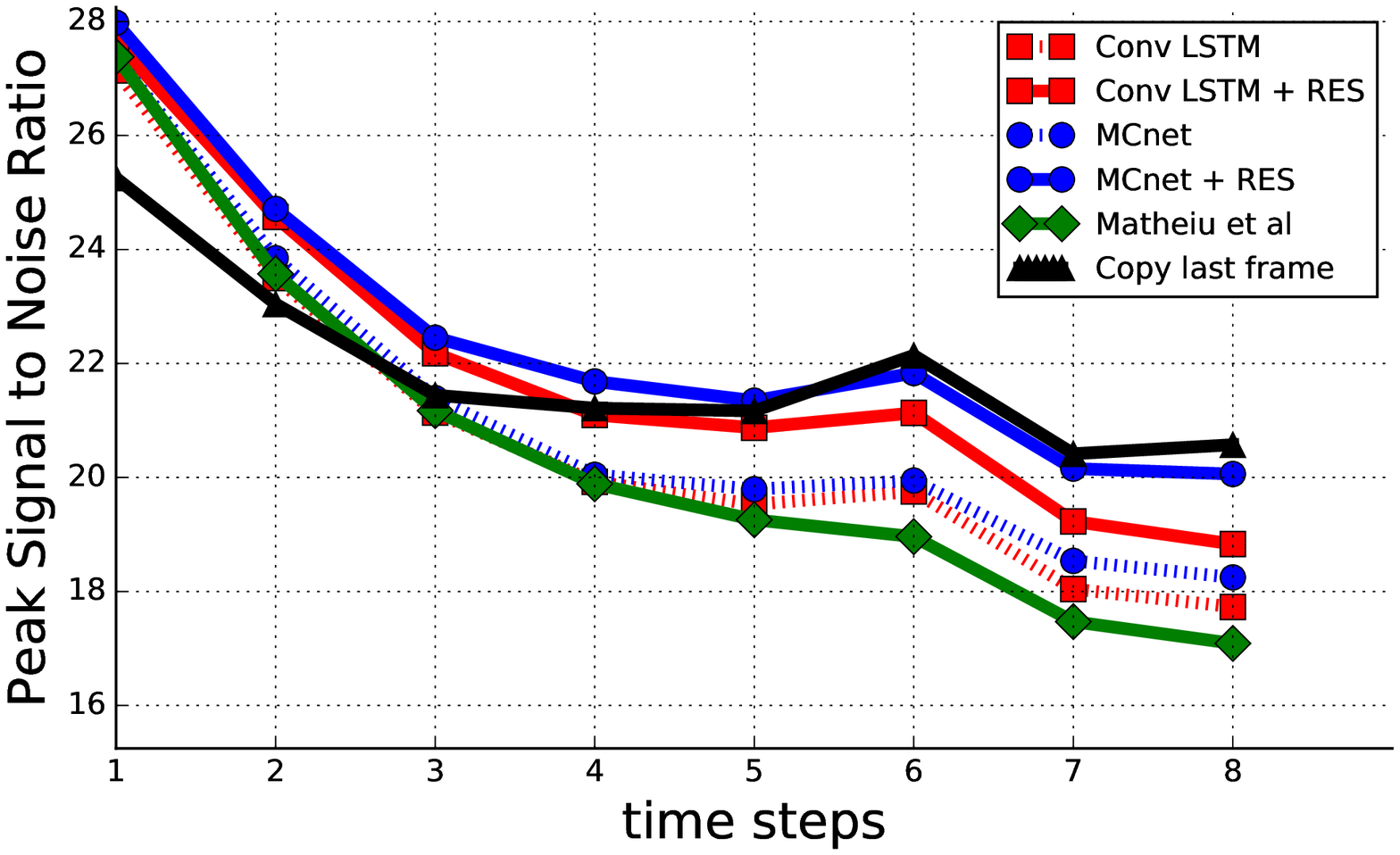} \hspace{.1cm}
\includegraphics[width=0.49\linewidth] {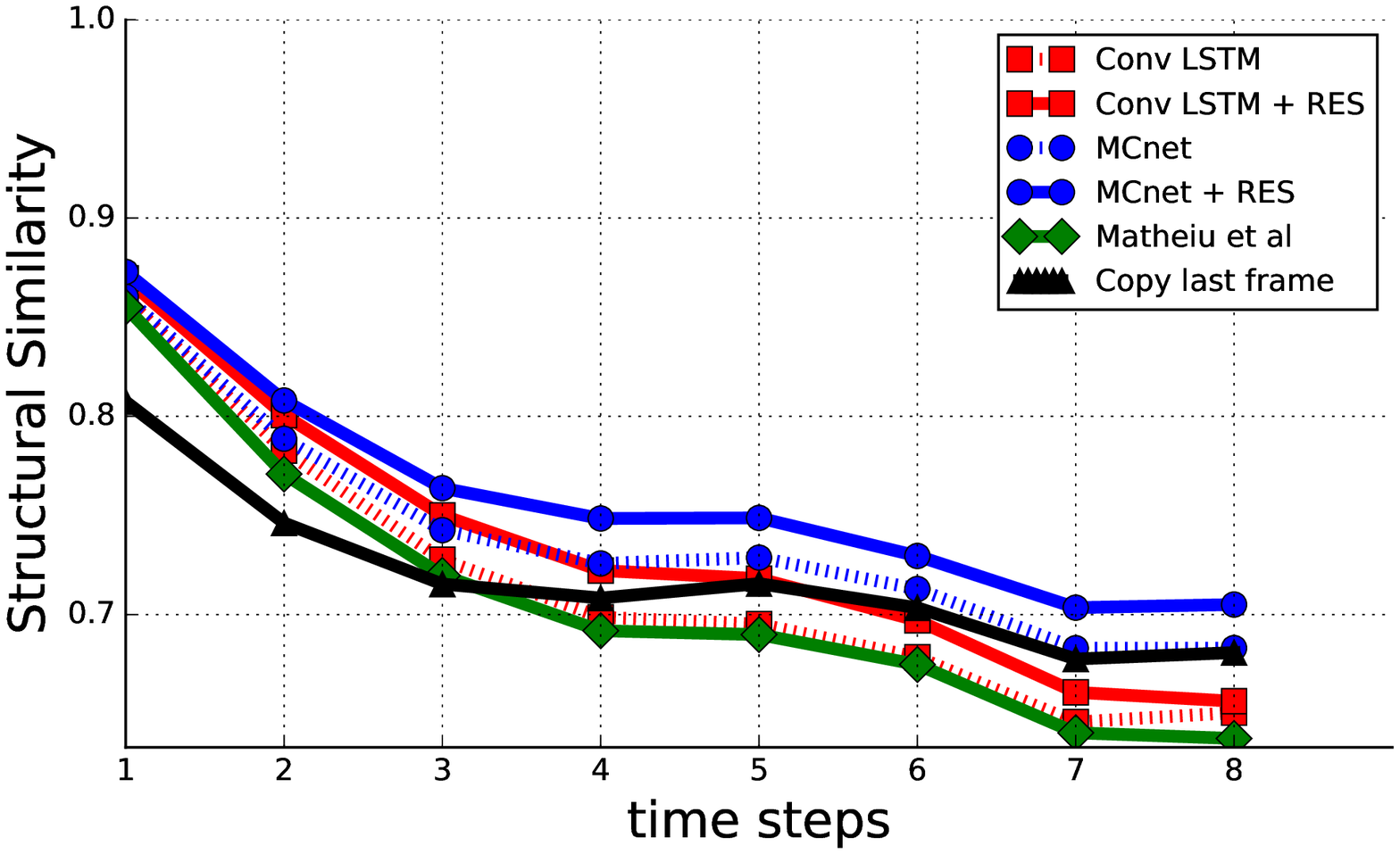} \vspace{-.6cm}\hspace{0.1cm}
\caption*{$9^{th}$ decile}
\vspace{-.4cm}
\includegraphics[width=0.49\linewidth] {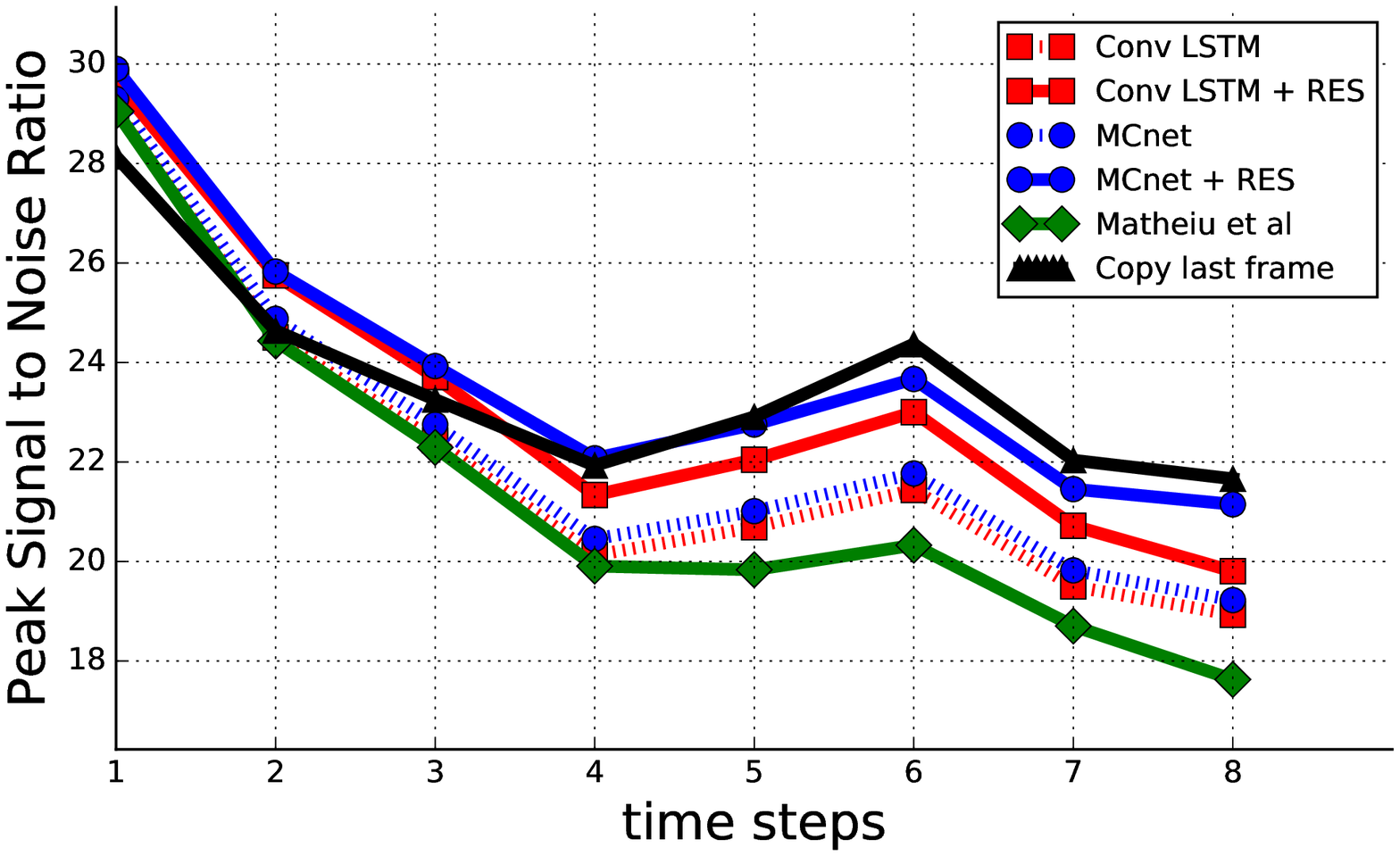} \hspace{.1cm}
\includegraphics[width=0.49\linewidth] {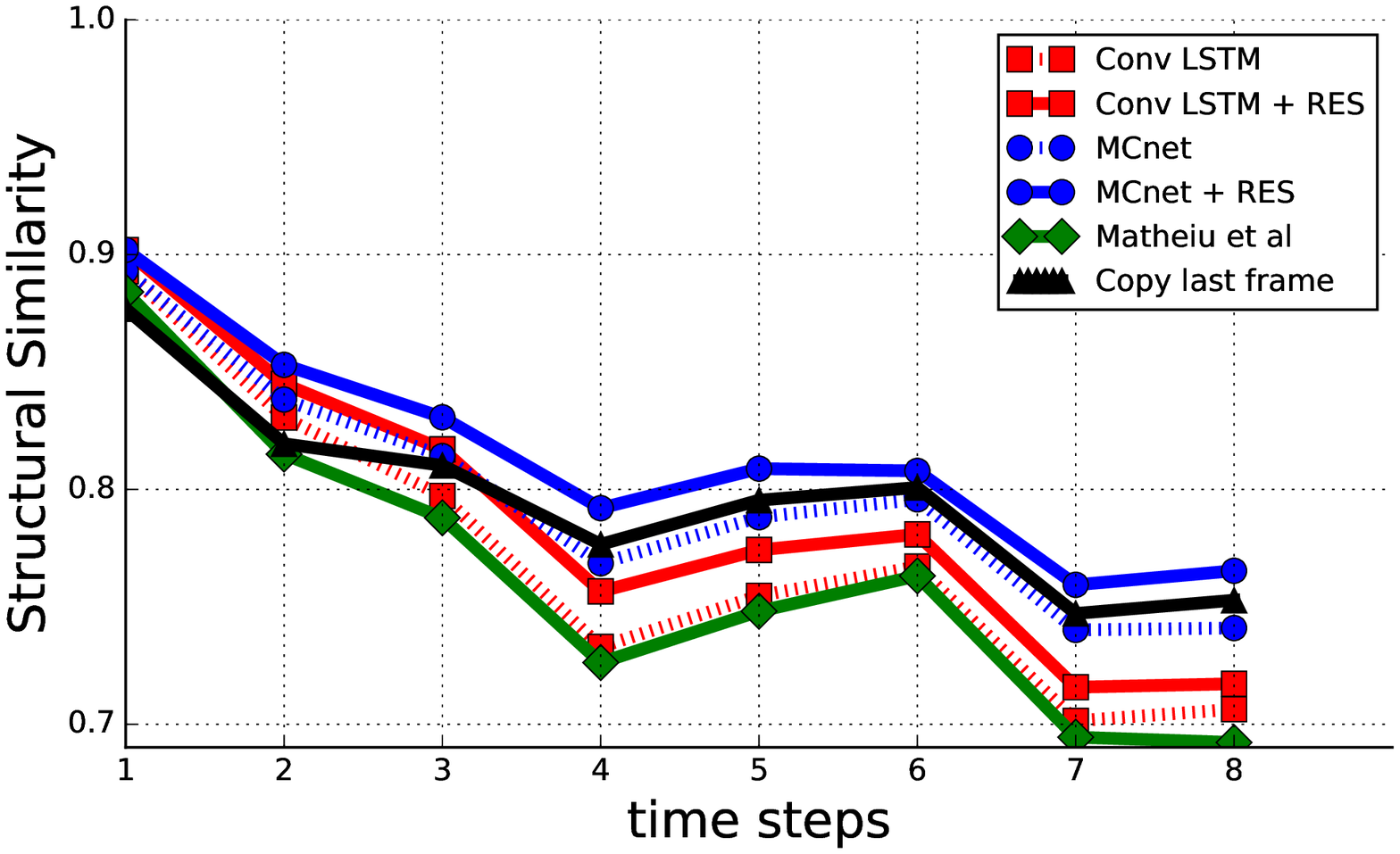} \vspace{-.6cm}\hspace{0.1cm}
\caption*{$8^{th}$ decile}
\vspace{-.4cm}
\includegraphics[width=0.49\linewidth] {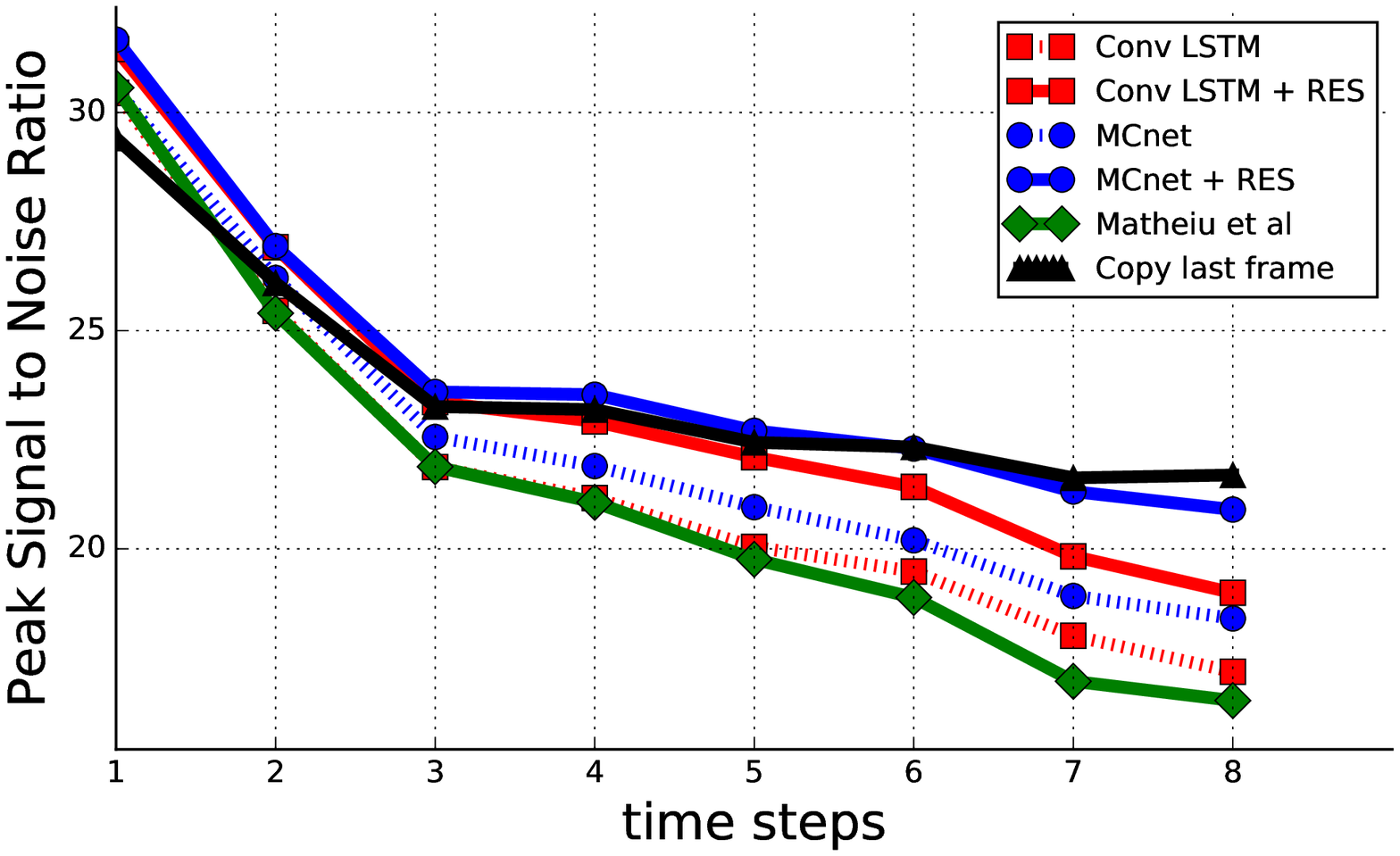} \hspace{.1cm}
\includegraphics[width=0.49\linewidth] {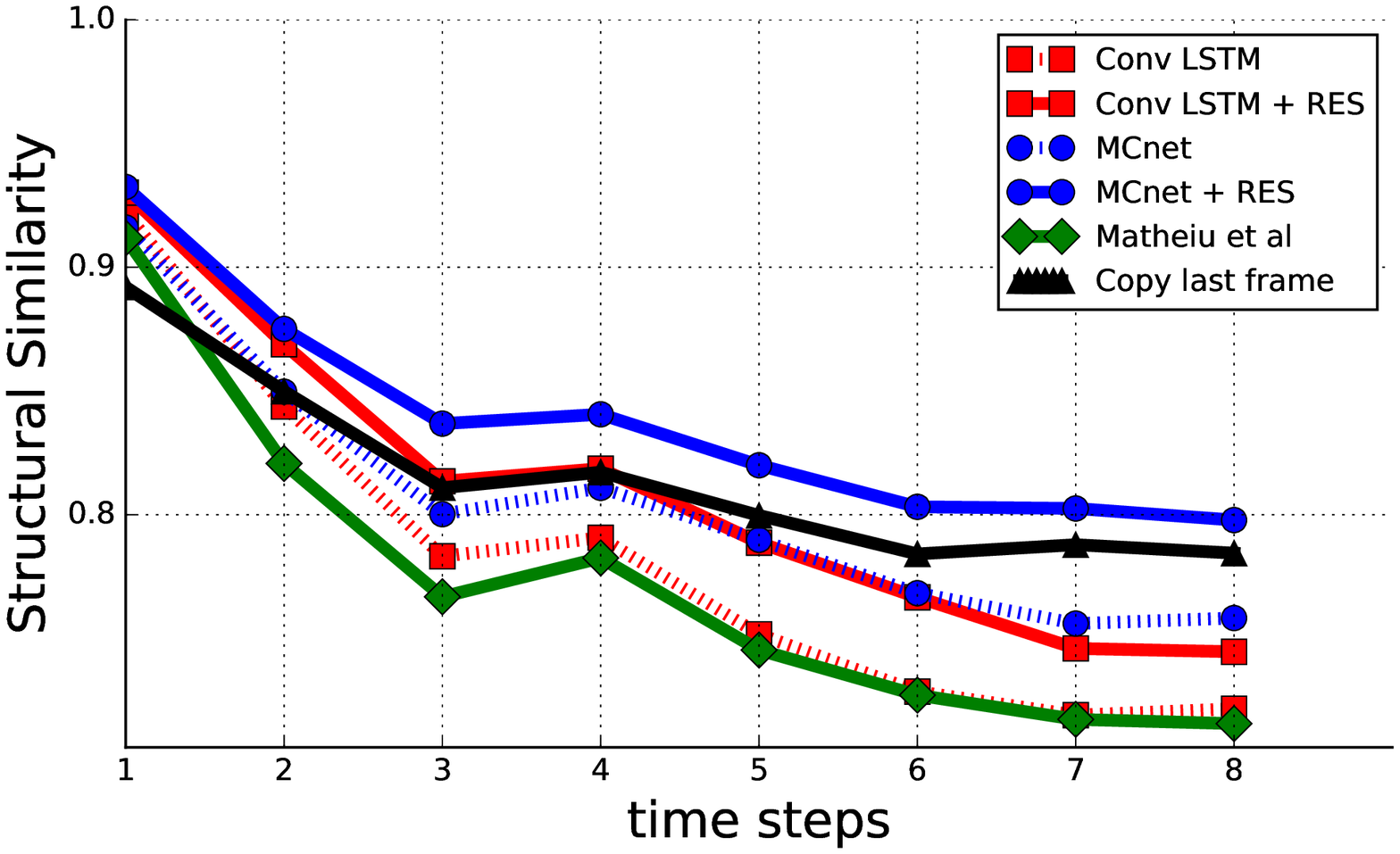} \vspace{-.6cm}\hspace{0.1cm}
\caption*{$7^{th}$ decile}
\vspace{-.4cm}
\includegraphics[width=0.49\linewidth] {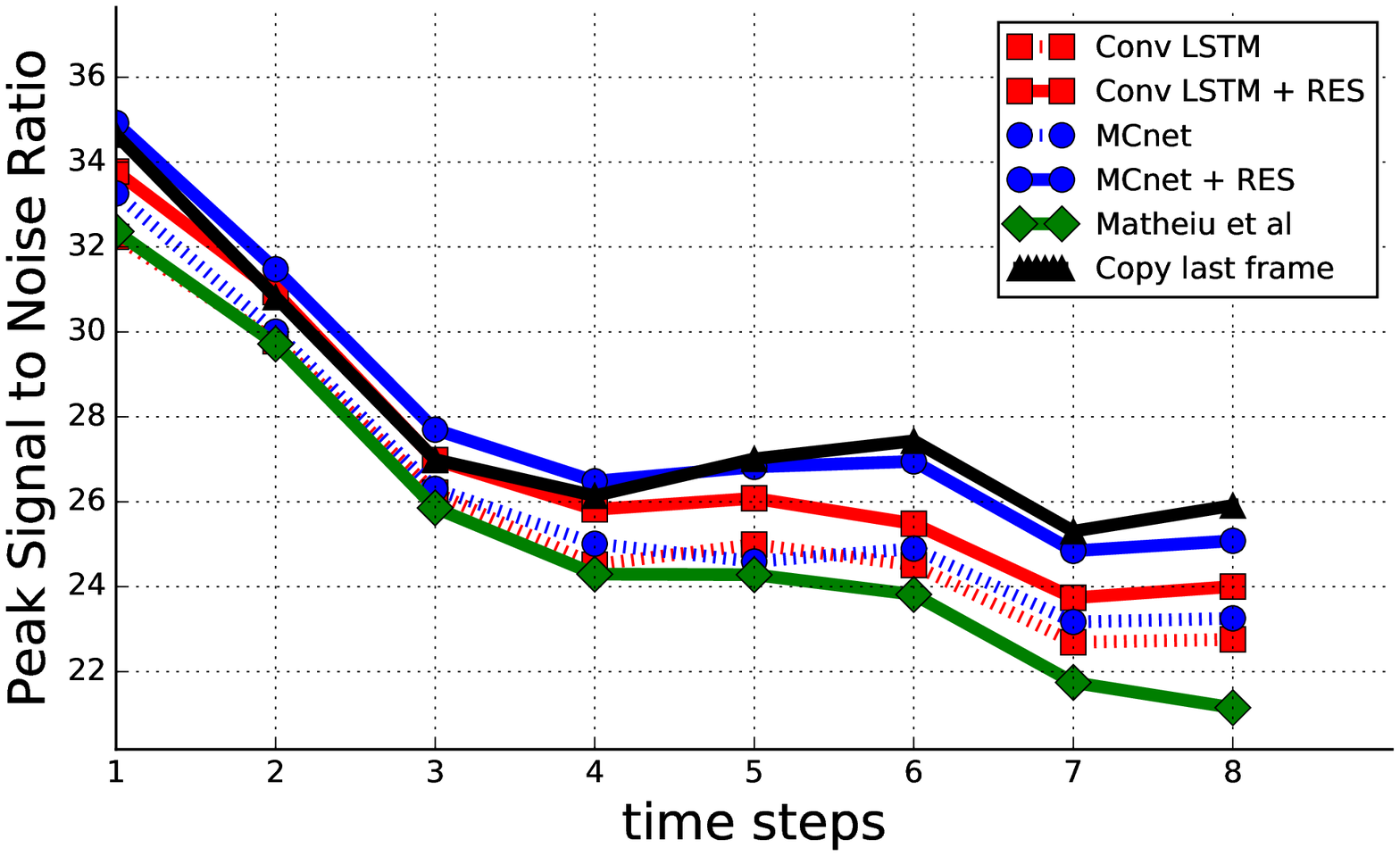} \hspace{.1cm}
\includegraphics[width=0.49\linewidth] {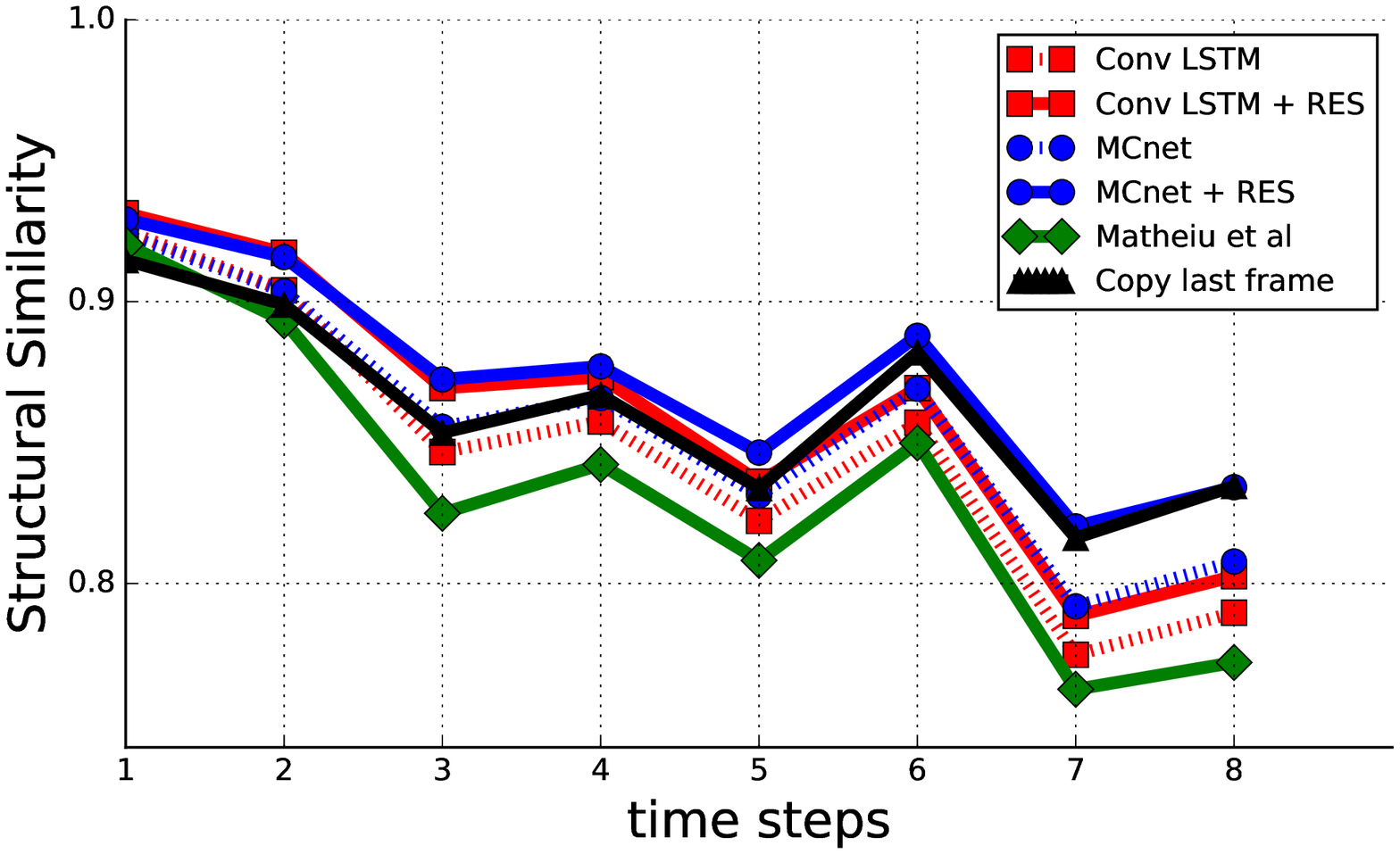} \vspace{-.6cm}\hspace{0.1cm}
\caption*{$6^{th}$ decile}
\vspace{-.4cm}
\includegraphics[width=0.49\linewidth] {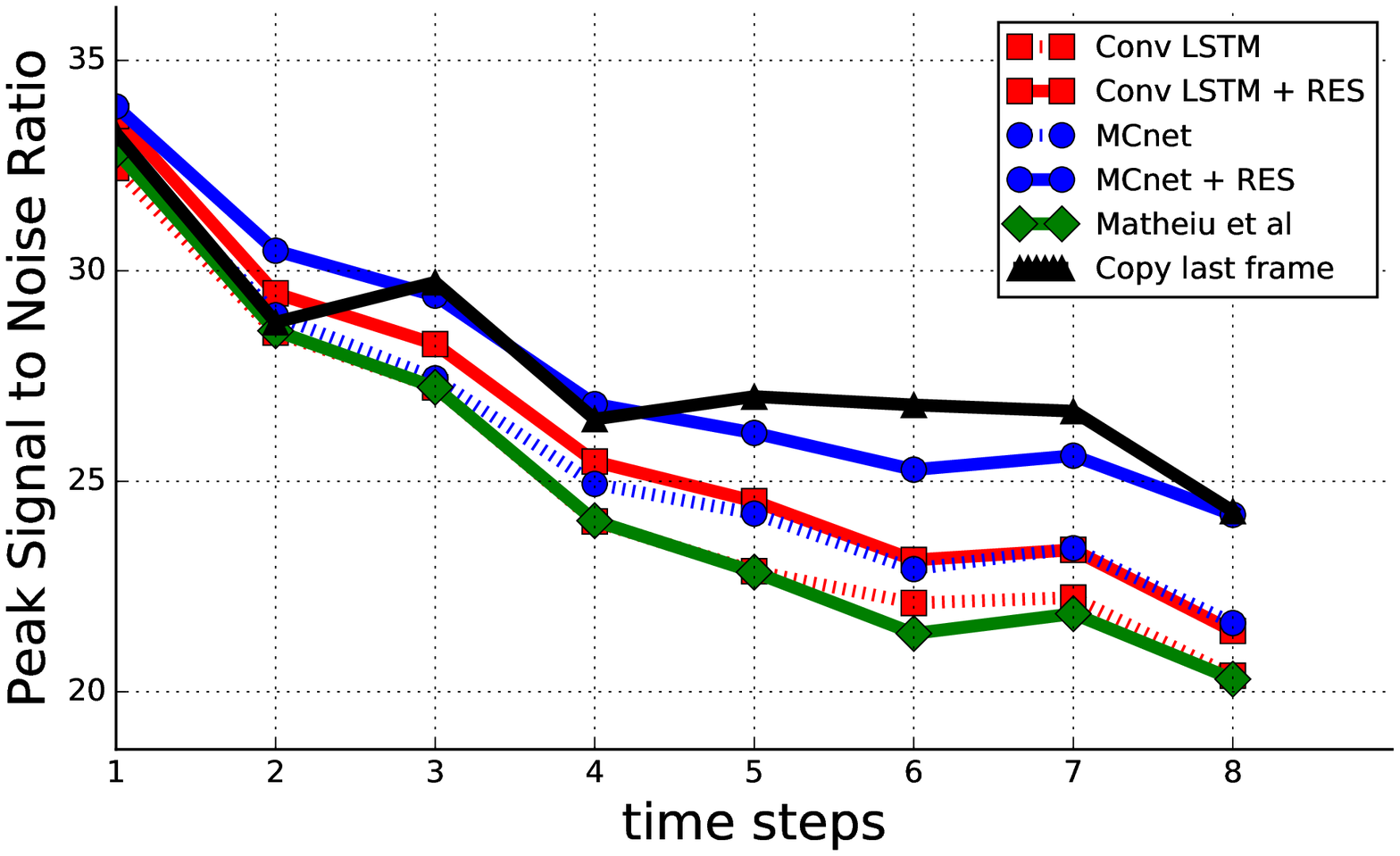} \hspace{.1cm}
\includegraphics[width=0.49\linewidth] {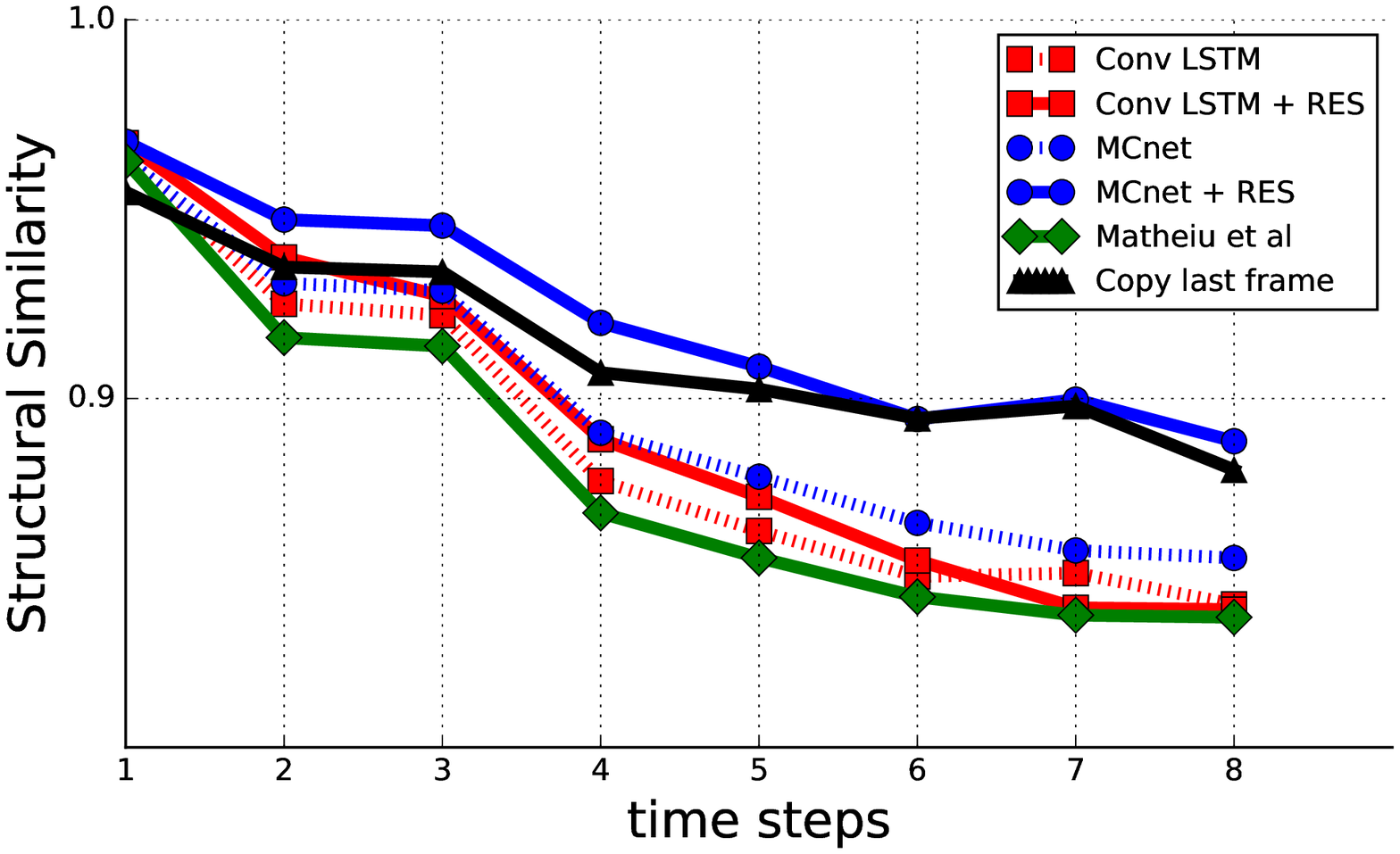} \hspace{0.1cm}
\caption{Quantitative comparison on UCF101 using motion based pixel mask, and separating dataset by average $\ell_2$-norm of time difference between target frames.}
\label{fig:extra_quantitative4}
\end{figure*}

\begin{figure*}[htb!]
\vspace{-.6cm}
\centering
\caption*{$5^{th}$ decile}
\vspace{-.4cm}
\includegraphics[width=0.49\linewidth] {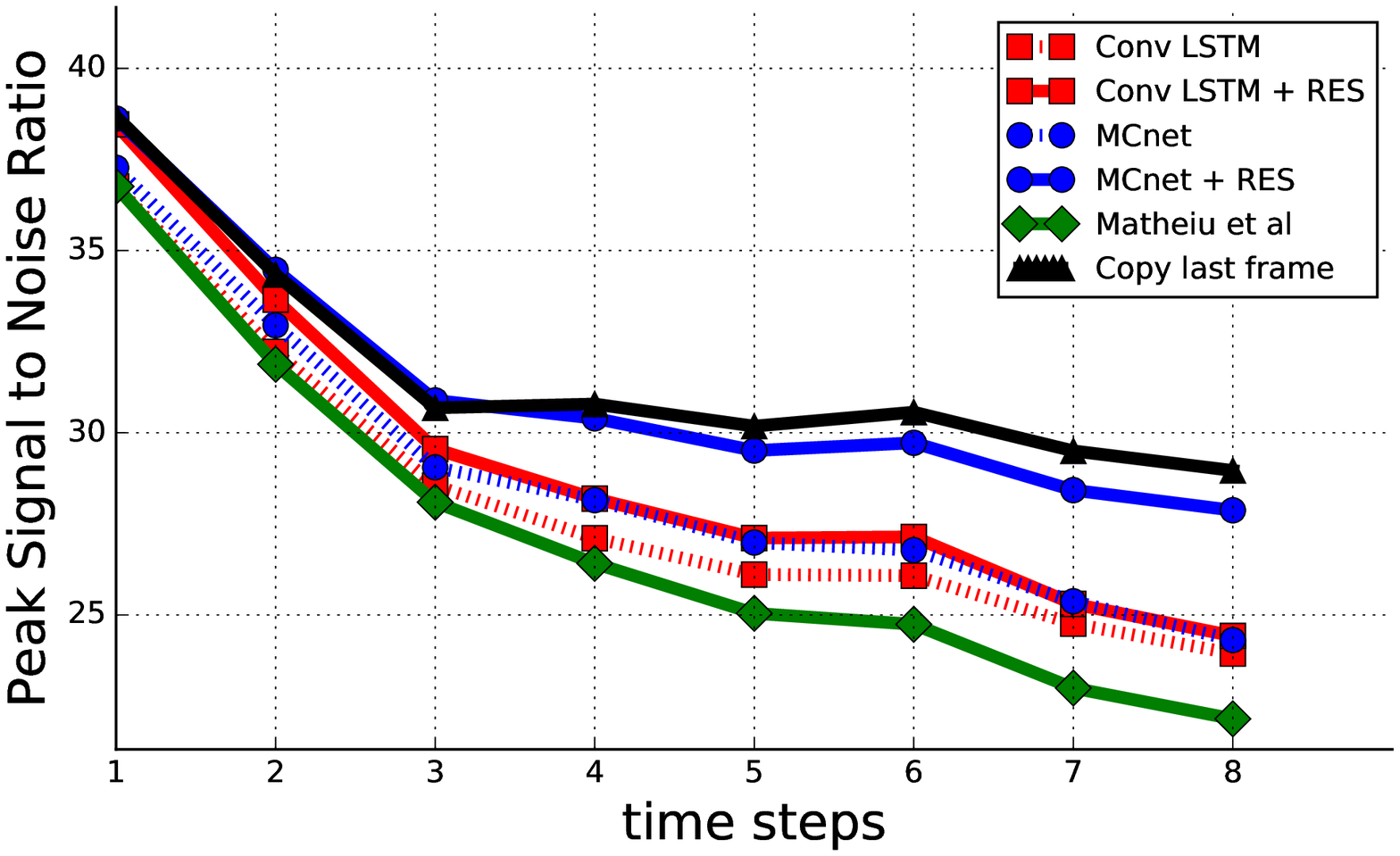}
\hspace{.1cm}
\includegraphics[width=0.49\linewidth] {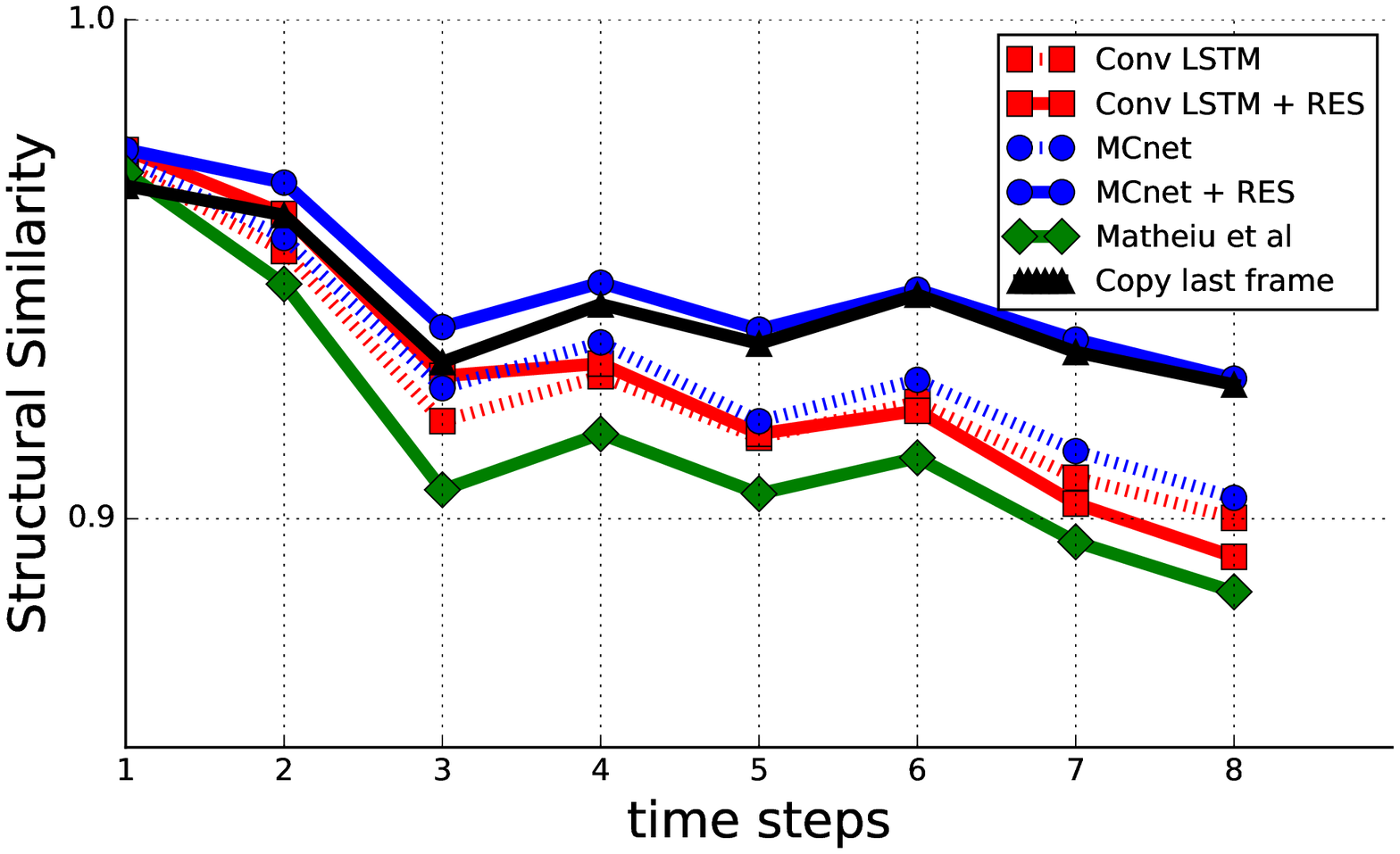}
\vspace{-.6cm}\hspace{0.1cm}
\caption*{$4^{th}$ decile}
\vspace{-.4cm}
\includegraphics[width=0.49\linewidth] {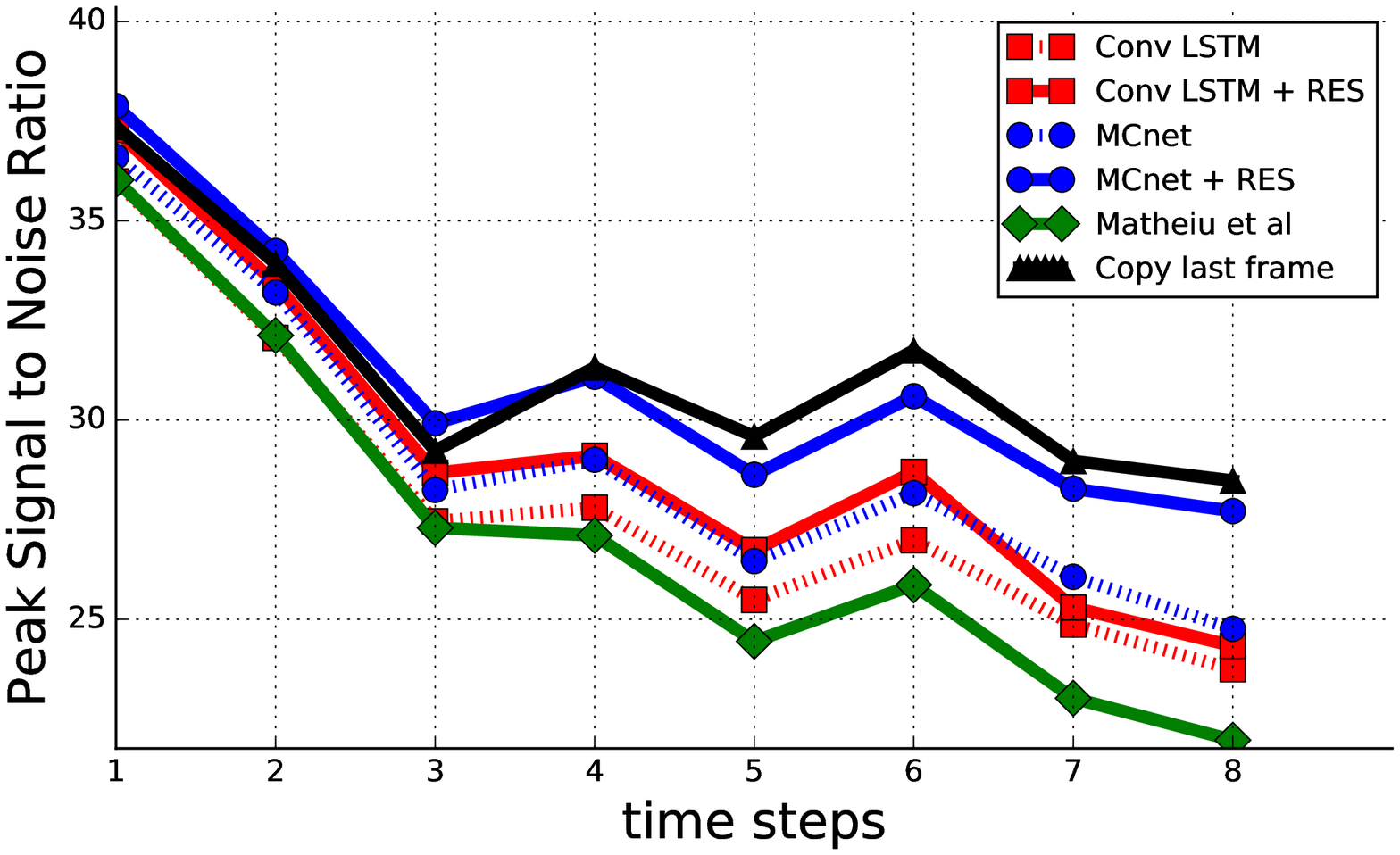} \hspace{.1cm}
\includegraphics[width=0.49\linewidth] {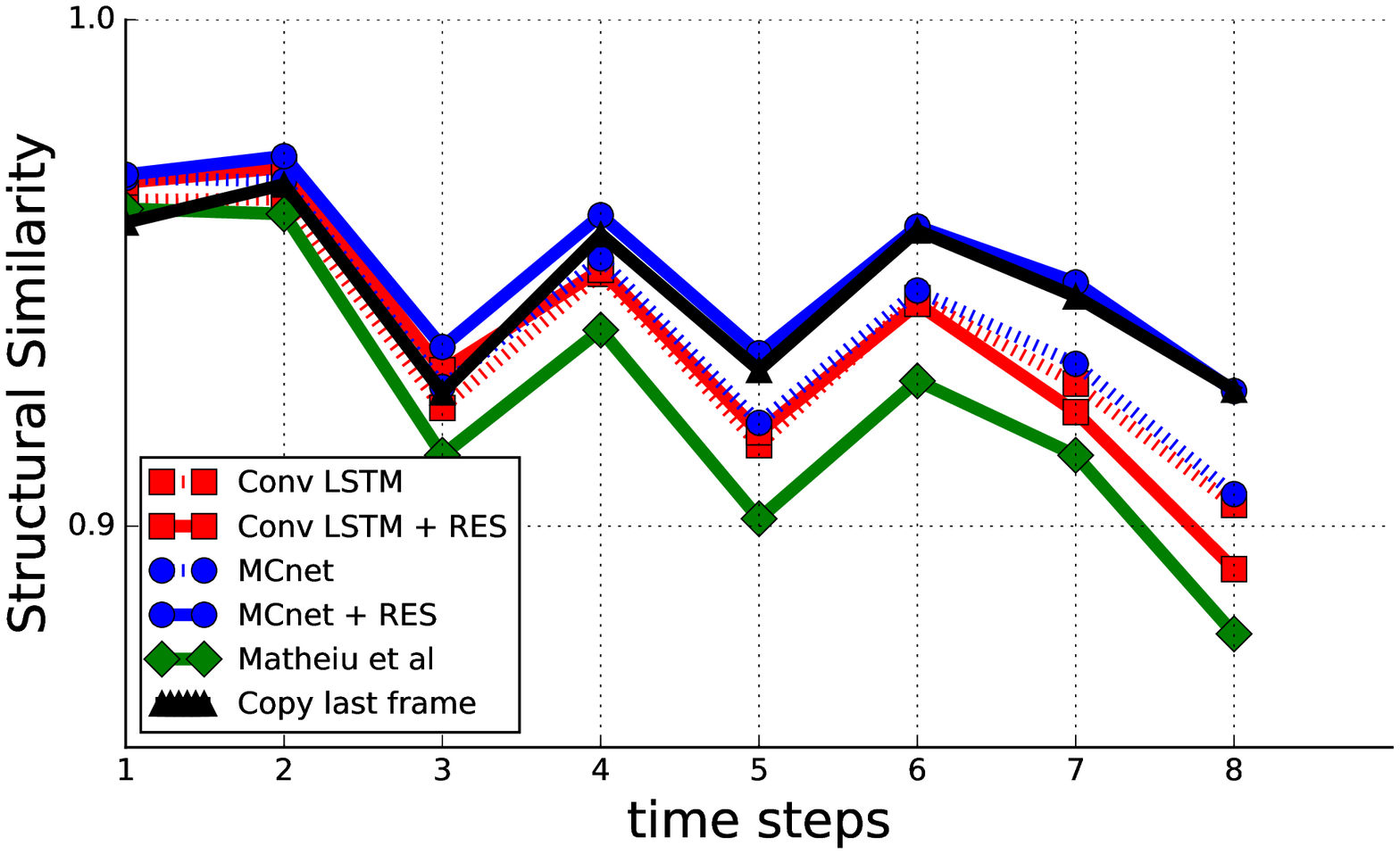} \vspace{-.6cm}\hspace{0.1cm}
\caption*{$3^{rd}$ decile}
\vspace{-.4cm}
\includegraphics[width=0.49\linewidth] {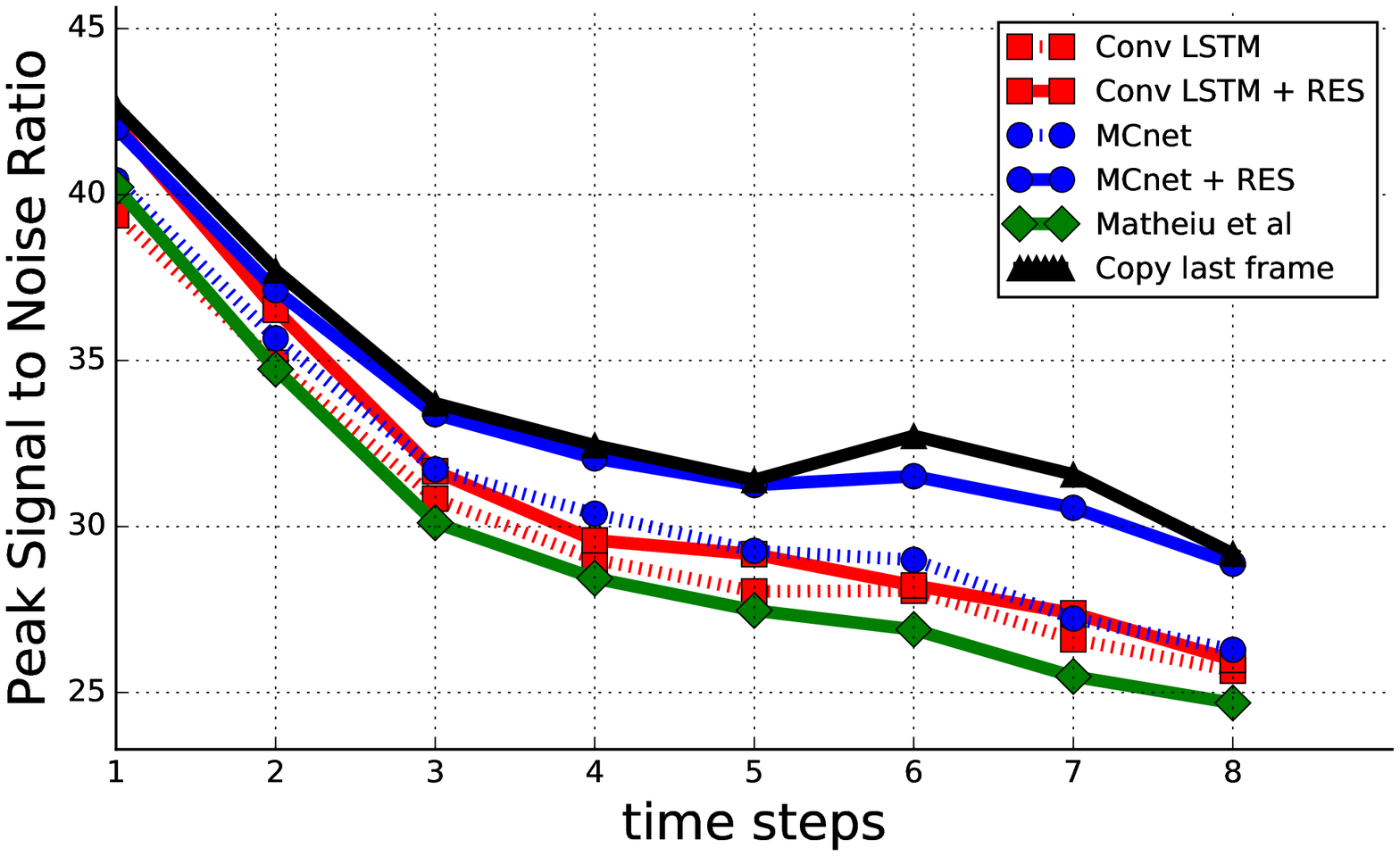} \hspace{.1cm}
\includegraphics[width=0.49\linewidth] {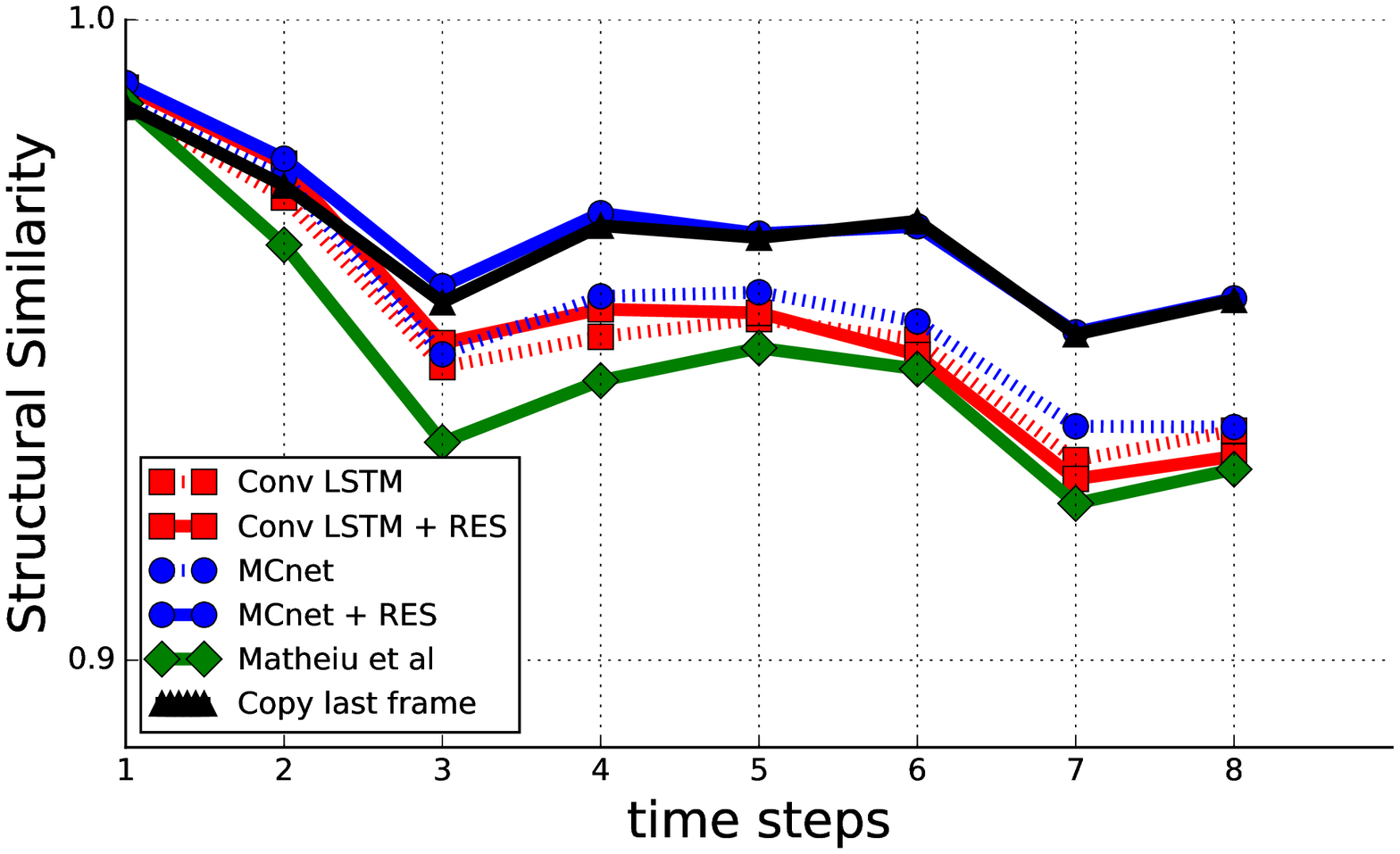} \vspace{-.6cm}\hspace{0.1cm}
\caption*{$2^{nd}$ decile}
\vspace{-.4cm}
\includegraphics[width=0.49\linewidth] {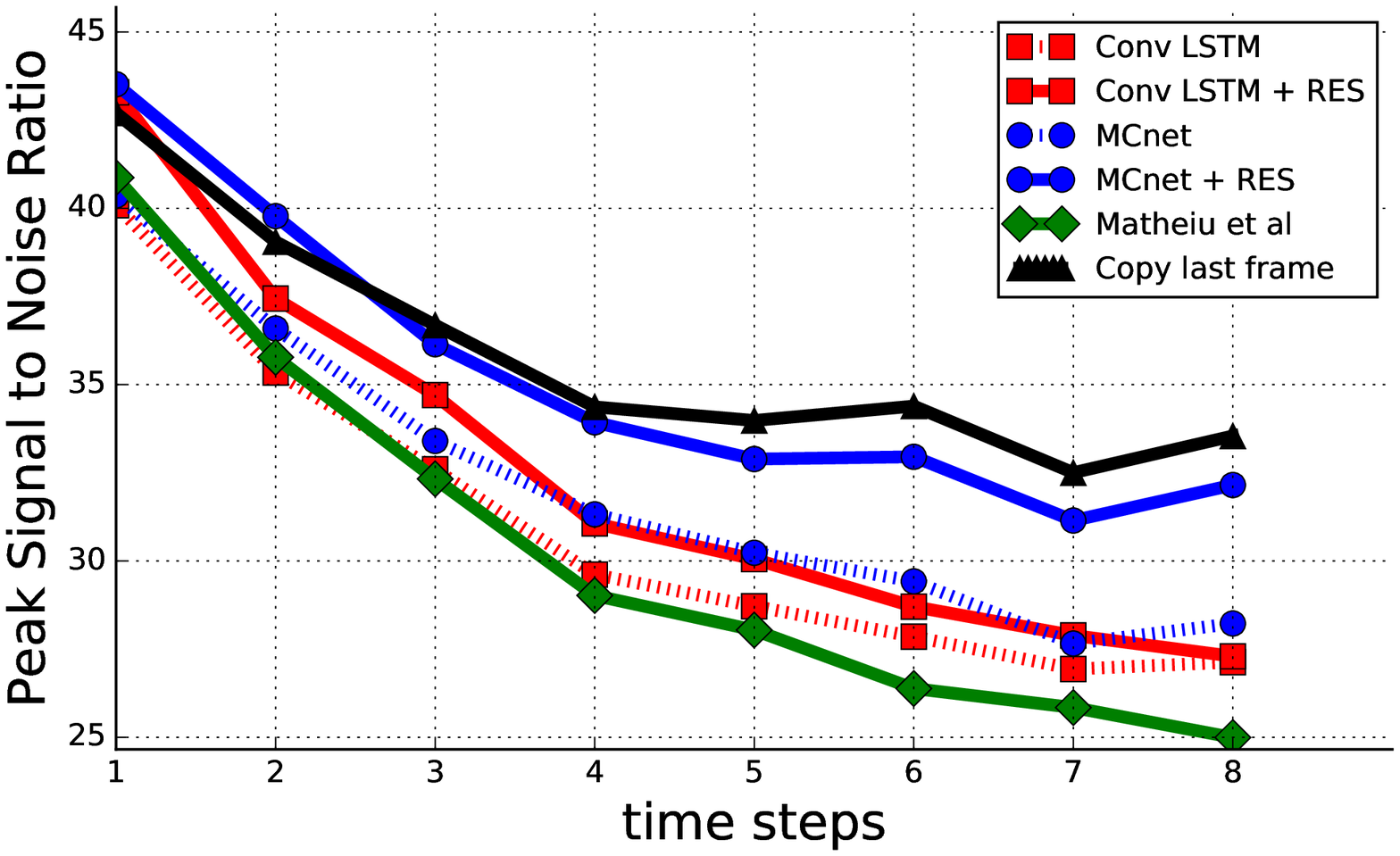} \hspace{.1cm}
\includegraphics[width=0.49\linewidth] {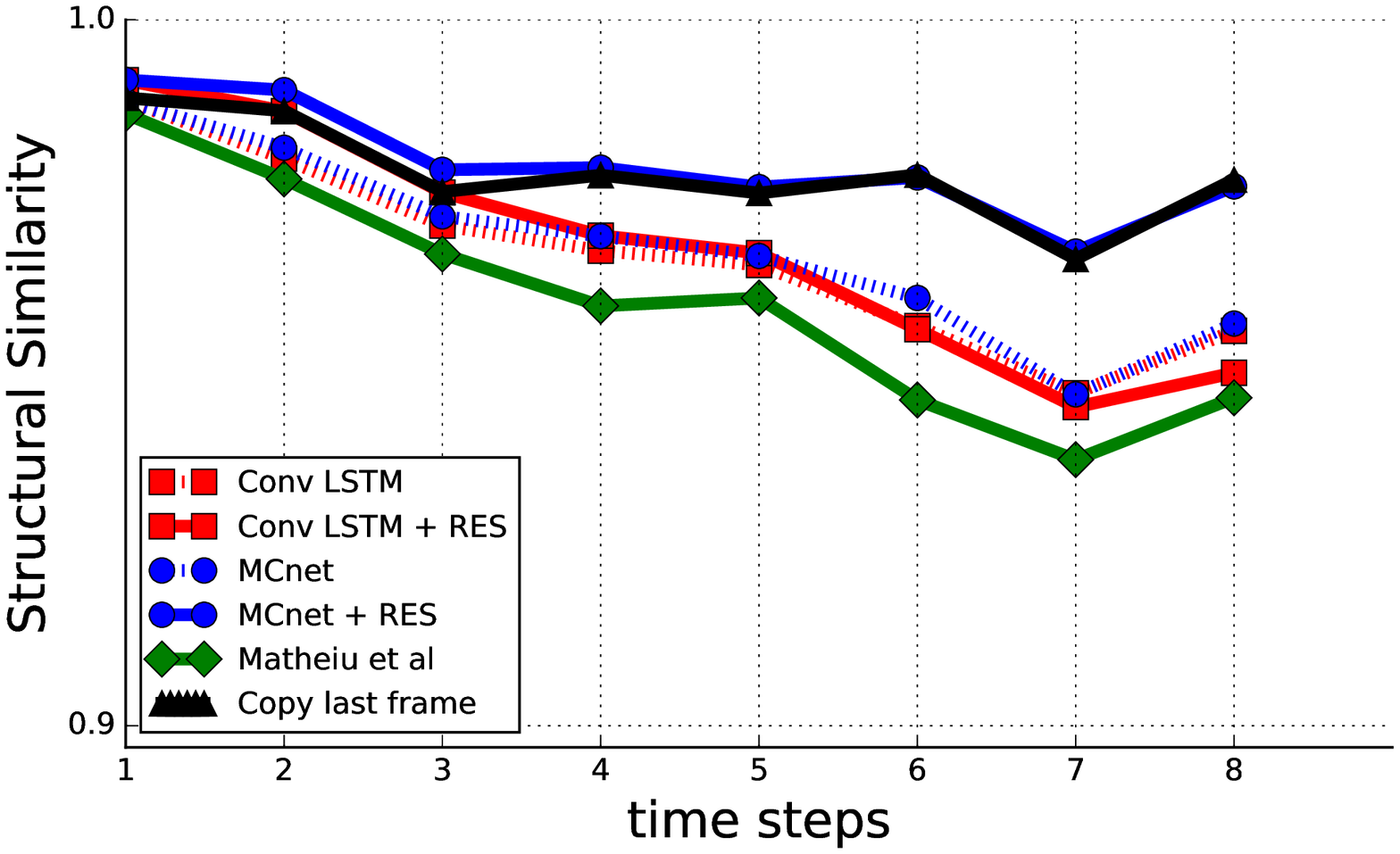} \vspace{-.6cm}\hspace{0.1cm}
\caption*{$1^{st}$ decile}
\vspace{-.4cm}
\includegraphics[width=0.49\linewidth] {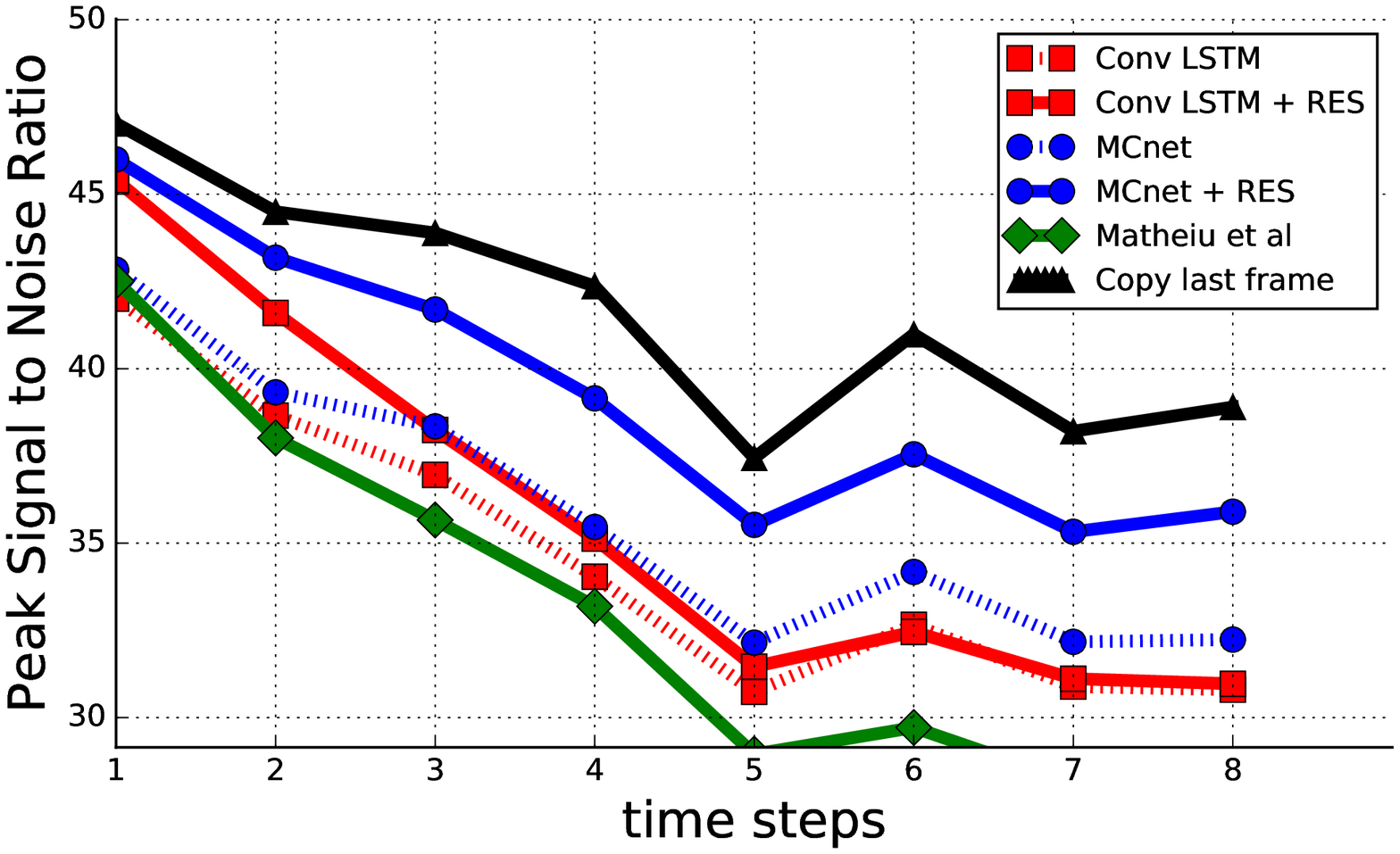} \hspace{.1cm}
\includegraphics[width=0.49\linewidth] {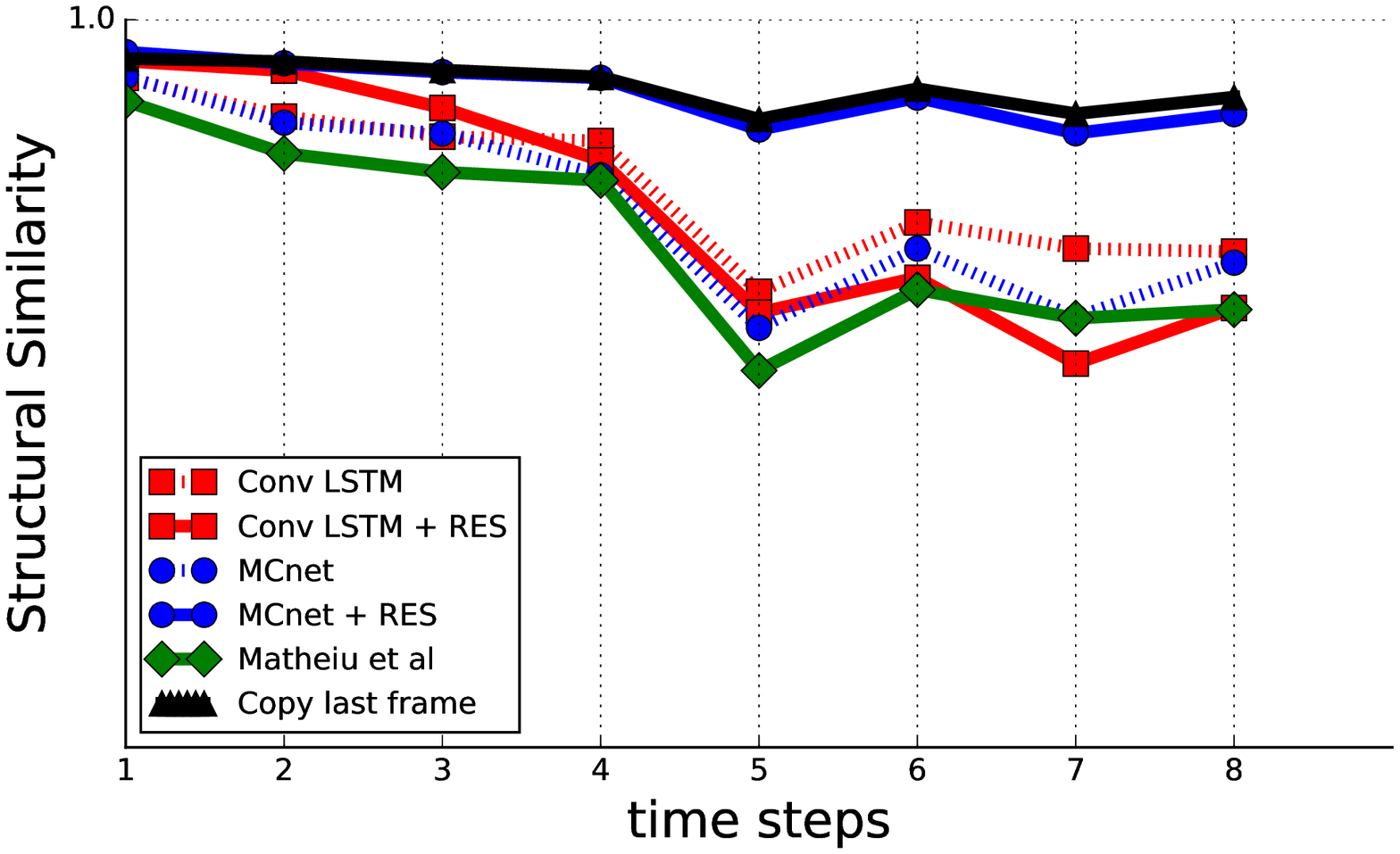} \hspace{0.1cm}
\caption{Quantitative comparison on UCF101 using motion based pixel mask, and separating dataset by average $\ell_2$-norm of time difference between target frames.}
\label{fig:extra_quantitative3}
\end{figure*}

Figures \ref{fig:extra_quantitative3} and \ref{fig:extra_quantitative4} show the evaluation by separating the test videos into deciles based on the average $\ell_2$-norm of time difference between target frames.
From this evaluation, it is proven that the copy last frame baseline scores higher in videos where motion is the smallest.
The first few deciles (videos with small motion) show that our network is not just copying the last observed frame through time, otherwise it would perform similarly to the copy last frame baseline.
The last deciles (videos with large motion) show our network outperforming all the baselines, including the copy last frame baseline, effectively confirming that our network does predict motion similar to the motion observed in the video.

\newpage

\clearpage

\section{Adversarial Training} \label{sec:GANs}
\cite{Mathieu15} proposed an adversarial training for frame prediction.
Inspired by \cite{NIPS2014_5423}, they proposed a training procedure that involves a generative model $G$ and a discriminative model $D$.
The two models compete in a two-player minimax game.
The discriminator $D$ is optimized to correctly classify its inputs as either coming from the training data (real frame sequence) or from the generator $G$ (synthetic frame sequence).
The generator $G$ is optimized to generate frames that \textit{fool} the discriminator into believing that they come from the training data.
At training time, $D$ takes the concatenation of the input frames that go into $G$ and the images produced by $G$.
The adversarial training objective is defined as follows:
\begin{equation*}
\min_{G} \max_{D} \ \ \log D\left(\left[\x_{1:t},\x_{t+1:t+T}\right]\right) +\log\left(1-D\left(\left[\x_{1:t},G\left(\x_{1:t}\right)\right]\right)\right),
\end{equation*}
where $\left[.,.\right]$ denotes concatenation in the depth dimension, $\x_{1:t}$ denotes the input frames to $G$, $\x_{t+1:t+T}$ are the target frames, and $G\left(\x_{1:t}\right)=\hat{\x}_{t+1:t+T}$ are the frames predicted by $G$.
In practice, we split the minimax objective into two separate, but equivalent, objectives: $\mathcal{L}_{\text{GAN}}$ and $\mathcal{L}_{\text{disc}}$.
During optimization, we minimize the adversarial objective alternating between $\mathcal{L}_{\text{GAN}}$ and $\mathcal{L}_{\text{disc}}$.
$\mathcal{L}_{\text{GAN}}$ is defined by
\begin{equation*}
\mathcal{L}_{\text{GAN}} = -\log D\left(\left[\x_{1:t},G\left(\x_{1:t}\right)\right]\right) ,
\end{equation*}
where we optimize the parameters of $G$ to minimize $\mathcal{L}_{\text{GAN}}$ while the parameters of $D$ stay untouched.
As a result, $G$ is optimized to generate images that make $D$ believe that they come from the training data.
Thus, the generated images look sharper, and more realistic.
$\mathcal{L}_{\text{disc}}$ is defined by
\begin{equation*}
\mathcal{L}_{\text{disc}} = -\log D\left(\left[\x_{1:t},\x_{t+1:t+T}\right]\right) -\log\left(1-D\left(\left[\x_{1:t},G\left(\x_{1:t}\right)\right]\right)\right),
\end{equation*}
where we optimize the parameters of $D$ to minimize $\mathcal{L}_{\text{disc}}$, while the parameters of $G$ stay untouched.
$D$ tells us whether its input came from the training data or the generator $G$.
Alternating between the two objectives, causes $G$ to generate very realistic images, and $D$ not being able to distinguish between generated frames and frames from the training data.

\end{appendix}

\end{document}